\documentclass[letterpaper]{article} 
\usepackage[]{aaai24}  
\usepackage{times}  
\usepackage{helvet}  
\usepackage{courier}  
\usepackage[hyphens]{url}  
\usepackage{graphicx} 
\usepackage{natbib}  
\usepackage{caption} 
\usepackage{algorithm}
\usepackage{algorithmic}
\usepackage{graphicx}
\usepackage{subcaption}

\usepackage{amsfonts}
\usepackage{amsmath}
\usepackage{bm}
\usepackage[T1]{fontenc}

\usepackage{newfloat}
\usepackage{listings}
\DeclareCaptionStyle{ruled}{labelfont=normalfont,labelsep=colon,strut=off} 
\floatstyle{ruled}
\newfloat{listing}{tb}{lst}{}
\floatname{listing}{Listing}
\usepackage{hyperref}

\title{Dimma: Semi-supervised Low Light Image Enhancement with Adaptive Dimming}

\author {
    Wojciech Koz{\l}owski\textsuperscript{\rm 1}\equalcontrib,
    Micha{\l} Szachniewicz\textsuperscript{\rm 1}\equalcontrib,
    Micha{\l} Stypu{\l}kowski\textsuperscript{\rm 2},
    Maciej Zi\k{e}ba\textsuperscript{\rm 1, 3}
}
\affiliations {
    \textsuperscript{\rm 1}Wroclaw University of Science and Technology\\
    \textsuperscript{\rm 2}University of Wroclaw\\
    \textsuperscript{\rm 3}Tooploox
}

\usepackage{bibentry}

\begin{document}

\maketitle

\begin{abstract}
Enhancing low-light images while maintaining natural colors is a challenging problem due to camera processing variations and limited access to photos with ground-truth lighting conditions. The latter is a crucial factor for supervised methods that achieve good results on paired datasets but do not handle out-of-domain data well. On the other hand, unsupervised methods, while able to generalize, often yield lower-quality enhancements.
To fill this gap, we propose Dimma, a semi-supervised approach that aligns with any camera by utilizing a small set of image pairs to replicate scenes captured under extreme lighting conditions taken by that specific camera. We achieve that by introducing a convolutional mixture density network that generates distorted colors of the scene based on the illumination differences. Additionally, our approach enables accurate grading of the dimming factor, which provides a wide range of control and flexibility in adjusting the brightness levels during the low-light image enhancement process.
To further improve the quality of our results, we introduce an architecture based on a conditional UNet. The lightness value provided by the user serves as the conditional input to generate images with the desired lightness.
Our approach using only few image pairs achieves competitive results compared to fully supervised methods. Moreover, when trained on the full dataset, our model surpasses state-of-the-art methods in some metrics and closely approaches them in others.\footnote{The code and pretrained models are available publicly at \href{https://github.com/WojciechKoz/Dimma}{github.com/WojciechKoz/Dimma}.}
\end{abstract}

\section{Introduction}
\begin{figure*}
    \centering
    \includegraphics[width=0.95\textwidth]{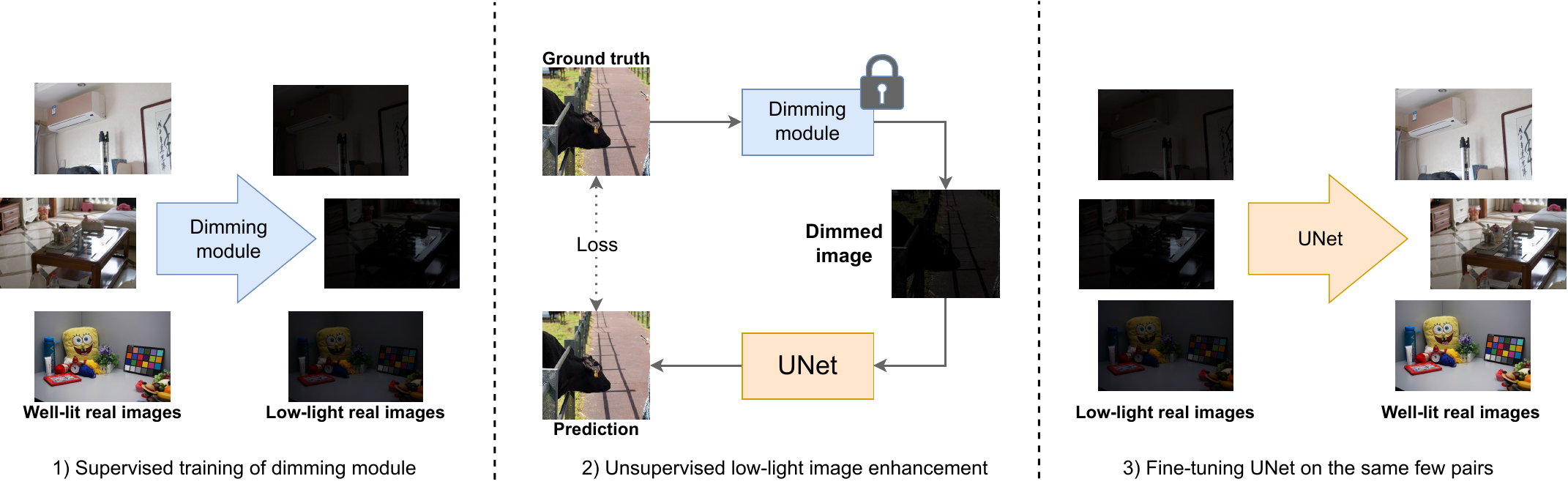}
    \caption{High-level picture of our proposed method. 1) We train the dimming module to mimic color distortion made by a specific camera. 2) We employ this dimming module to train a UNet \cite{unet} architecture in an unsupervised fashion to restore the original image. Note, that we could use any image in the second phase, making the model more generic despite having only a few real image pairs. 3) Then, we use the same real image pairs from the first stage to finetune the UNet.}
    \label{fig:big_picture}
\end{figure*}
The low-light image enhancement task aims to replicate the appearance of a photo taken with a longer exposure by reducing noise, increasing brightness, and preserving natural colors. Many research papers \cite{retinex_net, kind, kindpp, llflow, cai2023retinexformer} have reported promising results on paired datasets containing well-lit and dark images. However, the effectiveness of models trained on one dataset may not generalize well to images from different cameras. Variations in how individual cameras process images pose a challenge to achieving consistent and reliable enhancement across diverse camera models. Creating a comprehensive set of paired images for numerous devices is not only challenging but also expensive. Furthermore, relying on a single ground truth for each pair limits the model's ability to learn the optimal brightening factor accurately.

To address these challenges, we propose a semi-supervised approach (Figure \ref{fig:big_picture}) that can leverage a small set of real paired images to construct a dimming module capable of replicating the camera-specific dimming process. By utilizing a mixture density network and illumination statistics extracted from the real image pairs, the dimming module can effectively darken an image and mimic the color distortion associated with very dark images from a selected source.

Next, we create dimmed versions of images from various computer vision datasets using the constructed dimming module and train a UNet \cite{unet} model to restore their original light exposure.
Our UNet architecture incorporates a brightness level conditioning to represent light differences and predict the contrast between dark and light images, rather than directly predicting the light values themselves. This approach allows the model to learn conditioned image light enhancement, providing flexibility and adaptability to various lighting conditions.

Our proposed approach offers a practical and cost-effective solution for training image enhancement models to different cameras. The utilization of a dimming module based on a few real paired images enables the model to mimic the camera-specific dimming process accurately. Furthermore, the ability to adjust the brightening factor provides additional control over the image enhancement process, which is not feasible when relying solely on paired datasets.

Through extensive experimentation and evaluation, we demonstrate the effectiveness of our approach in enhancing low-light images while preserving natural colors. By addressing the challenges posed by camera variations, the limited availability of paired datasets, and the control over the brightening factor, our research contributes to advancing the field of low-light image enhancement and provides a practical and adaptable solution for real-world applications.

In summary, our contributions can be outlined as follows:
\textbf{(i)} We introduce a dimming module based on a mixture density network \cite{mdn} and illumination statistics calculated from a small set of ground truth image pairs. The dimming module effectively replicates the color distortion and darkening process of very dark images for any camera. 
\textbf{(ii)} We propose to use UNet \cite{unet} conditioned on light differences that predicts the residuals between dark and light images rather than the light values themselves.
\textbf{(iii)} We introduce two datasets: FewShow-Dark (FS-Dark) dataset containing a few real image pairs captured by mobile phone camera and MixHQ dataset, which comprises high-quality images selected from various datasets including COCO \cite{coco}, ImageNet \cite{imagenet}, Clic \cite{clic}, Inter4K \cite{inter4k}, LOL \cite{retinex_net}. Combined with our dimming module, MixHQ allows to create artificial training pairs for semi-supervised low-light image enhancement.
\textbf{(iv)} We combine a classic approach to retinex decomposition \cite{lime} with deep learning techniques, achieving state-of-the-art results on LOL, and VE-LOL datasets \cite{retinex_net} with fully supervised training and develop models supervised with as few as 3 ground truth image pairs competitive with existing fully supervised methods.



\begin{figure*}[t]
    \centering
    \includegraphics[width=0.9\textwidth]{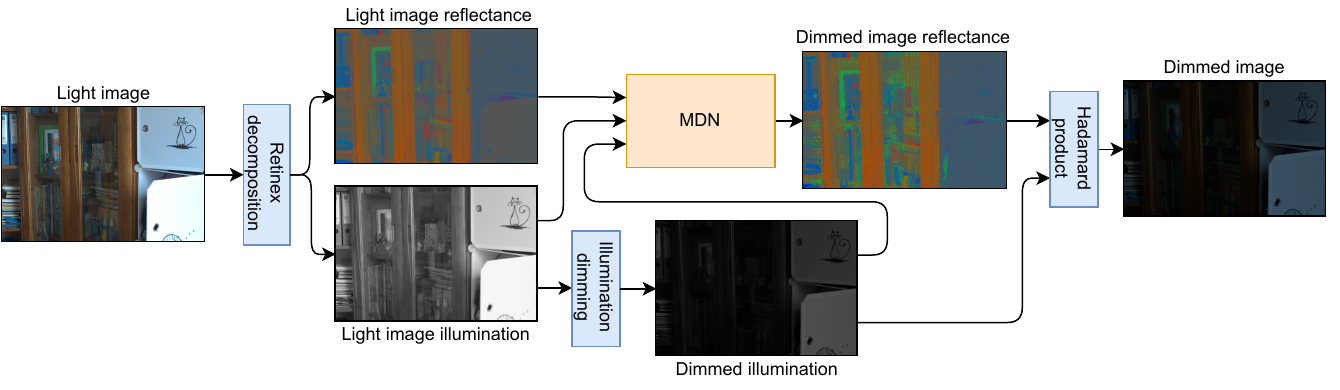}
    \caption{The dimming module utilizes camera color distortion information from a trained MDN model and dynamic illumination statistics calculated from a training set. It achieves a similar dimming effect to the original device, making it ideal for training brightening models on unlabeled images.}
    \label{fig:dimming_module}
\end{figure*}
\section{Related work}

Low-light image enhancement is a widely recognized task in the field of computer vision, attracting significant attention from researchers. Over the years, various classic approaches have emerged, including techniques such as histogram equalization, gamma correction, and more complex algorithms such as LIME (Low-light Image Enhancement) \cite{lime}. These classic methods are known for their remarkable speed and generalizability. However, they often produce images of relatively lower quality compared to more recent deep learning techniques.

Deep learning-based approaches for low-light image enhancement have gained popularity due to their ability to generate visually appealing images. Several methods \cite{retinex_net, kind, kindpp, llflow} have achieved notable results in this domain. In RetinexNet \cite{retinex_net}, the authors proposed to make use neural network for retinex decomposition. The model is trained to reconstruct both reflectance and illumination to match the well-lit images' maps. KinD and KinD++ \cite{kind, kindpp} share a similar idea but uses separate modules for each of the maps.

The current state-of-the-art supervised method, LLFlow \cite{llflow}, utilizes a normalizing flow that learns a distribution of images under normal light conditions given a low-light one. In addition, an encoder is used to extract an illumination-invariant color map that is then injected into the normalizing flow.

These approaches typically rely on datasets consisting of paired images, where one image is captured under low-light conditions, and the other is a corresponding well-lit reference. By learning the mapping from dark to well-lit images, these models can enhance low-light images effectively. However, it is important to note that these paired datasets represent the distribution specific to the cameras used, limiting the generalization capabilities of these approaches to other camera models.


Despite the challenges associated with low-light image enhancement, there have been attempts to develop unsupervised approaches using deep learning techniques. For example, EnlightenGAN \cite{enlightengan} proposes to leverage a large dataset of unpaired light and dark images and employs adversarial training to enhance low-light images. 

While unsupervised models offer flexibility and the potential to generalize to different cameras, due to the inherent nature of adversarial training, they often fall short in terms of producing high-quality images compared to approaches that employ ground-truth-based loss functions. The lack of explicit supervision in unsupervised methods makes it challenging to achieve the desired level of image quality and natural color preservation.

In \cite{retinex_net, kind, kindpp, enlightengan}, authors used UNet \cite{unet}, a convolutional architecture, originally created for the object segmentation task. The skip connections allow modeling features of lower frequencies, producing results of a higher visual quality.

In \cite{xu2022snr, cai2023retinexformer} the architecture of transformer was used instead of pure convolutional networks.
Zero-DCE \cite{zero-dce} instead of using UNet formulates the low-light image enhancement as monotonic curves estimation using a lightweight neural network. Authors of \cite{ruas} also focused on developing fast and lightweight model.

\section{Our method}

This section provides a detailed description of Dimma, a novel model for enhancing low-light images. Our approach comprises two key components:  dimming and brightening modules (see Figure \ref{fig:big_picture}). The first module mimics color distortion made by specific cameras to create darkened images. These are further used to train the brightening module represented by the UNet network responsible for restoring the original image. 

\subsection{Dimming module}
\paragraph{Retinex decomposition} Previous works \cite{retinex_net, kind, kindpp} have shown that the retinex decomposition of an image $\mathbf{I}$ to illumination $\mathbf{L}$ and reflectance $\mathbf{R}$ helps models to achieve a better quality of restored images. However, most of the works attempt to do this using neural networks making it very slow and inaccurate. Our approach is based on channel-wise normalization introduced in \cite{llflow} in which the $i$-th channel of reflectance $\mathbf{R}$ can be calculated as:

\begin{equation}
\mathbf{R}_i = \mathbf{I}_i \oslash (\frac{1}{3}\sum_{i=1}^3 \mathbf{I}_i)
\end{equation}
where $\oslash$ represents Hadamard (element-wise) division, $\mathbf{I}_i$ and $\mathbf{R}_i$ represent $i$-th channel of image $\mathbf{I}$ and reflectance $\mathbf{R}$ respectively. Assuming that the color map is a light-invariant reflectance known from retinex theory and taking into account that image is a Hadamard product of reflectance and illumination, we get the illumination equal $\mathbf{L}=\mathbf{I} \oslash \mathbf{R}$. We achieve similar results extremely quickly and without any errors related to restoration quality produced by the autoencoder. 

\begin{figure*}
    \centering
    \includegraphics[width=0.9\textwidth]{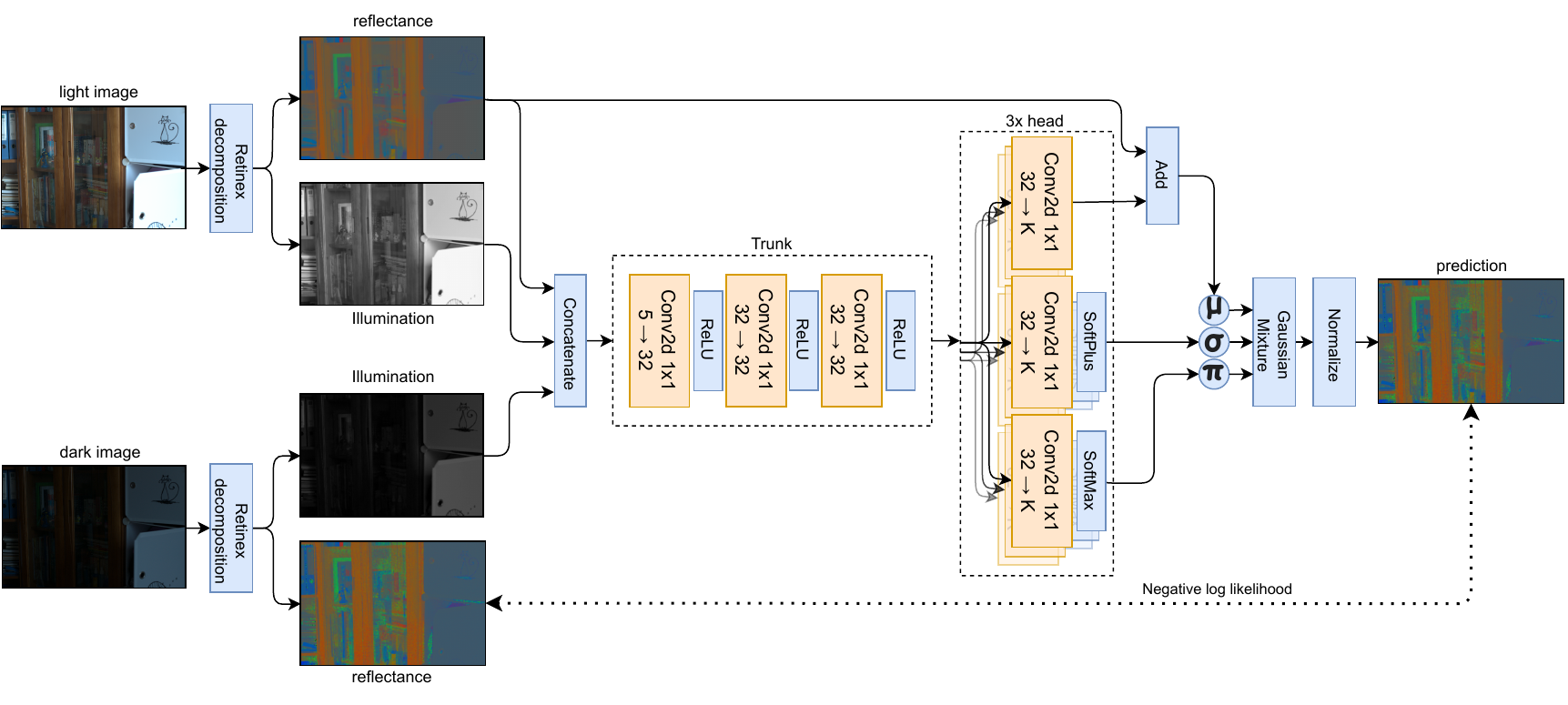}
    \caption{Mixture density network (MDN) - model used in our dimming module for modeling reflectance distortion of low-light images based on light image reflectance and illuminations from both images.}
    \label{fig:mdn}
\end{figure*}

\paragraph{Dimming process} In Figure \ref{fig:dimming_module}, we present the dimming procedure that creates the corresponding dark images essential for training the brightening module in the second stage. The initial photo is split into illumination $\mathbf{L}$ and reflectance $\mathbf{R}$ components using retinex decomposition described in the previous paragraph. Next, illumination dimming is applied on $\mathbf{L}$ to create the darker equivalent $\mathbf{L}_D$. In order to align the dimming process to specific camera conditions, we utilize Mixture Density Network (MDN), which is responsible for adjusting reflectance $\mathbf{R}$. This module is trained in supervised mode using a few labeled examples, for which the corresponding ground truth reflectance is known. The output dark image $\mathbf{I}_D$ is created by the application of Hadamard product between dark illumination $\mathbf{L}_D$ and reflectance $\mathbf{R}_D$ sampled from the MDN component, $\mathbf{I}_D=\mathbf{R}_D \odot \mathbf{L}_D$.

\paragraph{Mixture Density Network (MDN)} The architecture of MDN network is provided in Figure \ref{fig:mdn}. This component is responsible for calibrating the reflectance crucial to create a darker image. We postulate to use a probabilistic model due to the following reasons. First, we aim to mimic the noise produced by the camera on the same scene in low-light conditions. Second, we train this component using only a few examples, so the precise reflectance prediction may be difficult to obtain using limited data. Third, the randomness injected into the process of generating dark images serves as a regularization technique for the brightening module. The ablation study (Table \ref{tab:fewshot_results}) shows the rationality of our approach.

The MDN module takes reflectance $\mathbf{R}$ from the light image $\mathbf{I}$ and illuminations $\mathbf{L}$ and $\mathbf{L}_D$ from both light and dark images. This enables the network to learn color noise, considering the local brightness of the source and target image. This approach is motivated by the observation that dark regions in photos tend to be more noisy. The reflectance $\mathbf{R}$ and illuminations $\mathbf{L}$ and $\mathbf{L}_D$ are concatenated to create 5-channel input. To make the model as general as possible and to avoid overfitting while training using only a few images, we use only 1x1 convolution filters in MDN architecture. Practically, it means that we process 5-dimensional representations for each pixel independently, with the MLP module that shares the weights across the pixels. Let denote the 5 dimensional representation $\mathbf{x}_{i,j}=[\mathbf{r}_{i,j},l_{i,j},l_{D,i,j}]$ that concatenates the pixel values at $(i,j)$ location from input reflectance $\mathbf{R}$, and illuminations $\mathbf{L}$ and $\mathbf{L}_D$ respectively.
The distribution over $r_{D,i, j, k}$ representing the scalar element of $\mathbf{R}_D$ on $(i,j)$-th position in channel $k$ can be described as:

\begin{align}
\label{eq:r_prob}
p(r_{D,i, j, k}|\mathbf{x}_{i, j}; \boldsymbol \theta)= \sum_{m=1}^M \pi_{k, m} (\mathbf{x}_{i, j};\boldsymbol \theta) \phi_{D, i, j, k, m},
\end{align}
where $M$ is the number of components in mixture of Gaussians, $\boldsymbol \theta$ are the parameters of MDN network, and the density of a single Gaussian $\phi_{D, i, j, k, m}$ is defined as:
\begin{align}
 \mathcal{N}(r_{i,j,k} + \mu_{k,m}(\mathbf{x}_{i, j};\boldsymbol \theta),\sigma_{k,m}(\mathbf{x}_{i, j};\boldsymbol \theta)),
\end{align}
where $\pi_{k,m} (\mathbf{x}_{i, j};\boldsymbol \theta)$, $\mu_{k,m}(\mathbf{x}_{i, j};\boldsymbol \theta)$ and $\sigma_{k,m}(\mathbf{x}_{i, j};\boldsymbol \theta))$ are the parameters of Gaussian mixture predicted by MDN. Each color channel has its own head, producing means $\mu_{k,m}(\mathbf{x}_{i, j};\boldsymbol \theta)$, standard deviations $\sigma_{k,m}(\mathbf{x}_{i, j};\boldsymbol \theta))$, and mixing coefficients $\pi_{k,m} (\mathbf{x}_{i, j};\boldsymbol \theta)$. The means produced by the model are added to the original color map to make the network predict color differences instead of the total reflectance, which could not be possible with the limited training data. For the same reason, we trained the model to predict color maps instead of entire dark images, which makes it more invariant to illumination conditions in the training data. The joined distribution over $\mathbf{R}_D$ can be defined as:

\begin{equation}
p(\mathbf{R}_D|\mathbf{R},\mathbf{L}, \mathbf{L}_D; \boldsymbol \theta) =\prod_{i,j,k} p(r_{D,i, j, k}|\mathbf{x}_{i, j}; \boldsymbol \theta)
\label{eq:R_prob}
\end{equation}

Assume access to few ($K$) labelled training examples $\mathcal{D}_K=\{(\mathbf{I}^{(n)}, \mathbf{I}_D^{(n)})\}_{n=1}^K=\{(\mathbf{R}^{(n)}, \mathbf{L}^{(n)}, \mathbf{R}_D^{(n)},\mathbf{L}_D^{(n)})\}_{n=1}^K$, composed of light $\mathbf{I}^{(n)}$ and corresponding dark images $\mathbf{I}_D^{(n)}$. We optimize the parameters $\boldsymbol{ \theta}$ of MDN network by minimizing conditional negative log-likelihood:

\begin{equation}
\boldsymbol{ \theta^*}=\arg \min_{\boldsymbol{ \theta}}\sum_{n} \sum_{i,j,k} -\log{p(r_{D,i, j, k}^{(n)}|\mathbf{x}_{i, j}^{(n)}; \boldsymbol \theta)}
\end{equation}
where $p(r_{D,i, j, k}^{(n)}|\mathbf{x}_{i, j}^{(n)}; \boldsymbol \theta)$ is given by Equation \ref{eq:r_prob}. The vector $\mathbf{x}_{i,j}^{(n)}=[\mathbf{r}_{i,j}^{(n)},l_{i,j}^{(n)},l_{D,i,j}^{(n)}]$  concatenates the pixel values at $(i,j)$ location from $n$-th example reflectance $\mathbf{R}^{(n)}$, and illuminations $\mathbf{L}^{(n)}$ and $\mathbf{L}_D^{(n)}$.

\subsection{Brightening module}
\begin{figure*}
    \centering
    \includegraphics[width=0.9\textwidth]{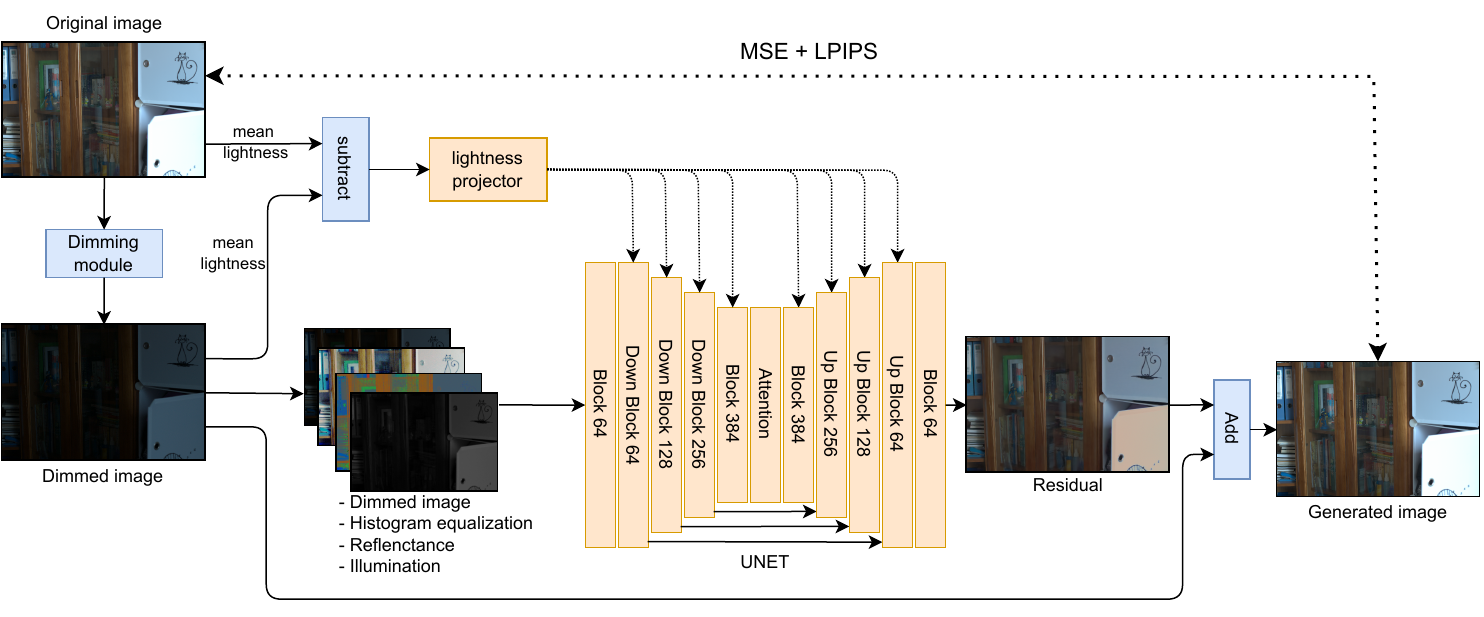}
    \caption{Architecture of our brightening module. In the training process, we used the difference between original and dimmed image lightness to learn the model to understand the concept of lightness. Later, in the inference mode, this is provided by the user.}
    \label{fig:unet-diag}
\end{figure*}

\begin{table*}[t]
\centering
\begin{tabular}{cccccccc}
Method & PSNR $\uparrow$ & SSIM $\uparrow$ & RGB-SSIM $\uparrow$ & LPIPS $\downarrow$ & DeltaE $\downarrow$ & NIQE $\downarrow$ & Train pairs $\downarrow$ \\
\hline
RetinexNet & 16.77 / 14.68 & 0.52 / 0.46 & 0.42 / 0.36 & 0.47 / 0.65 & 15.89 / 20.98 & 9.73 / 10.36 & 485 \\
KinD & 17.65 / 18.42 & 0.83 / 0.75 & 0.77 / 0.70 & 0.18 / 0.29 & 12.49 / 13.73 & 3.89 / \textbf{4.09} & 450 \\
Zero-DCE & 14.86 / 16.07 & 0.65 / 0.58 & 0.56 / 0.47 & 0.34 / 0.48 & 18.82 / 18.51 & 8.22 / 8.92 & 360 \\
EnlightenGAN & 17.48 / 16.90 & 0.70 / 0.62 & 0.65 / 0.56 & 0.32 / 0.43 & 14.48 / 16.45 & 4.89 / 6.88 & 0 \\
RUAS & 16.40 / 13.16 & 0.70 / 0.56 & 0.50 / 0.41 & 0.27 / 0.48 & 16.83 / 22.25 & 5.93 / 8.06 & 100 \\
KinD++ & 21.80 / 20.41 & 0.88 / \textbf{0.79} & 0.83 / \textbf{0.76} & 0.16 / \underline{0.28} & 8.50 / 11.37 & 4.00 / 4.47 & 460 \\
SNR-Net & 24.61 / \underline{22.93} & 0.90 / 0.76 & 0.84 / 0.70 & \underline{0.15} / 0.32 & 6.85 / 11.32 & 4.02 / \underline{4.14} & 485 \\
LLFlow & \underline{25.19} / 22.38 & \textbf{0.93} / 0.73 & \textbf{0.86} / 0.69 & \textbf{0.11} / 0.32 & \underline{6.40} / \underline{11.13} & 4.08 / 5.83 & 485 \\
Retinexformer$^*$ & 25.16 / - & - / - & \underline{0.85} / - & - / - & - / - & - / - & 485 \\
Dimma 3 pairs & 23.54 / 20.18 & 0.83 / 0.72 & 0.76 / 0.65 & 0.26 / 0.38 & 9.20 / 13.24 & 3.93 / 5.30 & 3 \\
Dimma 5 pairs & 24.49 / 20.56 & 0.84 / 0.72 & 0.76 / 0.66 & 0.25 / 0.37 & 7.98 / 12.90 & 3.81 / 5.17 & 5 \\
Dimma 8 pairs & 24.70 / 20.75 & 0.86 / 0.74 & 0.78 / 0.67 & 0.23 / 0.36 & 7.81 / 12.84 & \underline{3.56} / 5.14 & 8 \\
Dimma full & \textbf{27.39} / \textbf{25.00} & \underline{0.91} / \underline{0.77} & \textbf{0.86} / \underline{0.72} & \textbf{0.11} / \textbf{0.26} & \textbf{5.54} / \textbf{9.63} & \textbf{3.14} / 4.79 & 480 \\
\end{tabular}
\caption{Quantitive comparison on LOL and VE-LOL datasets separated by slashes. All of the models were trained on LOL. The best results are bold, the second best are underlined. $^*$Limited evaluation of Retinexformer caused by lack of a source code.}
\label{tab:results_lol}
\end{table*}

Once the dimming module has been trained, it can be utilized for unsupervised training of a brightening model, as depicted in Figure \ref{fig:big_picture}.
We chose to use UNet \cite{unet} as it proved to be a very good architecture for low light image enhancement task \cite{sid, kind, kindpp}. We decided to further improve it by using conditioning known from diffusion models \cite{ho2020denoising, nichol2021improved} to give the model information about the desired target lightness.
By injecting lightness into feature maps, the model acquires a lot of information regarding the image's final appearance. Consequently, it can generate more visually realistic images across various brightness levels. 
Previous works attempted to achieve this by scaling the dark image illumination \cite{sid, kind, llflow}, which is a more naive approach and could not lead to visually appealing images with varying brightness levels that differ from those present in the training set. The detailed architecture of our brightening module is shown in Figure \ref{fig:unet-diag}.

Denote $U_{\boldsymbol{\beta}}$ as a brightening UNet network with parameters $\boldsymbol{\beta}$. Inspired by \cite{llflow}, our model takes the concatenation of the original dark image $\mathbf{I}_{D}$, its histogram equalization $H(\mathbf{I}_{D})$, color map $\mathbf{R}_D$, and illumination as input $\mathbf{L}_D$. Compared to reference approaches, our model also takes a lightness degree $\Delta m$ that controls the light level of the output image. This information is injected into UNet in a similar way as time embedding is used in diffusion models. Thanks to this approach, we can control the level of lightness while enhancing the low-light image. The model returns the residual map $\mathbf{U}$ that is further added to the dark image $\mathbf{I}_D$ to obtain the output image $\mathbf{I}=\mathbf{I}_D + \mathbf{U}$, We use sigmoid activation functions on the output of $U_{\boldsymbol{\beta}}$ as this aligns with the intuition that the model should only increase the image's brightness.

The model $U_{\boldsymbol{\beta}}$ is trained using the set of light images $\mathcal{D}_N=\{\mathbf{I}^{(n)}\}_{n=1}^N=\{(\mathbf{R}^{(n)}, \mathbf{L}^{(n)})\}_{n=1}^N$. The corresponding darker images are generated using the Dimming module by sampling from the distribution given by Equation \ref{eq:R_prob}. Since we train $U_{\boldsymbol{\beta}}$ in unsupervised mode, we do not have access to illumination $\mathbf{L}_D$. We solve this issue by calculating $\mathbf{L}_D = \mathbf{\Phi} \odot \mathbf{L}$, where components of dimming matrix $\mathbf{\Phi}$ are sampled from normal distribution $\mathcal{N}(\gamma \cdot \mu_{l_{i,j}}, \alpha \cdot \sigma_{l_{i,j}}^2)$, where $l_{i,j}$ represent the element in light illumination $\mathbf{L}$. The values of $\mu_{k}$ and $\sigma_{k}$ are means and deviations for each light value calculated from the few training pairs $\mathcal{D}_K$, where $k \in \{0,\dots, 255\}$. The detailed estimation procedure is provided in Appendix in Section \ref{sec:detailed_architecture}.  In addition, we propose to scale the Gaussian means (Equation \ref{eq:r_prob}), by parameter $\gamma$, where the values are sampled from the uniform distribution, $\gamma \sim \mathcal{U}(a_0,a_1)$. Thanks to this, we achieve diverse levels of darkness for output images. Second, we utilize the \emph{temperature trick} usually applied in generative models \cite{kingma2018glow} by scaling the variances from Equation \ref{eq:r_prob} by $\alpha<1$.


\begin{figure*}[t]
    \centering
    
    \begin{subfigure}{0.19\textwidth}
        \includegraphics[width=\linewidth]{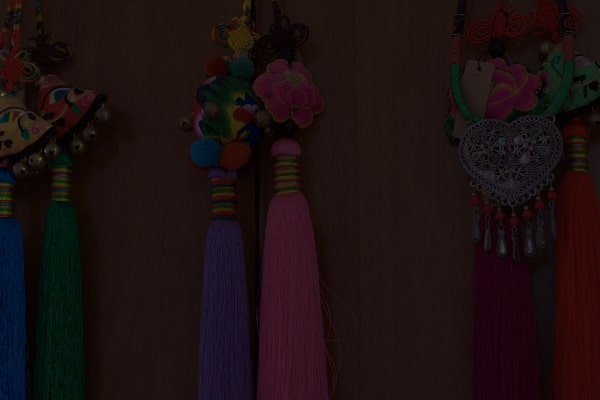}
        \caption*{Input}
    \end{subfigure}
    \begin{subfigure}{0.19\textwidth}
        \includegraphics[width=\linewidth]{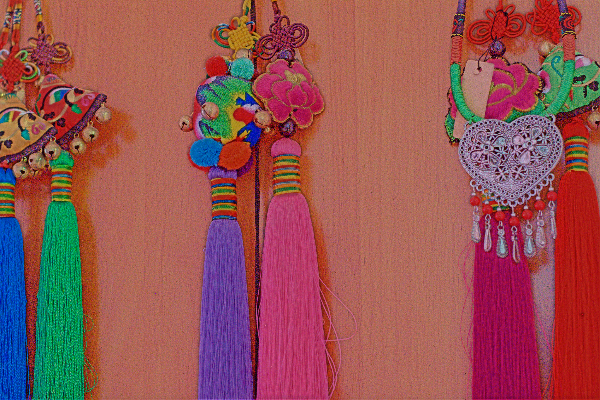}
        \caption*{RetinexNet}
    \end{subfigure}
    \begin{subfigure}{0.19\textwidth}
        \includegraphics[width=\linewidth]{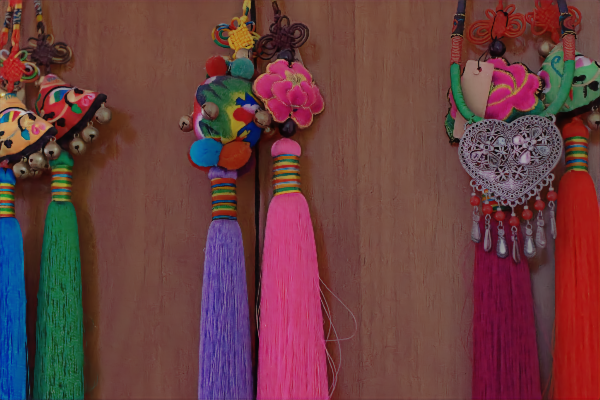}
        \caption*{KinD++}
    \end{subfigure}
    \begin{subfigure}{0.19\textwidth}
        \includegraphics[width=\linewidth]{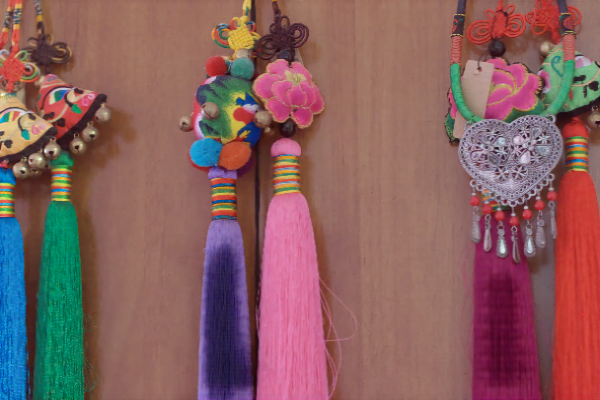}
        \caption*{SNR-Net}
    \end{subfigure}
    \begin{subfigure}{0.19\textwidth}
        \includegraphics[width=\linewidth]{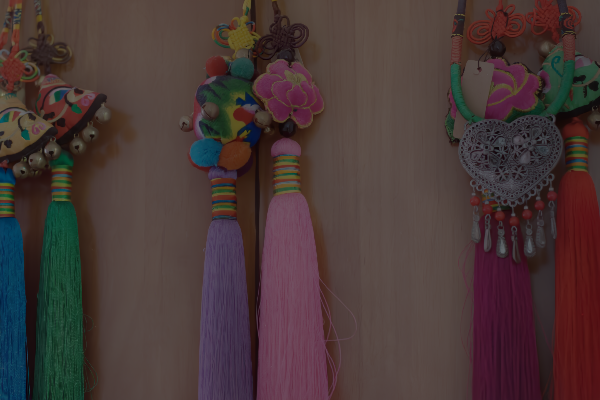}
        \caption*{LLFlow}
    \end{subfigure}
    
    \vspace{0.5em} 
    
    \begin{subfigure}{0.19\textwidth}
        \includegraphics[width=\linewidth]{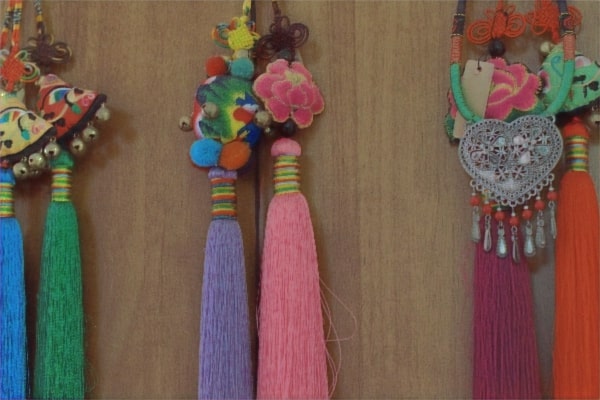}
        \caption*{Dimma 3 pairs}
    \end{subfigure}
    \begin{subfigure}{0.19\textwidth}
        \includegraphics[width=\linewidth]{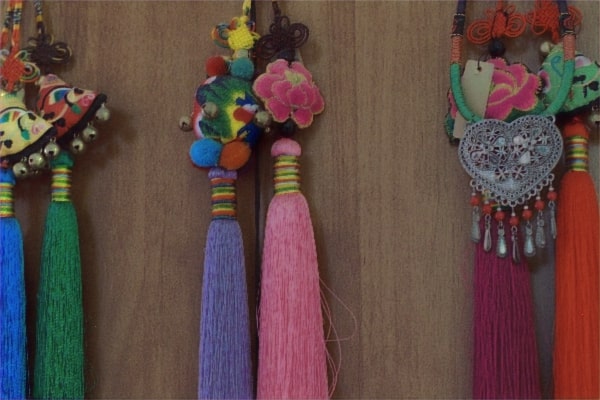}
        \caption*{Dimma 5 pairs}
    \end{subfigure}
    \begin{subfigure}{0.19\textwidth}
        \includegraphics[width=\linewidth]{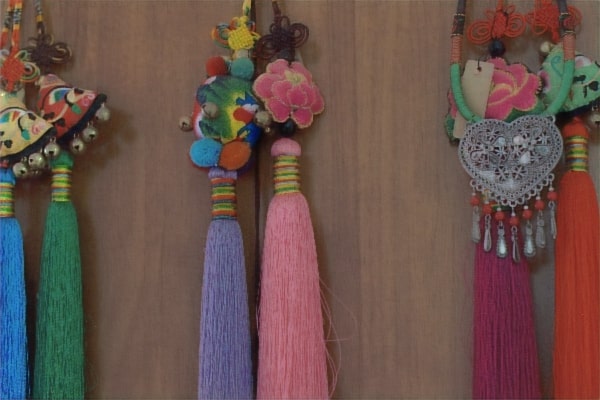}
        \caption*{Dimma 8 pairs}
    \end{subfigure}
    \begin{subfigure}{0.19\textwidth}
        \includegraphics[width=\linewidth]{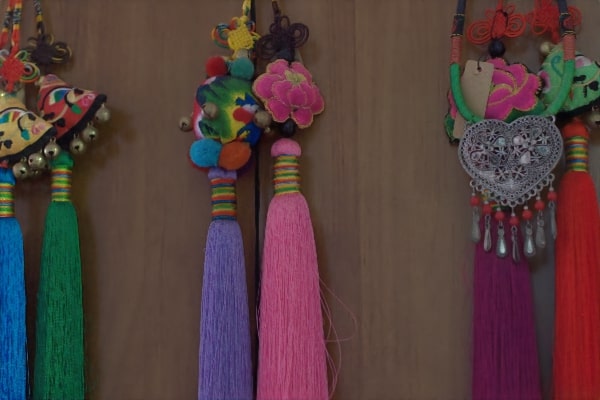}
        \caption*{Dimma full}
    \end{subfigure}
    \begin{subfigure}{0.19\textwidth}
        \includegraphics[width=\linewidth]{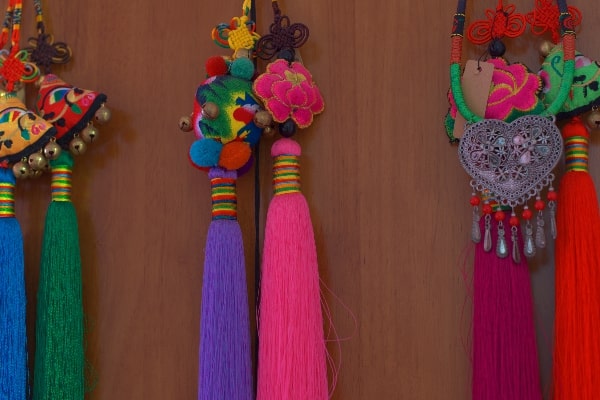}
        \caption*{Ground truth}
    \end{subfigure}
    
    \caption{Visual comparison of Dimma with different fully supervised methods on image pair from LOL dataset.}
    \label{fig:comparison}
\end{figure*}

\begin{table*}[t]
\centering
\begin{tabular}{ccccccc}
Method & PSNR $\uparrow$ & SSIM $\uparrow$ & RGB-SSIM $\uparrow$ & LPIPS $\downarrow$ & DeltaE $\downarrow$ & NIQE $\downarrow$ \\
\hline
SNR-Net & 19.43 & 0.78 & 0.75 & 0.42 & 9.59 & 4.61 \\
LLFlow & 19.46 & 0.81 & 0.79 & 0.35 & 9.69 & 3.50 \\
Dimma (ours) & \textbf{24.14} & \textbf{0.83} & \textbf{0.81} & \textbf{0.27} & \textbf{6.14} & \textbf{2.93} \\
\end{tabular}
\caption{Quantitive comparison on FS-Dark. All of the models were trained on FS-Dark with the same setup as for LOL (except for the smaller batch size due to the dataset size).}
\label{tab:results_our_ds}
\end{table*}

The light image $\mathbf{I}^{(n)}$ together with generated dark image $\mathbf{\tilde{I}}_{D}^{(n)}$ sampled from MDN representing the dimming module is further used as one of the training pairs to estimate the parameters $\boldsymbol{\beta}$ of the brightening component. The model $U_{\boldsymbol{\beta}}$ requires generated dark image $\mathbf{\tilde{I}}_{D}^{(n)}$, the corresponding histogram equalization $H(\mathbf{\tilde{I}}_{D}^{(n)})$, color map $\mathbf{\tilde{R}}_{D}^{(n)}$, and illumination as input $\mathbf{\tilde{L}}_{D}^{(n)}$. In addition, the parameter $\Delta m^{(n)}$ that controls the light level should be delivered to the model. The $\Delta m^{(n)}$ is calculated as the difference between the mean lightnesses of the original light image $\mathbf{I}^{(n)}$ and generated dark image $\mathbf{\tilde{I}}_{D}^{(n)}$. The total loss used to train the model $U_{\boldsymbol{\beta}}$ is represented by the sum of mean squared error $\mathcal{L}_{MSE}$ and perceptual loss $\mathcal{L}_{P}$:

\begin{align}
\mathcal{L}_{MSE} + \mathcal{L}_{P}=
&\sum_{n=1}^N ||\mathbf{I}^{(n)} - (\mathbf{\tilde{I}}_{D}^{(n)} + \mathbf{U}^{(n)})||_2^2 + \nonumber \\
&\sum_{n=1}^N \lambda||f(\mathbf{I}^{(n)}) -  f(\mathbf{\tilde{I}}_{D}^{(n)} + \mathbf{U}^{(n)})||_2^2
\end{align}
where $\mathbf{U}^{(n)}=U_{\boldsymbol{\beta}}(\mathbf{\tilde{I}}_{D}^{(n)},H(\mathbf{\tilde{I}}_{D}^{(n)}), \mathbf{\tilde{R}}_{D}^{(n)}, \mathbf{\tilde{L}}_{D}^{(n)}, \Delta m^{(n)})$ is residual map returned by UNet for a given dark image $\mathbf{\tilde{I}}_{D}^{(n)}$, $\lambda$ is hyperparameter scaling perceptual loss and $f(\cdot)$ is the non-trainable feature extractor for perceptual loss. Note that $\mathbf{\tilde{I}}_{D}^{(n)}$ represents a different sample from the dimming module for each of the epochs. 

\section{Experiments}

We conducted a series of experiments aiming to demonstrate the superiority of Dimma over previous methods: standard quantitative and qualitative comparison, generalizability test, training with limited data, and light level conditioning. Our model was compared with RetinexNet \cite{retinex_net}, KinD \cite{kind}, Zero-DCE \cite{zero-dce}, EnlightenGAN \cite{enlightengan}, RUAS \cite{ruas}, KinD++ \cite{kindpp}, SNR-Net \cite{xu2022snr}, LLFlow \cite{llflow}, and Retinexformer \cite{cai2023retinexformer}. For the evaluation of results, we employed PSNR, SSIM \cite{ssim} on both grayscale and RGB, LPIPS \cite{lpips, lpips2}, DeltaE and NIQE \cite{niqe} metrics. Finally, we empirically justified our design choices regarding dimming module and final finetuning of Dimma.

\paragraph{Quantitative}
The first experiment involved a quantitative comparison on the LOL dataset \cite{retinex_net}.
In the first phase of semi-supervised training, we used 3, 5, 8, or 480 real image pairs from LOL dataset to train the dimming module. 
For the second phase (see Figure \ref{fig:big_picture}) we introduced MixHQ dataset, which contained more than 15000 natural images from COCO2017 \cite{coco}, ImageNet \cite{imagenet}, Clic \cite{clic}, Inter4K \cite{inter4k} and light images from train split of LOL dataset. We selected good-quality images only, using the rules described in Appendix. Those were used as unlabeled samples for the initial (unsupervised) training of the brightening module. For validation, we used 5 pairs from LOL. 
After the unsupervised training, we finetuned the UNet on the same pairs which had been used to construct dimming module, and tested it on 15 pairs from LOL test split.

As depicted in Table \ref{tab:results_lol}, our semi-supervised models trained on few real image pairs achieved results comparable to existing supervised methods and outperformed all unsupervised methods. Dimma trained on 480 pairs achieves state-of-the-art performance in terms of PSNR, LIPIS, DeltaE, and NIQE metrics and competitive results compared to other models in SSIM. High PSNR and DeltaE indicate that our model exhibits exceptional accuracy in terms of lighting and colors while high LPIPS and NIQE metrics suggest great realism of generated images. For visual comparison see Figure \ref{fig:comparison} and other Figures in Appendix.

\paragraph{Generalizability}
We proceeded with a cross-dataset comparison on the VE-LOL dataset \cite{ve-lol}, focusing exclusively on models trained on the LOL dataset and unsupervised methods. This particular experiment aimed to evaluate the ability of our method to generalize to unseen image source. We used 100 real and 100 synthetic pairs from VE-LOL as a test set and evaluate the same methods mentioned in the previous paragraph. Results depicted in Table \ref{tab:results_lol} show the superiority of our model in terms of PSNR, LPIPS, and DeltaE metrics in cross-dataset assessments.

\paragraph{Limited data}
To examine the usefulness of our framework in real-case scenarios, when we have only a few real image pairs, we collected our own dataset called FewShot-Dark (FS-Dark). The collecting procedure was similar to LOL \cite{retinex_net} - we took two photos of the same scene with different camera light exposure. FS-Dark consists a total of 14 image pairs (6 training, 4 validation, and 4 testing) taken by a Samsung Galaxy M52 mobile phone. For a fair comparison, we trained our Dimma model, LLFlow \cite{llflow} and SNR-Net \cite{xu2022snr}.

The results are shown in Table \ref{tab:results_our_ds} and a visual comparison is placed in Appendix. Our method achieves better results in terms of all metrics, which proves Dimma strongly outperforms other methods when only a few ground truth pairs are available for training.

\begin{table}
\centering
\begin{tabular}{cccc}
\#Pairs              & 3                     & 5                     & 8                     \\ \hline 
                      \multicolumn{4}{c}{Supervised (w/o dimming module)}                                        \\ \hline 

PSNR                  & 21.05 ± 4.16          & 21.62 ± 4.19          & 22.18 ± 4.40          \\
SSIM                  & 0.81 ± 0.09           & 0.82 ± 0.09           & 0.83 ± 0.09           \\
LPIPS                 & 0.27 ± 0.09           & \textbf{0.25 ± 0.09}  & \textbf{0.23 ± 0.08}  \\ \hline
                      \multicolumn{4}{c}{Semi-supervised with deterministic dimming}           \\ \hline 
PSNR                  & 22.93 ± 2.31          & 24.07 ± 2.41          & 24.26 ± 2.40          \\
SSIM                  & 0.78 ± 0.06           & 0.78 ± 0.06           & 0.79 ± 0.06           \\
LPIPS                 & 0.32 ± 0.08           & 0.31 ± 0.08           & 0.30 ± 0.08           \\ \hline 
\multicolumn{4}{c}{Semi-supervised with MDN}                                               \\ \hline 
PSNR                  & 23.24 ± 2.25          & 23.83 ± 2.26          & 24.47 ± 2.22          \\
SSIM                  & 0.78 ± 0.06           & 0.79 ± 0.06           & 0.79 ± 0.05           \\
LPIPS                 & 0.31 ± 0.08           & 0.30 ± 0.08           & 0.29 ± 0.07           \\  \hline
\multicolumn{4}{c}{Semi-supervised with MDN + finetuning (Dimma)}                                    \\ \hline 
PSNR                  & \textbf{23.54 ± 2.79} & \textbf{24.50 ± 2.53} & \textbf{24.70 ± 2.59} \\
SSIM                  & \textbf{0.83 ± 0.06}  & \textbf{0.84 ± 0.06}  & \textbf{0.86 ± 0.06}  \\
LPIPS                 & \textbf{0.26 ± 0.07}  & \textbf{0.25 ± 0.06}  & \textbf{0.23 ± 0.06} 
\\ 
\end{tabular}
\caption{Ablation study on LOL dataset. Each experiment was conducted five times with different training samples. The files which were used for training are listed in Table \ref{tab:few_shot_files} in Appendix.}
\label{tab:fewshot_results}
\end{table}

\paragraph{Light conditioning}
To compare our light conditioning mechanism used in Dimma, we used SICE \cite{sice} dataset consisting of image sequences with different lightness. Since most low-light image enhancement methods (e.g. \cite{sid, kind, kindpp, llflow}) are capable of lightness conditioning, we created the visual comparison of generated images with different conditioning values. Thanks to the lightning conditioning mechanism of UNet, we achieve remarkable precision in generating images with a certain level of lightning. The qualitative results, placed in Appendix, emphasize the superiority of Dimma in terms of light conditioning.

\paragraph{Ablation study}
To further establish the efficacy of our approach, we compared the results of semi-supervised training using Dimma with training UNet without the dimming module. The intention behind these experiments was to validate the rationality of our dimming module with MDN with an extremely small dataset. We also showed that using a deterministic convolutional network with the same architecture as the backbone of MDN does not give as appealing results as our MDN-based dimming module.

Our ablation study examines the usefulness of the learnable dimming module. We compare it to a more naive approach of training only the UNet on the very same few pairs. Table \ref{tab:fewshot_results} shows the rationality of the Dimma pipeline.

\section{Conclusion}
In this paper, we propose a novel semi-supervised learning approach for low-light image enhancement. We address the problem of insufficient large datasets containing image pairs for various cameras by utilizing only a few real image pairs, which can be easily captured and prepared. Based on these pairs, our model aims to mimic the characteristics of a specific camera in low-light conditions. With this approach, we can effectively train existing models to restore dark images captured by that particular camera. Furthermore, our approach allows for the grading of the dimming factor during training, enabling the model to adapt to different light conditions and generate images with any desired illumination. As a result, our method significantly outperforms all existing unsupervised models and achieves competitive results compared to fully supervised approaches.

\paragraph{Limitations} 
The main limitation of Dimma is the necessity to train the UNet backbone using a dimming module calibrated for particular camera settings. As a potential future work, having a more general backbone with changing camera-specific modules on top would enable few-shot fine-tuning, limiting the time and resources needed.

Our method achieves good results using only a couple of paired ground truth images, making the data-gathering process easier and cheaper but still necessary. The development of fully unsupervised methods that are on par with supervised models is the next crucial step in low-light image enhancement research.

\bibliography{bibliography}

\newpage
\appendix

\section{Detailed architecture}\label{sec:detailed_architecture}
\paragraph{Dimming module}
Our dimming module utilizes the Retinex Decomposition to deal separately with reflectance and illumination. To manipulate the reflection, we use Mixture Density Network (MDN) to be able to align well with the color distortion distribution present in the dark images. We set the number of mixture components to 4 and each pixel can have mean and standard deviation sampled from different components. This approach together with using only convolution with 1x1 filters allows us to treat each pixel as an individual sample, preventing quick overfitting with an insufficient number of pairs. We train the MDN model for 1000 epochs with a learning rate set to $0.01$ for each training.

Since illumination has only one value per pixel, it is less complicated and we can model its dynamic without parameterized models. We compute dimmed illumination from the formula:
\begin{equation}
\mathbf{L}_D = \mathbf{\Phi} \odot \mathbf{L}
\end{equation}
where elements of $\mathbf{\Phi}$ are sampled from the Gaussian distribution:
\begin{equation}
\phi_{i,j} = \mathcal{N}(\gamma \cdot \mu_{l_{i,j}}, \alpha \cdot \sigma_{l_{i,j}})
\end{equation}
where $\gamma$ denotes dimming factor, $\alpha$ is a noise temperature. $\mu_{l_{i,j}}$ and $\sigma_{l_{i,j}}$ are means and standard deviations of lightness ratios for each illumination value calculated from the few training pairs:
\begin{align}
\mu_{k} &= \frac{1}{\sum_{n,i,j} \mathbb{I}(l_{n,i,j} = k)} \sum_{n,i,j} \frac{l_{D,n,i,j} \mathbb{I}(l_{n,i,j} = k)}{l_{n,i,j}} \\
\sigma_{k} &= \sqrt{\frac{\sum_{n,i,j} \left(\frac{l_{D,n,i,j} \mathbb{I}(l_{n,i,j} = k)}{l_{n,i,j}} - \mu_k \right)^2}{\sum_{n,i,j} \mathbb{I}(l_{n,i,j} = k) - 1}}
\end{align}
where $k \in (0, 255)$ are all possible light values, $l_{n,i,j}$ is a illumination value from an n-th light image at $(i,j)$ pixel location and $l_{D,n,i,j}$ is the illumination value of the corresponding pixel from a dark image. Operator $\mathbb{I}$ returns 1 if arguments are true and 0 otherwise. In the rare case when $\sum_{n,i,j} \mathbb{I}(l_{n,i,j} = k)$ is 0 or 1, we interpolate values from the closest known lightness values. 

\paragraph{Conditioned UNet}
Our UNet model consists of three upsample, downsample, and four normal layers. Each layer has two resnet blocks \cite{diffusers}. We apply an attention mechanism with 64 heads for feature maps with the smallest resolution. We condition UNet layers with a mean lightness difference between the original image and the image we want to produce. This difference is projected with sine and cosine functions, transformed with an MLP, and injected into the UNet feature map during its forward pass.

\paragraph{Training settings}
For the semi-supervised approach, we utilized 3, 5, 8, or 480 pairs from LOL. Each training was repeated five times (three for 480 pairs) using different training sets. We trained the UNet on random crops of size 256$\times$256, with a batch size of 4. The initial learning rate was set to $10^{-5}$, and we employed a cosine annealing scheduler for 5k iterations, with early stopping based on the validation set. The optimization was performed using the Adam optimizer. 

Dimming factor $\gamma \sim \mathcal{U}(a_0,a_1)$ where $a_0 = 0.3$ and $a_1 = 2$. $\gamma < 1$ produces slightly darker images than those from the training set and $\gamma > 1$ creates lighter images. This way model can be trained on a wider range of light conditions.

Finally, we fine-tuned the UNet on real images for 2k iterations. However, early stopping was usually triggered after a few hundred updates due to overfitting on the small training set. While fine-tuning on the entire training set, we set the number of iterations to 150k to fully utilize the potential of the large dataset. The rest of the hyperparameters remained constant.

\section{Datasets}
\paragraph{MixHQ}
To train UNet in an unsupervised way, we combine five real image datasets to give the model knowledge about the distribution of good-quality photos. During the training process, we dim them according to the Dimma paradigm. The unsupervised training proved to increase the quality of generated images especially if we do not have too many real image pairs. To provide only the best quality images, we select them based on the following conditions:
\begin{itemize}
    \item COCO2017 unlabeled - images with at least 500x500 resolution
    \item Clic - images reduced to 512x512 size
    \item ImageNet1K - images from validation and test sets with resolution between 550x550 and 1100x1100. We discarded images with white backgrounds.
    \item Inter4K - The first and the last frame of each clip resized by a factor of 0.3.
    \item LOL - original 485 training split. Note that we only used light images without knowledge about the dark image distribution.
\end{itemize}

\begin{table*}
\centering
\begin{tabular}{cccccccccc}

Exp. & \#pairs & File 1 & File 2 & File 3 & File 4 & File 5 & File 6 & File 7 & File 8 \\ \hline
  & 3 & 2.png & 5.png & 6.png & & & & & \\
1 & 5 & 2.png & 5.png & 6.png & 9.png & 10.png & & & \\
  & 8 & 2.png & 5.png & 6.png & 9.png & 10.png & 12.png & 13.png & 14.png\\ \hline
  & 3 & 17.png & 18.png & 21.png & & & & & \\
2 & 5 & 17.png & 18.png & 21.png & 24.png & 25.png & & & \\
  & 8 & 17.png & 18.png & 21.png & 24.png & 25.png & 26.png & 27.png & 28.png\\ \hline
  & 3 & 36.png & 38.png & 39.png & & & & & \\
3 & 5 & 36.png & 38.png & 39.png & 40.png & 42.png & & & \\
  & 8 & 36.png & 38.png & 39.png & 40.png & 42.png & 43.png & 44.png & 46.png\\ \hline
  & 3 & 50.png & 51.png & 52.png & & & & & \\
4 & 5 & 50.png & 51.png & 52.png & 53.png & 54.png & & & \\
  & 8 & 50.png & 51.png & 52.png & 53.png & 54.png & 56.png & 57.png & 58.png\\ \hline
  & 3 & 61.png & 62.png & 63.png & & & & & \\
5 & 5 & 61.png & 62.png & 63.png & 64.png & 67.png & & & \\
  & 8 & 61.png & 62.png & 63.png & 64.png & 67.png & 68.png & 69.png & 70.png \\ 
\end{tabular}
\caption{Files from LOL dataset \cite{retinex_net} which are used for semi-supervised training. We train $15$ MDN models and $15$ UNets to compare the effectiveness of our method. Results are shown in Table \ref{tab:fewshot_results}.}
\label{tab:few_shot_files}
\end{table*}

\paragraph{Semi-supervised training}
For semi-supervised learning experiments, we use different subsets of the LOL training set. The filenames used in each experiment are shown in Table \ref{tab:few_shot_files}. Files from this table were used for MDN training, illumination dynamic estimation, and finetuning the UNet model after training it on MixHQ in an unsupervised manner.

\section{Ablation studies}

\begin{figure*}[b]
    \centering
    \subfloat[\centering light image]{{\includegraphics[width=0.3\textwidth]{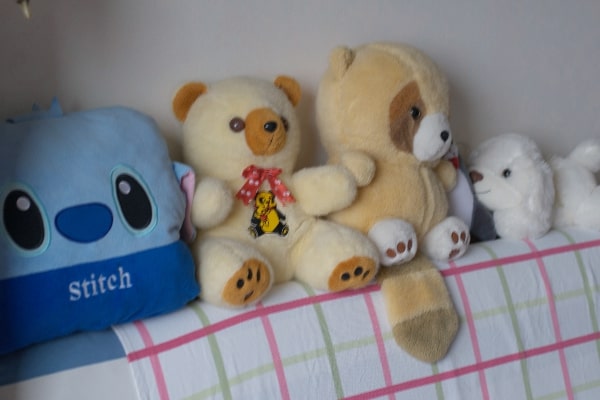} }}
    \subfloat[\centering dimmed image]{{\includegraphics[width=0.3\textwidth]{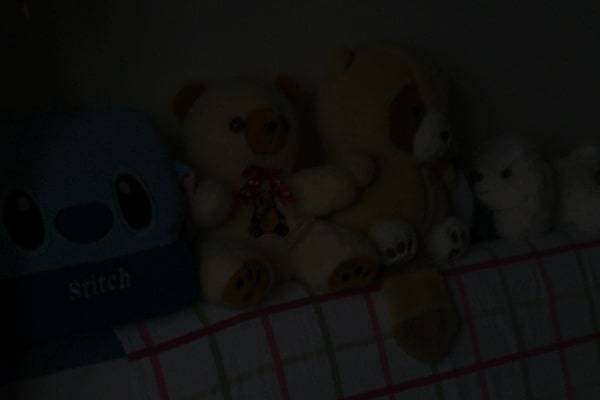} }}
    \subfloat[\centering original dark image]{{\includegraphics[width=0.3\textwidth]{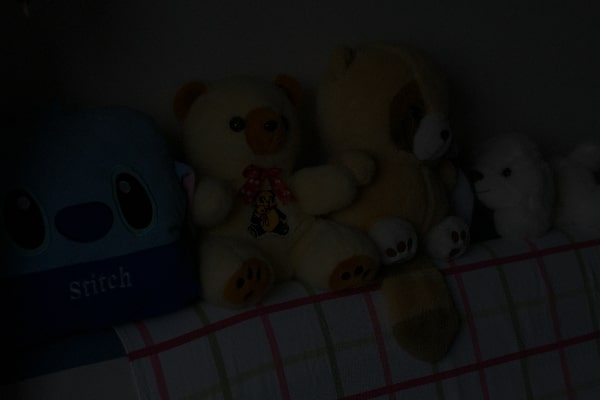} }}
    
    \subfloat[\centering light image reflectance]{{\includegraphics[width=0.3\textwidth]{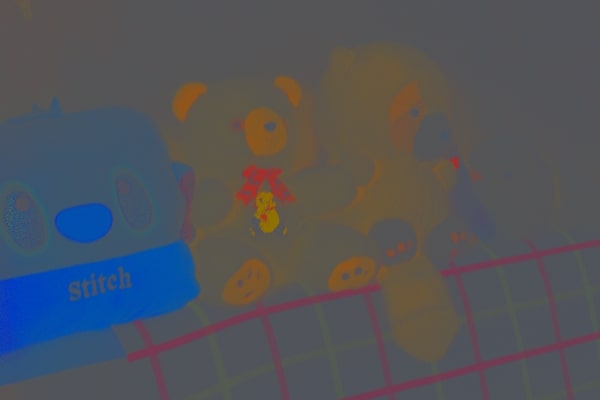} }}
    \subfloat[\centering dimmed image reflectance]{{\includegraphics[width=0.3\textwidth]{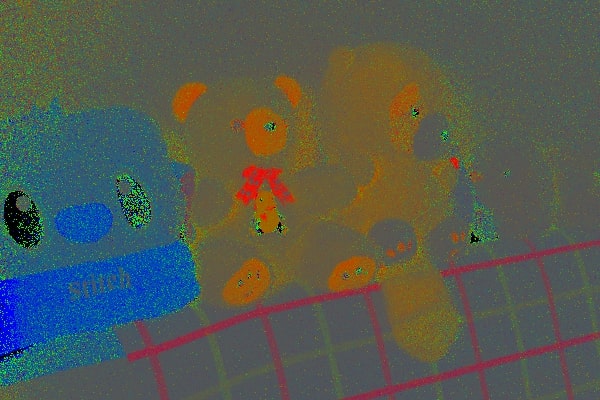} }}
    \subfloat[\centering dark image reflectance]{{\includegraphics[width=0.3\textwidth]{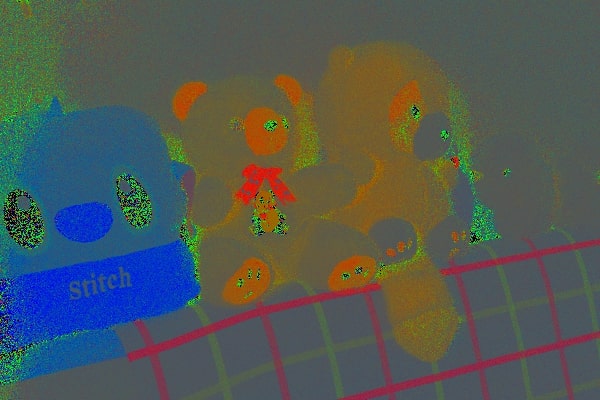} }}

    \subfloat[\centering distribution of color changes for light pixel]{{\includegraphics[width=0.45\textwidth]{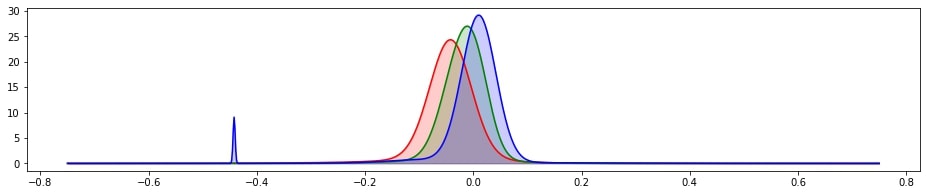} }}
    \subfloat[\centering distribution of color changes for dark pixel]{{\includegraphics[width=0.45\textwidth]{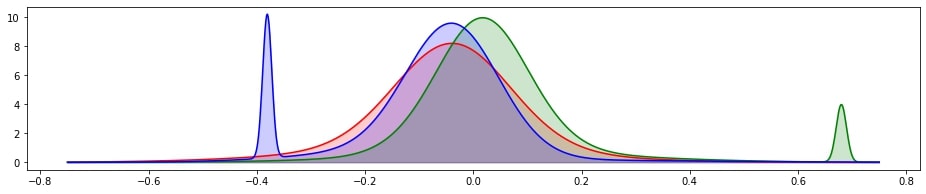} }}
    
    \caption{Output of Mixture Density Network trained on 8 image pairs. Figures (a)-(f) show how the dimming module aligns with real image pairs. Figures (g) and (h) illustrate MDN's output for gray pixels with different illumination differences. (g) from 0.2 to 0.15 (h) from 0.2 to 0.05. The characteristic of images from the LOL dataset is that green noise appears in darker regions which is visible in the (h) distribution. The second observation is that the standard deviation is higher for dark images making them more noisy in general. We can see that in the image (e).}
    \label{fig:mdn_results}
\end{figure*}

To examine the rationality of our approach, we conducted a series of experiments comparing the supervised approach, standard Dimma method and Dimma with deterministic convolutional network as a dimming model. We utilized the same architecture of UNet and the training sets mentioned in Table \ref{tab:few_shot_files} for all three approaches. The results, presented in Table \ref{tab:fewshot_results}, provide insights into the performance of each method. Our Dimma method consistently outperforms any other approach, especially when the available real image pairs are limited.

Moreover, we aimed to demonstrate the advantages of incorporating a generative model, specifically the Mixture Density Network (MDN), in the dimming procedure. Figure \ref{fig:mdn_results} visualizes the output of our dimming module, showcasing its ability to produce noise that closely resembles the real noise found in dark images. By leveraging a distribution-based approach, the MDN generates varying degrees of noise, capturing the nuances of dim photos. Table \ref{tab:fewshot_results} provides further evidence supporting the superiority of the MDN over the deterministic network. In almost all cases, the results obtained using the MDN consistently surpass or at least match the performance of the deterministic network. 

Overall, our experiment demonstrates that the Dimma method, utilizing a generative model like the MDN, yields superior results compared to both standard supervised training and deterministic approaches.

\section{Visualizations}
In this section, we present visual results of our method on different datasets as well as visual comparisons with previous methodologies.

\begin{figure*}
    \centering
    \includegraphics[width=0.19\textwidth]{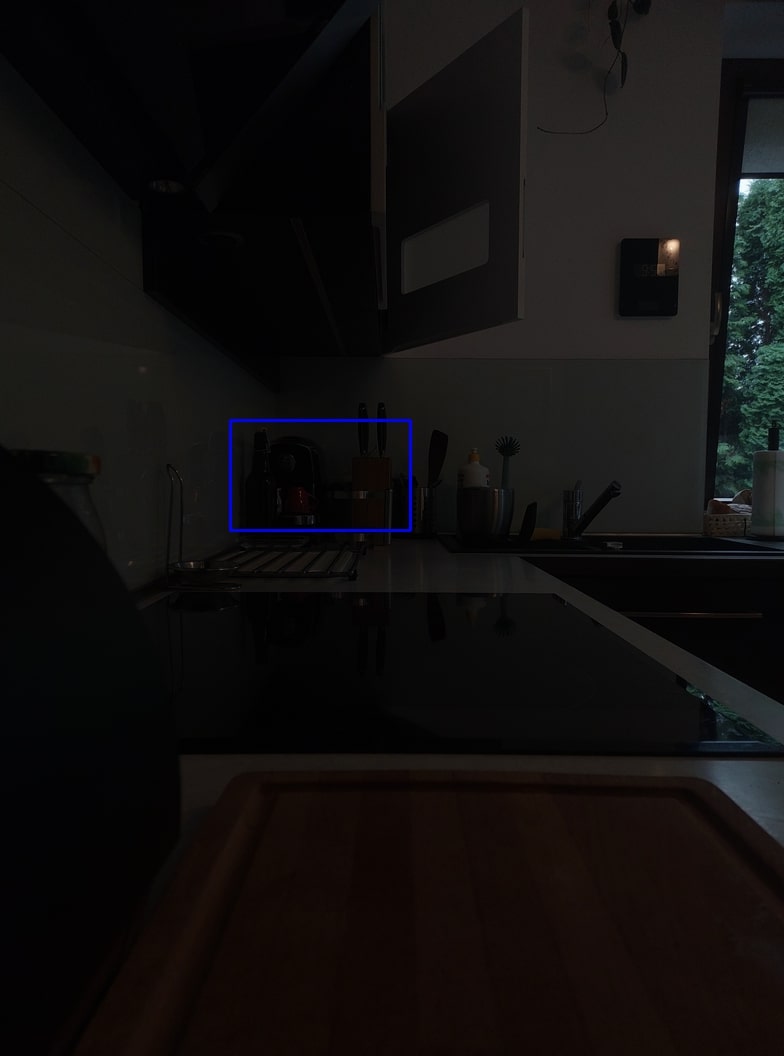}
    \includegraphics[width=0.19\textwidth]{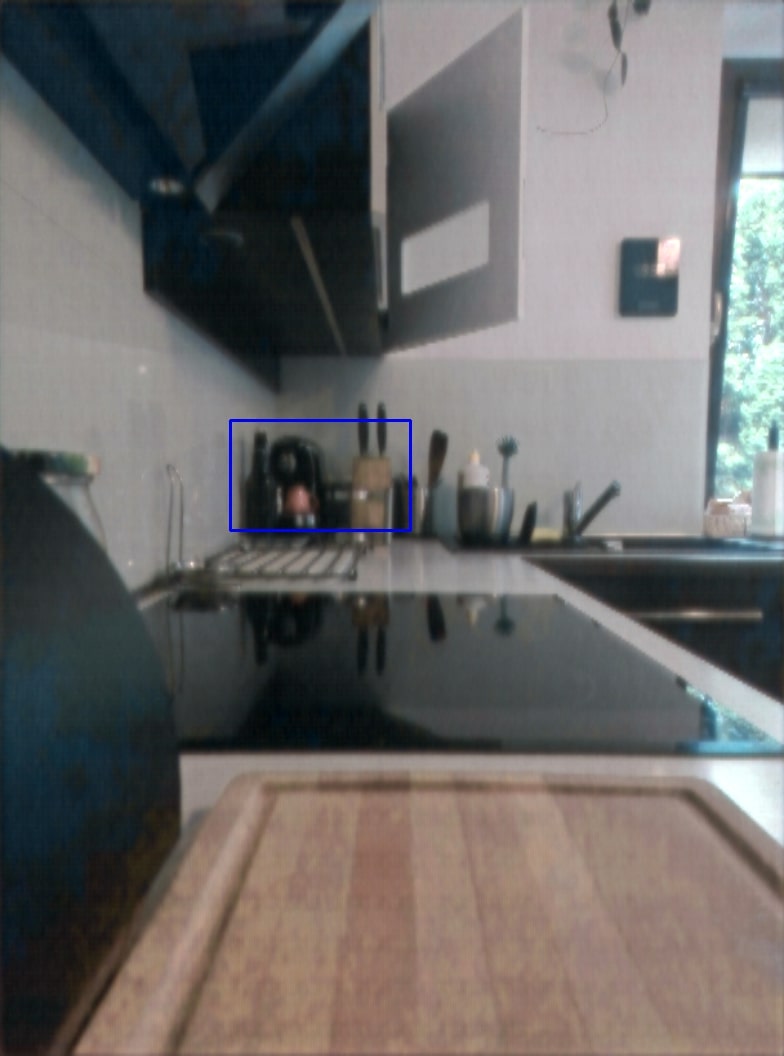}
    \includegraphics[width=0.19\textwidth]{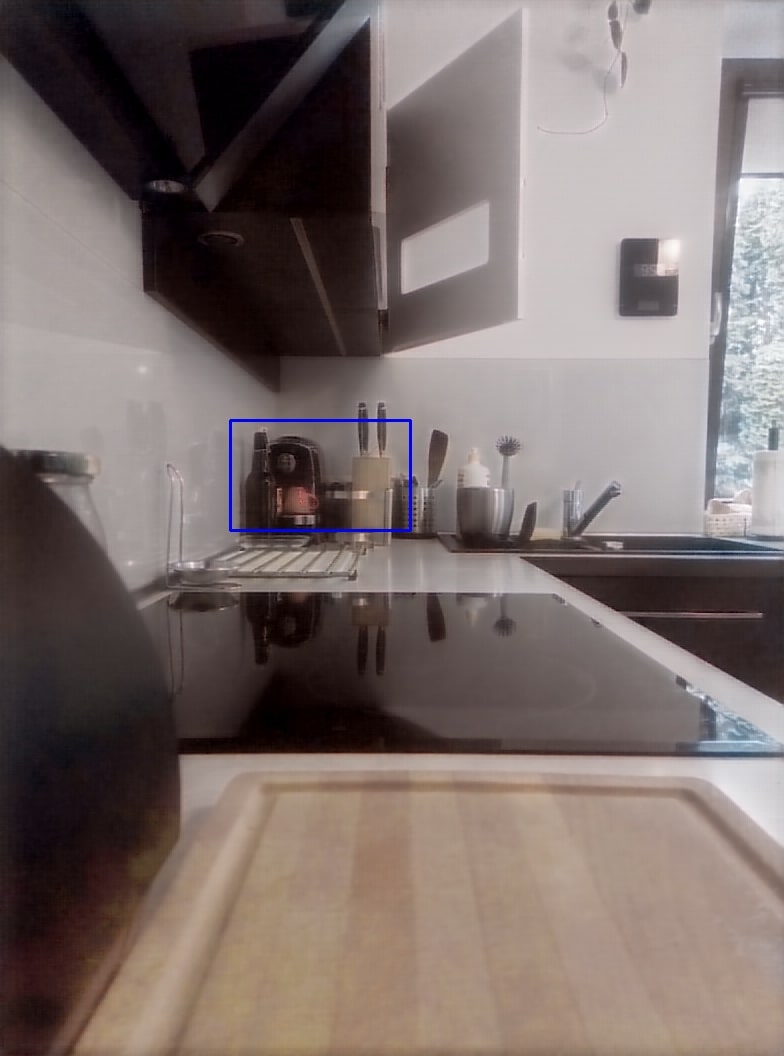}
    \includegraphics[width=0.19\textwidth]{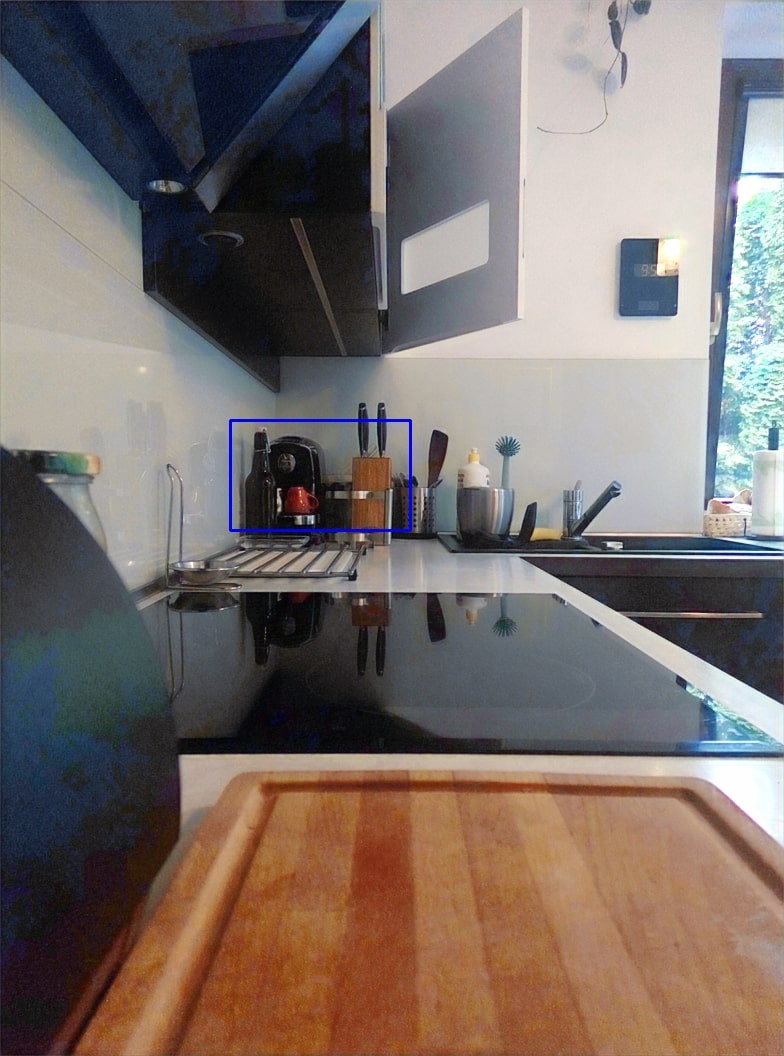}
    \includegraphics[width=0.19\textwidth]{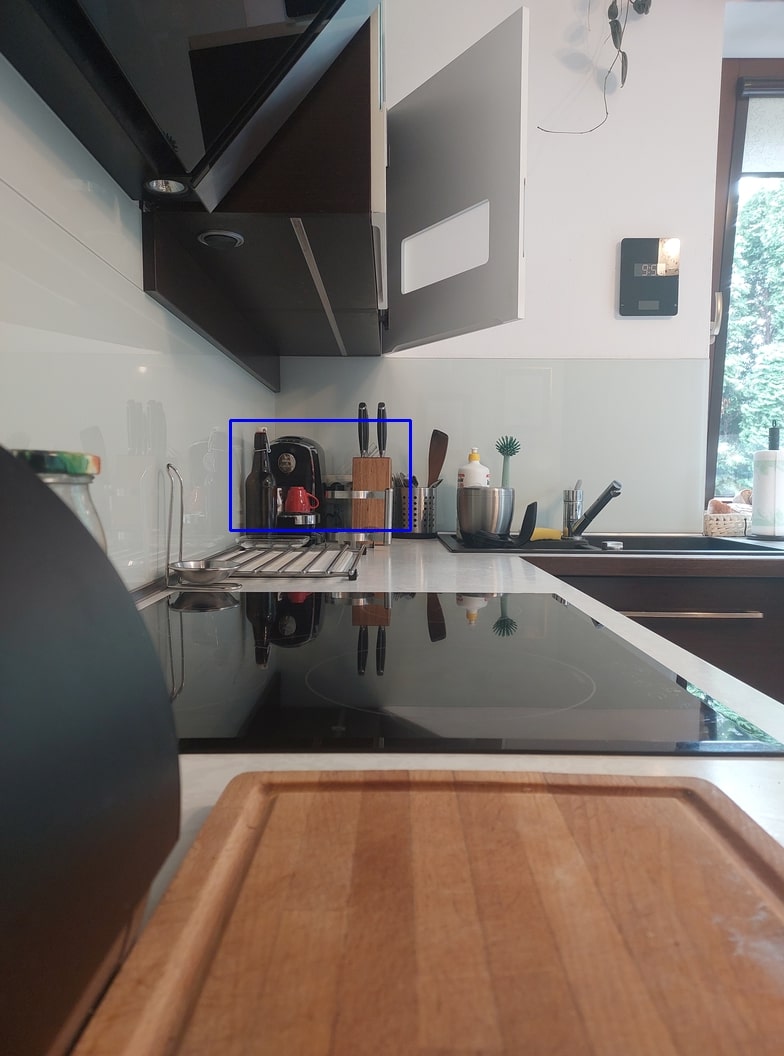}
    
    \begin{subfigure}{0.19\textwidth}
        \includegraphics[width=\textwidth]{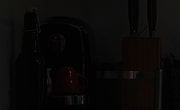}
        \caption*{Input}
    \end{subfigure}
    \begin{subfigure}{0.19\textwidth}
        \includegraphics[width=\textwidth]{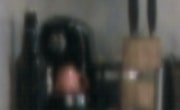}
        \caption*{SNR-Net}
    \end{subfigure}
    \begin{subfigure}{0.19\textwidth}
        \includegraphics[width=\textwidth]{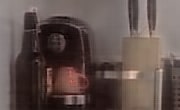}
        \caption*{LLFlow}
    \end{subfigure}
    \begin{subfigure}{0.19\textwidth}
        \includegraphics[width=\textwidth]{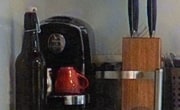}
        \caption*{Dimma}
    \end{subfigure}
    \begin{subfigure}{0.19\textwidth}
        \includegraphics[width=\textwidth]{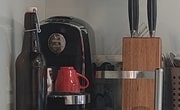}
        \caption*{Ground truth}
    \end{subfigure}
    
    \includegraphics[width=0.19\textwidth]{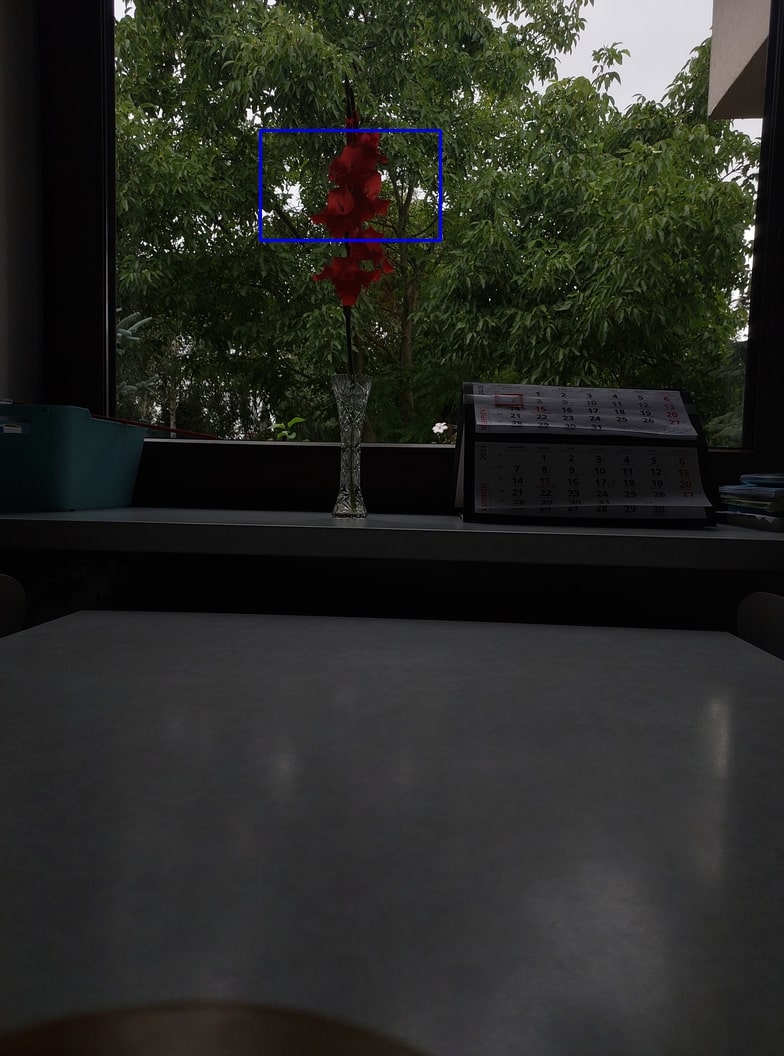}
    \includegraphics[width=0.19\textwidth]{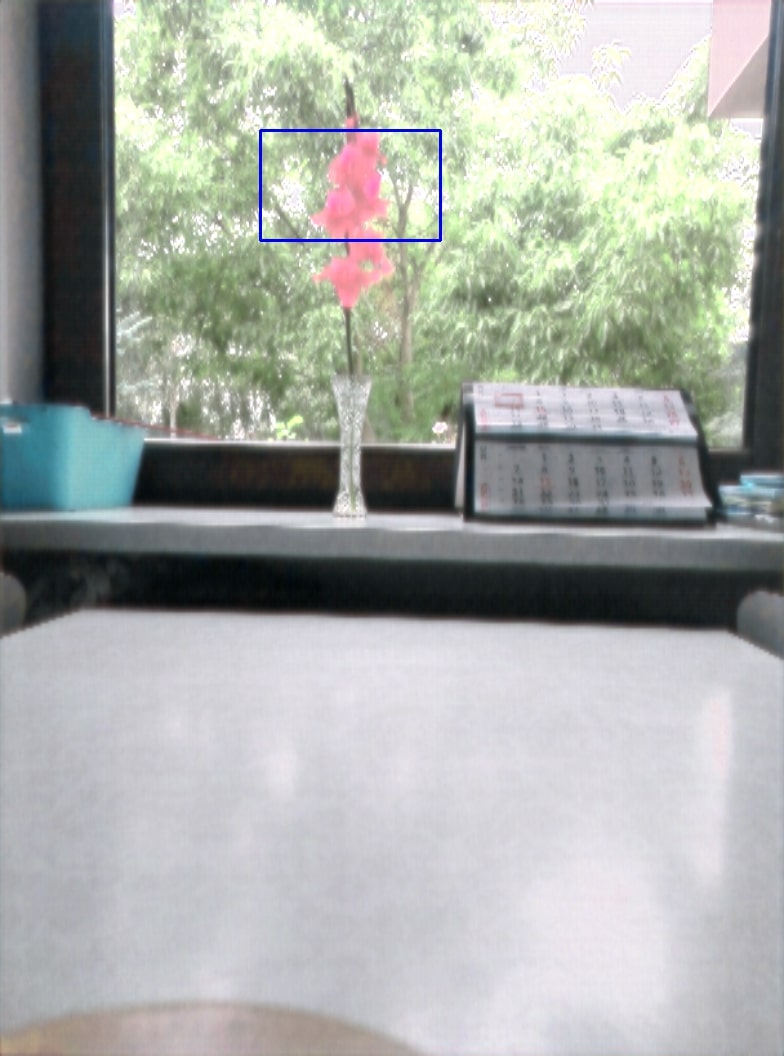}
    \includegraphics[width=0.19\textwidth]{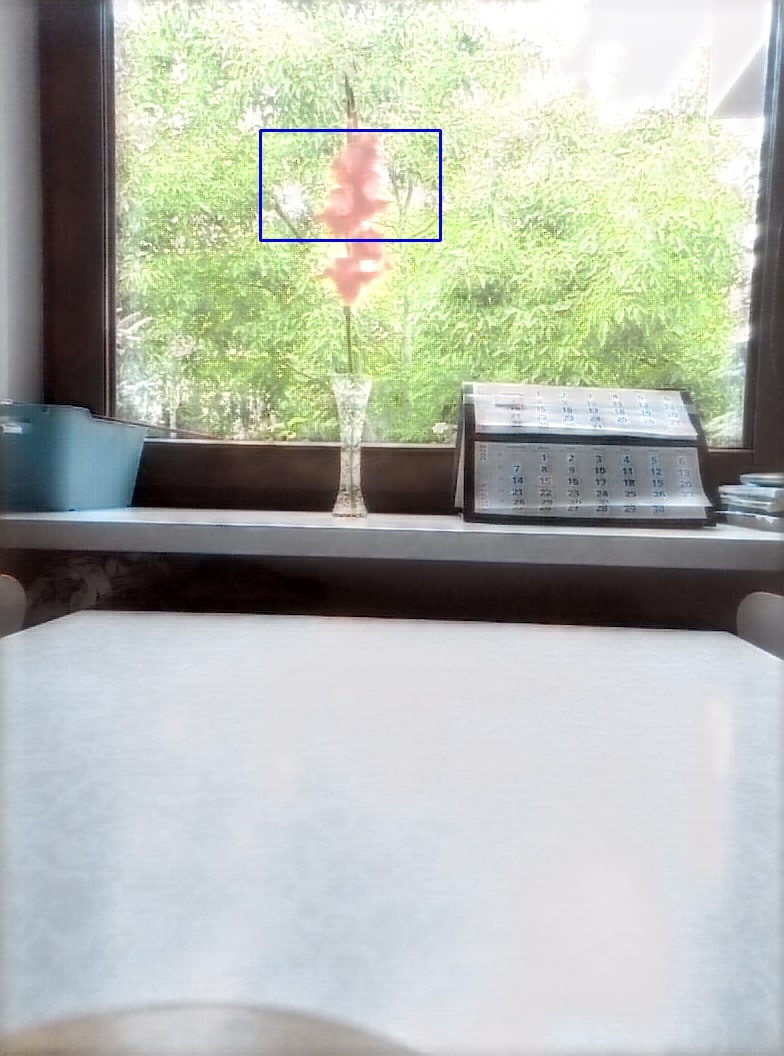}
    \includegraphics[width=0.19\textwidth]{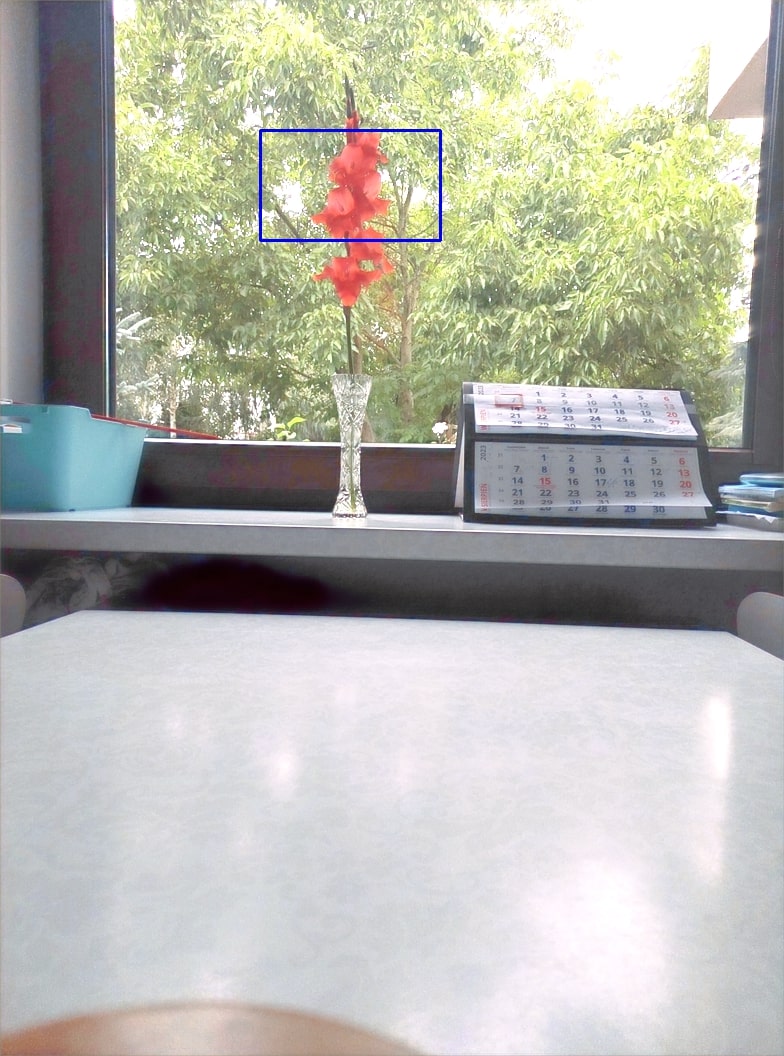}
    \includegraphics[width=0.19\textwidth]{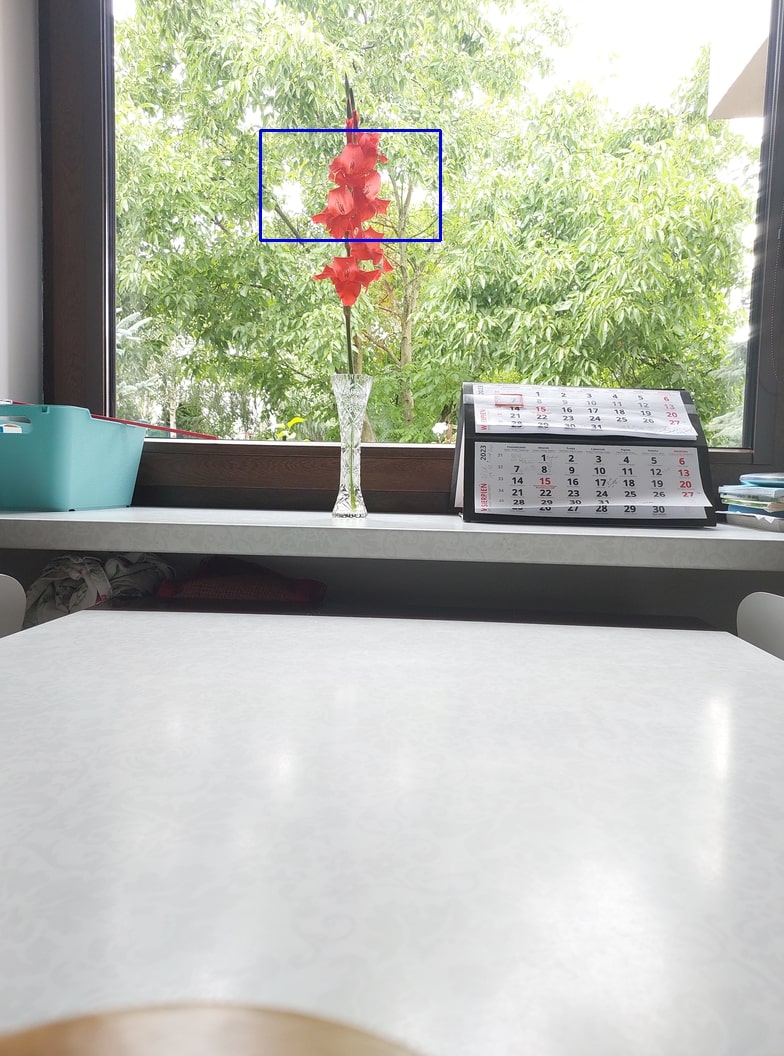}
    
    \begin{subfigure}{0.19\textwidth}
        \includegraphics[width=\textwidth]{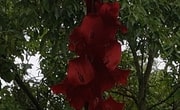}
        \caption*{Input}
    \end{subfigure}
    \begin{subfigure}{0.19\textwidth}
        \includegraphics[width=\textwidth]{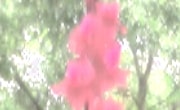}
        \caption*{SNR-Net}
    \end{subfigure}
    \begin{subfigure}{0.19\textwidth}
        \includegraphics[width=\textwidth]{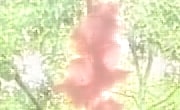}
        \caption*{LLFlow}
    \end{subfigure}
    \begin{subfigure}{0.19\textwidth}
        \includegraphics[width=\textwidth]{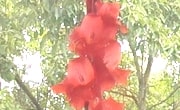}
        \caption*{Dimma}
    \end{subfigure}
    \begin{subfigure}{0.19\textwidth}
        \includegraphics[width=\textwidth]{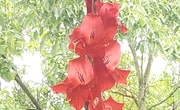}
        \caption*{Ground truth}
    \end{subfigure}
    
    \includegraphics[width=0.19\textwidth]{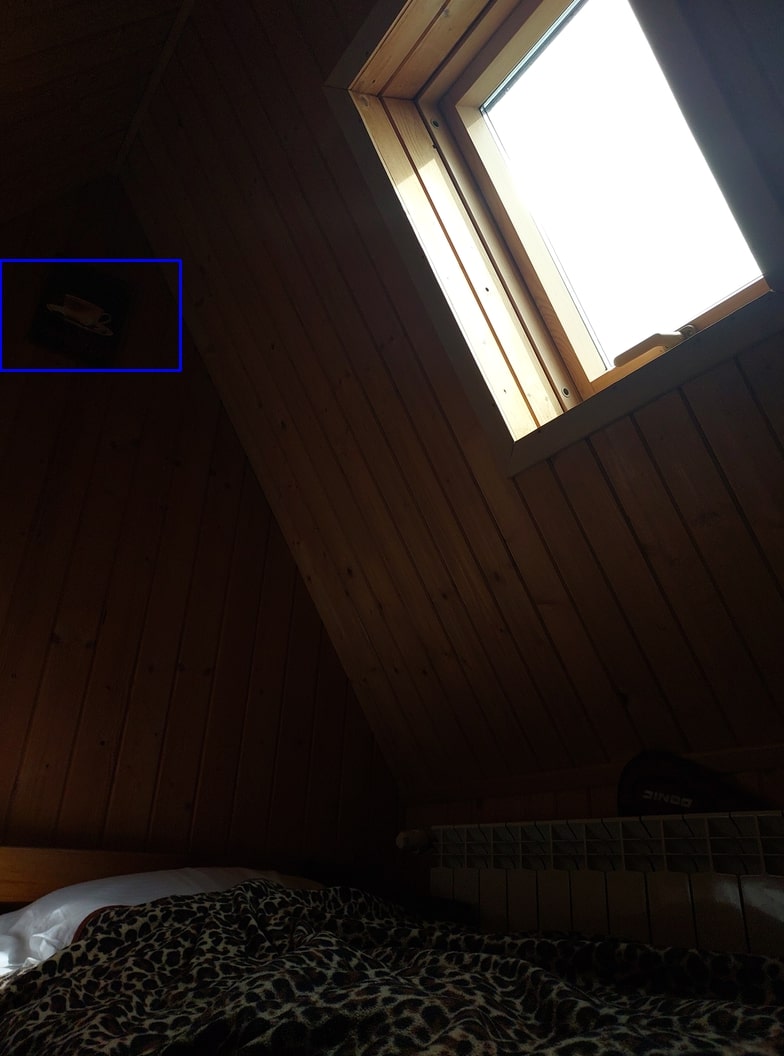}
    \includegraphics[width=0.19\textwidth]{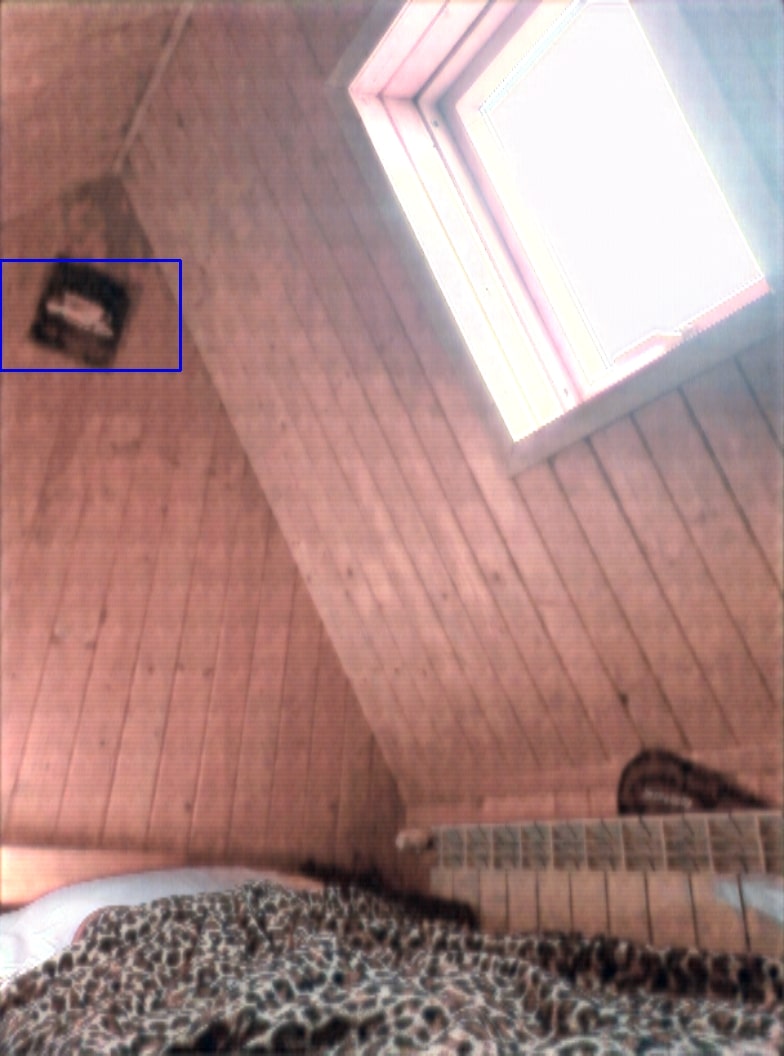}
    \includegraphics[width=0.19\textwidth]{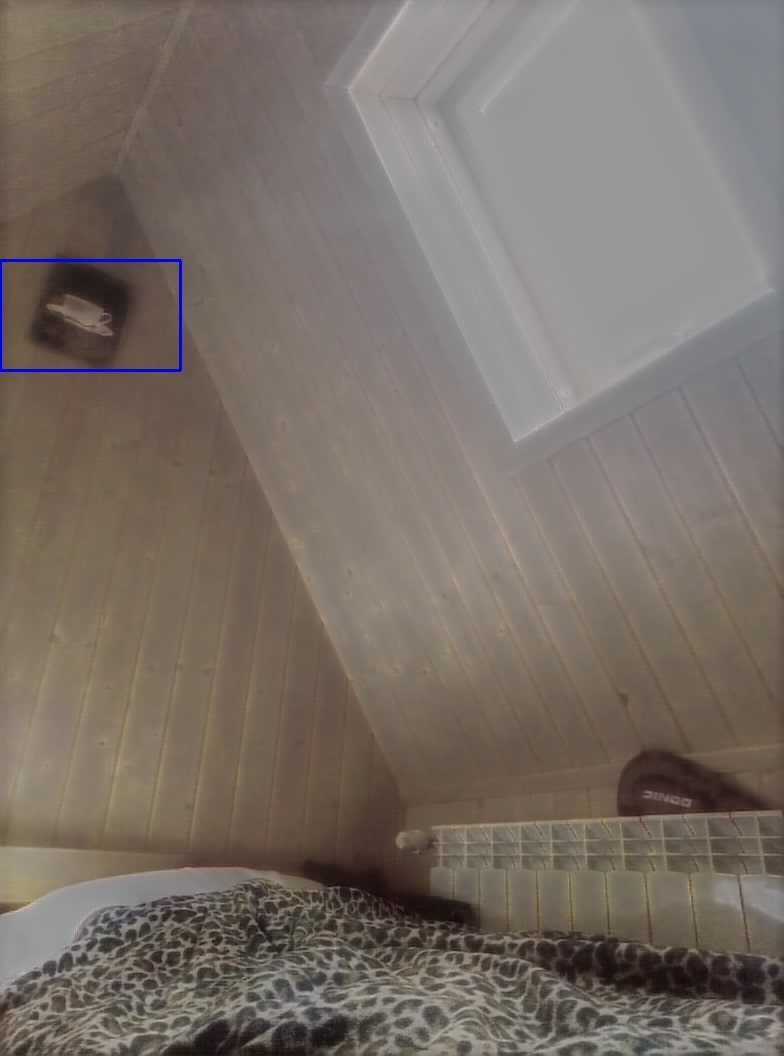}
    \includegraphics[width=0.19\textwidth]{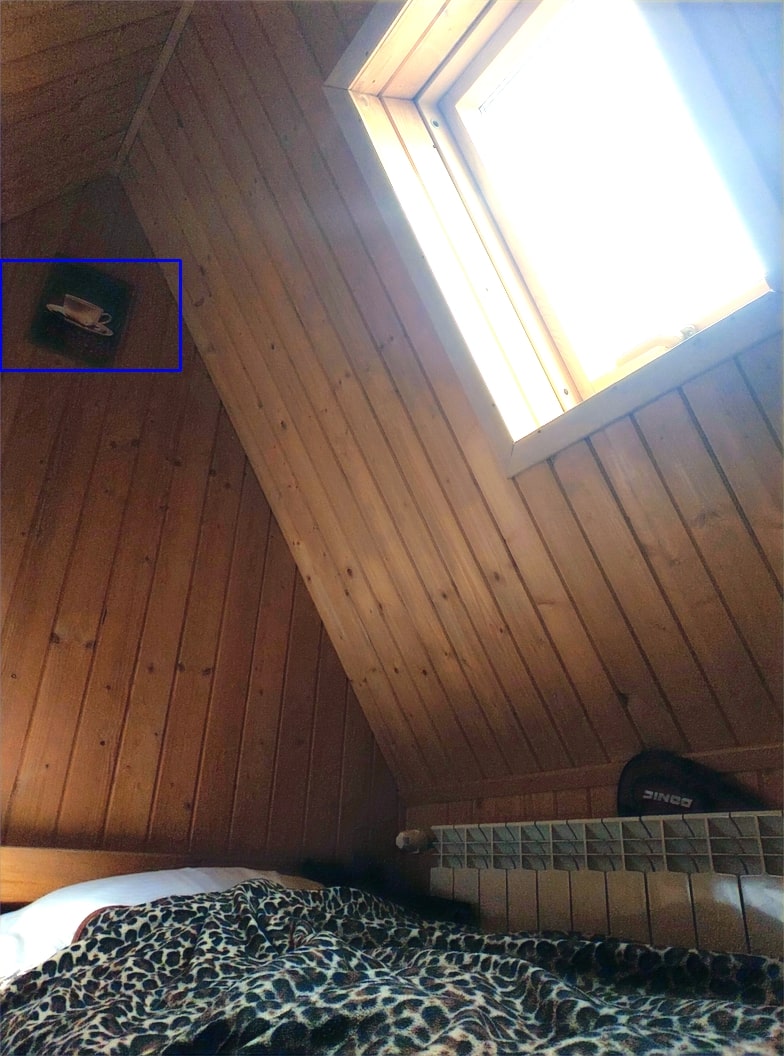}
    \includegraphics[width=0.19\textwidth]{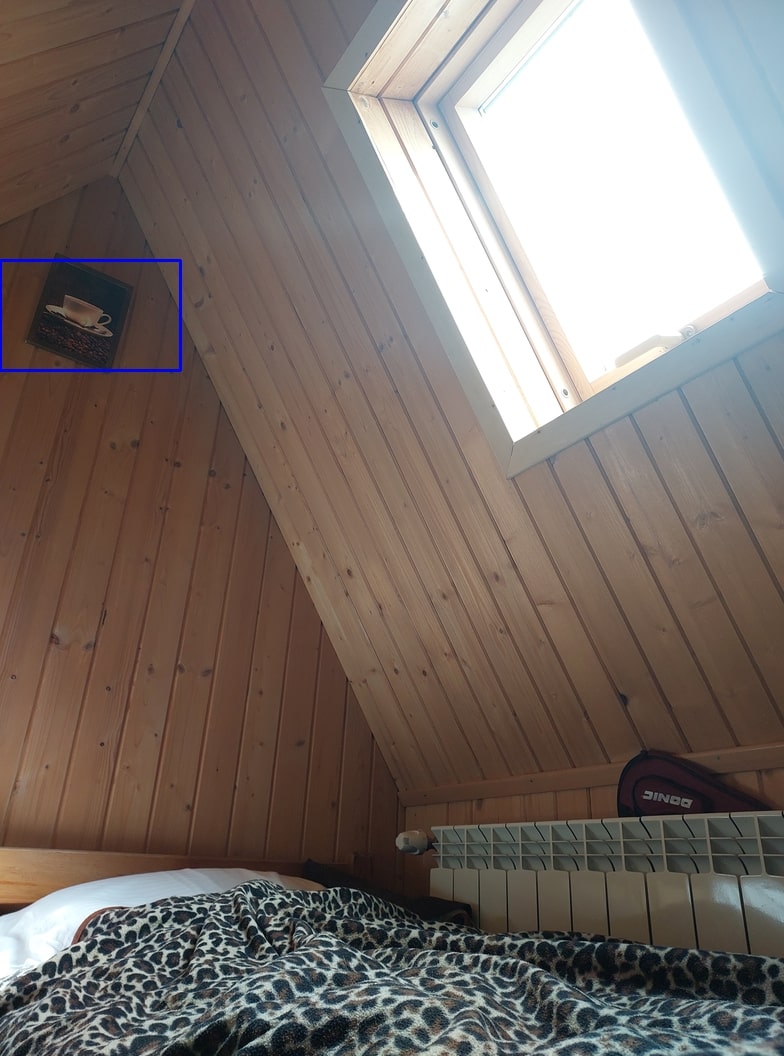}
    
    \begin{subfigure}{0.19\textwidth}
        \includegraphics[width=\textwidth]{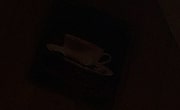}
        \caption*{Input}
    \end{subfigure}
    \begin{subfigure}{0.19\textwidth}
        \includegraphics[width=\textwidth]{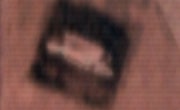}
        \caption*{SNR-Net}
    \end{subfigure}
    \begin{subfigure}{0.19\textwidth}
        \includegraphics[width=\textwidth]{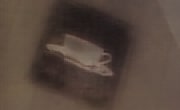}
        \caption*{LLFlow}
    \end{subfigure}
    \begin{subfigure}{0.19\textwidth}
        \includegraphics[width=\textwidth]{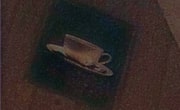}
        \caption*{Dimma}
    \end{subfigure}
    \begin{subfigure}{0.19\textwidth}
        \includegraphics[width=\textwidth]{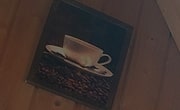}
        \caption*{Ground truth}
    \end{subfigure}
    
    \caption{Results on three test image pairs from our FewShow-Dark dataset. All three methods were trained on train split of this dataset consisting of only six training image pairs. Training setup for each method was the same as for LOL dataset.}
    \label{fig:fsd_comparison}
\end{figure*}

Figure \ref{fig:fsd_comparison} presents the results of Dimma, SNR-Net, and LLFlow on our dataset called FewShow-Dark (FS-Dark). These methods were trained using six pairs of images from FS-Dark's training set. While SNR-Net and LLFlow perform very well on the larger LOL dataset with many training pairs, they struggle when given only a few training samples. In contrast, Dimma adapts to different camera setups with just a few training pairs, generating high-quality samples for the FS-Dark dataset. 

\begin{figure*}
    \centering

    \hspace{0.22\textwidth}
    \begin{subfigure}{0.22\textwidth}
        \includegraphics[width=\linewidth]{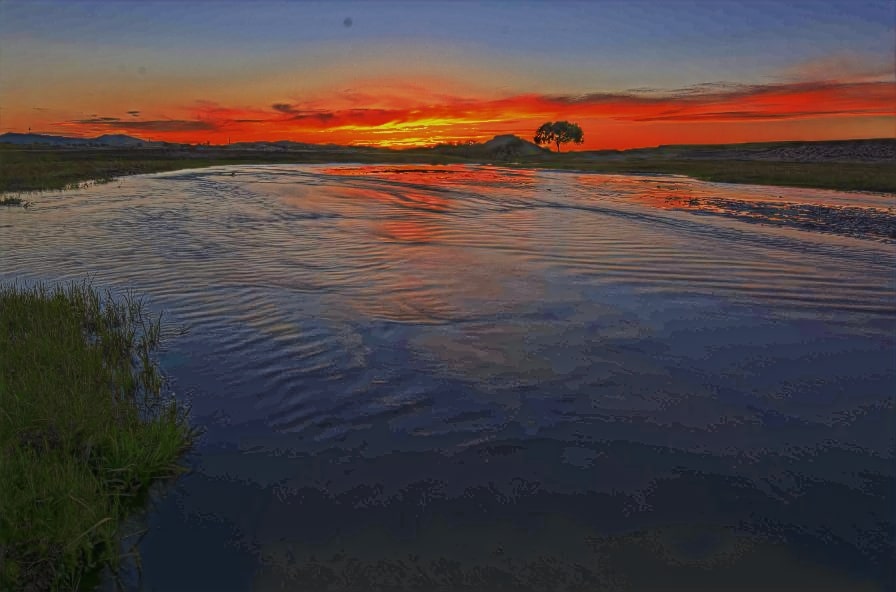}
        \caption*{KinD++ (23\% lum)}
    \end{subfigure}
    \begin{subfigure}{0.22\textwidth}
        \includegraphics[width=\linewidth]{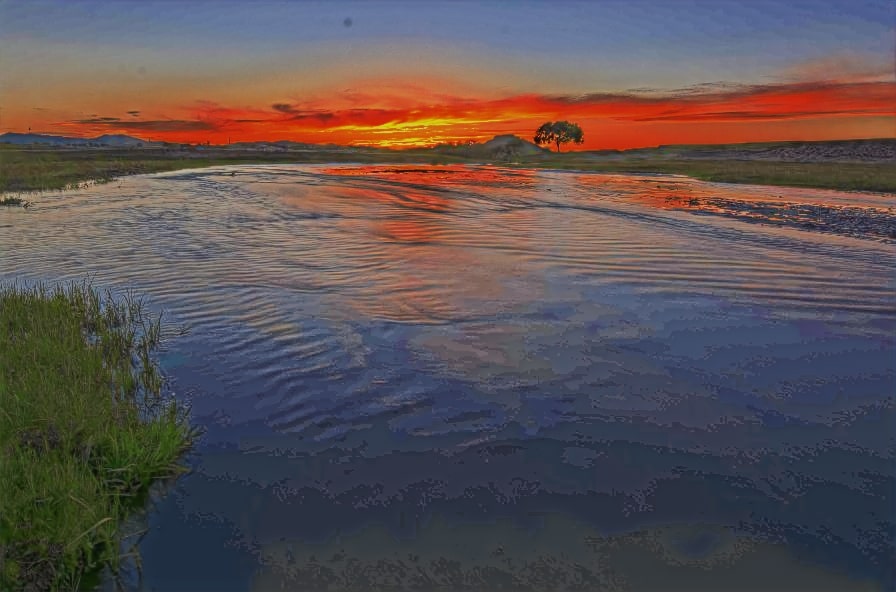}
        \caption*{KinD++ (27\% lum)}
    \end{subfigure}
    \begin{subfigure}{0.22\textwidth}
        \includegraphics[width=\linewidth]{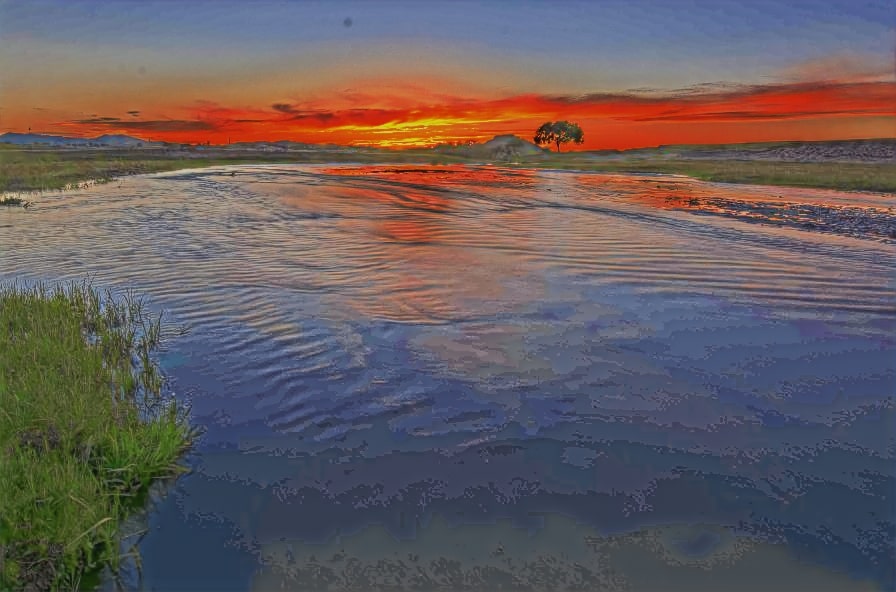}
        \caption*{KinD++ (31\% lum)}
    \end{subfigure}
    
    \hspace{0.22\textwidth}
    \begin{subfigure}{0.22\textwidth}
        \includegraphics[width=\linewidth]{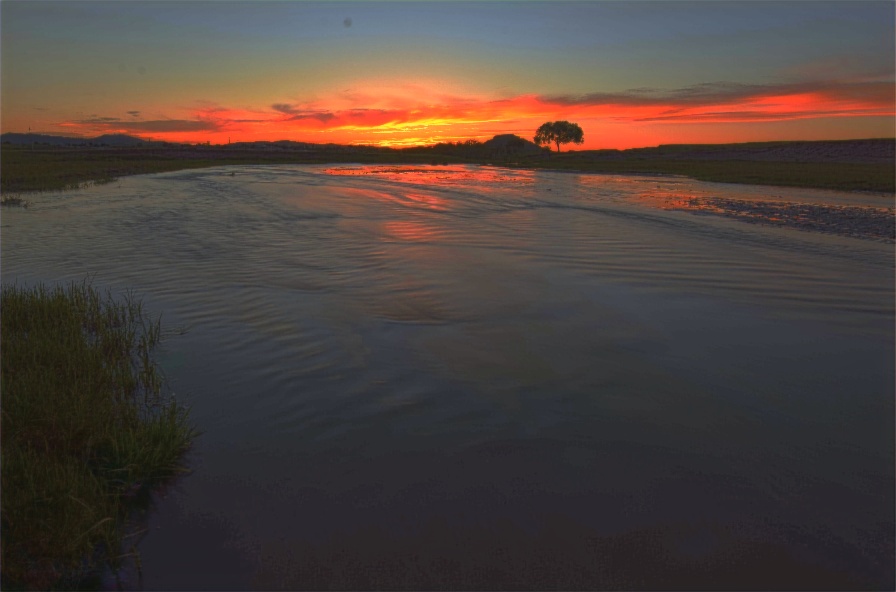}
        \caption*{Dimma 3 pairs (27\% lum)}
    \end{subfigure}
    \begin{subfigure}{0.22\textwidth}
        \includegraphics[width=\linewidth]{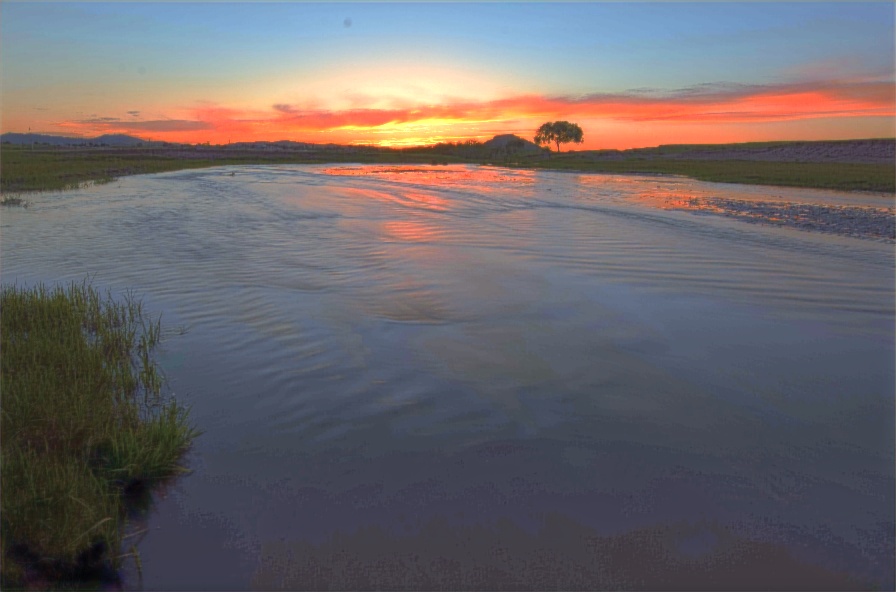}
        \caption*{Dimma 3 pairs (41\% lum)}
    \end{subfigure}
    \begin{subfigure}{0.22\textwidth}
        \includegraphics[width=\linewidth]{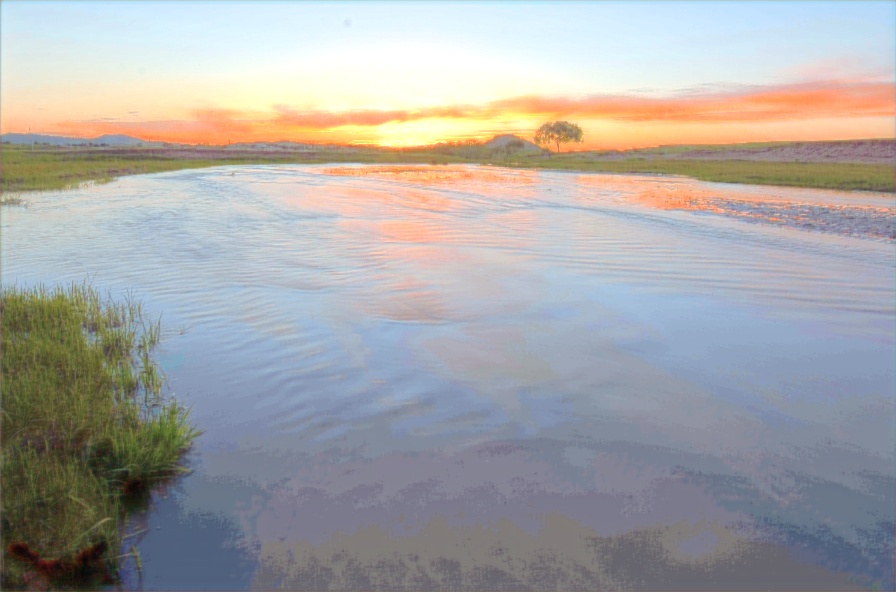}
        \caption*{Dimma 3 pairs (65\% lum)}
    \end{subfigure}
    
    \hspace{0.22\textwidth}
    \begin{subfigure}{0.22\textwidth}
        \includegraphics[width=\linewidth]{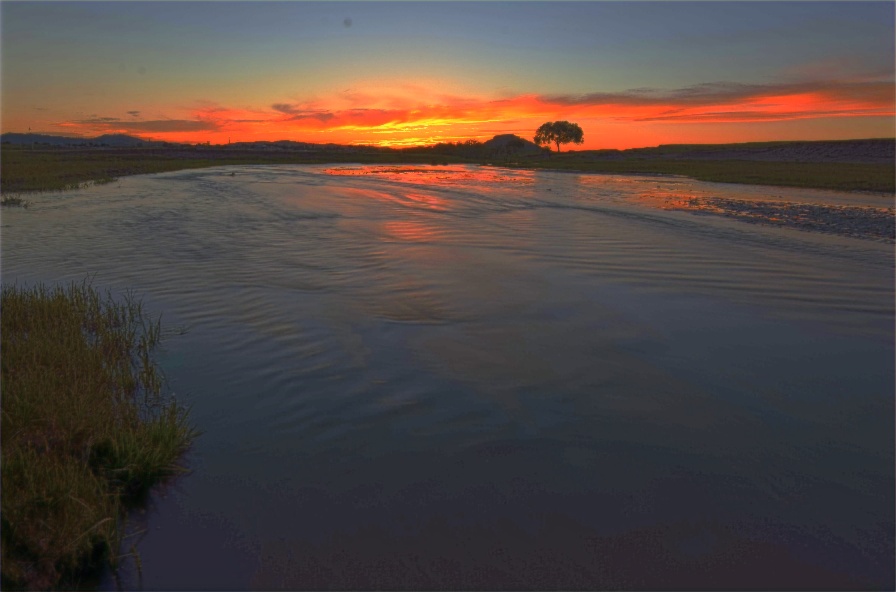}
        \caption*{Dimma 5 pairs (29\% lum)}
    \end{subfigure}
    \begin{subfigure}{0.22\textwidth}
        \includegraphics[width=\linewidth]{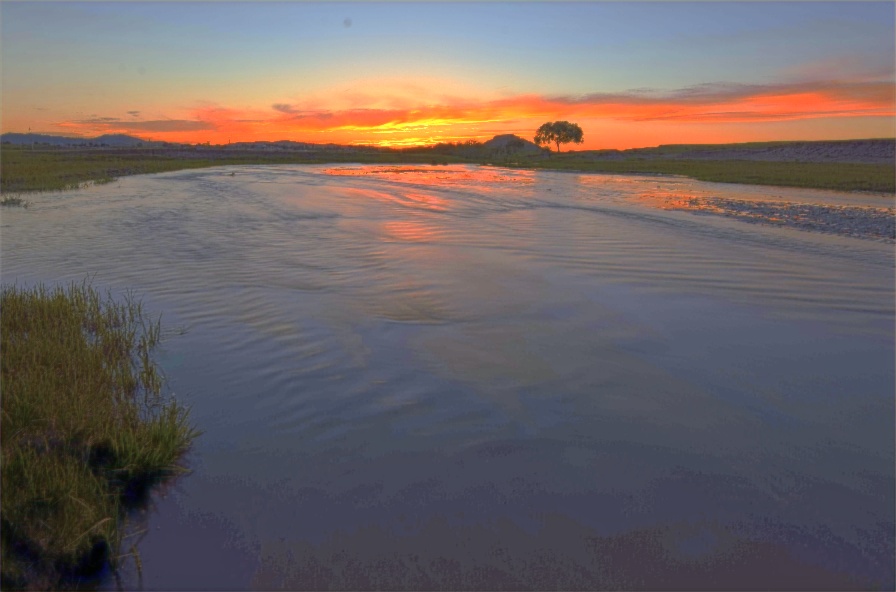}
        \caption*{Dimma 5 pairs (41\% lum)}
    \end{subfigure}
    \begin{subfigure}{0.22\textwidth}
        \includegraphics[width=\linewidth]{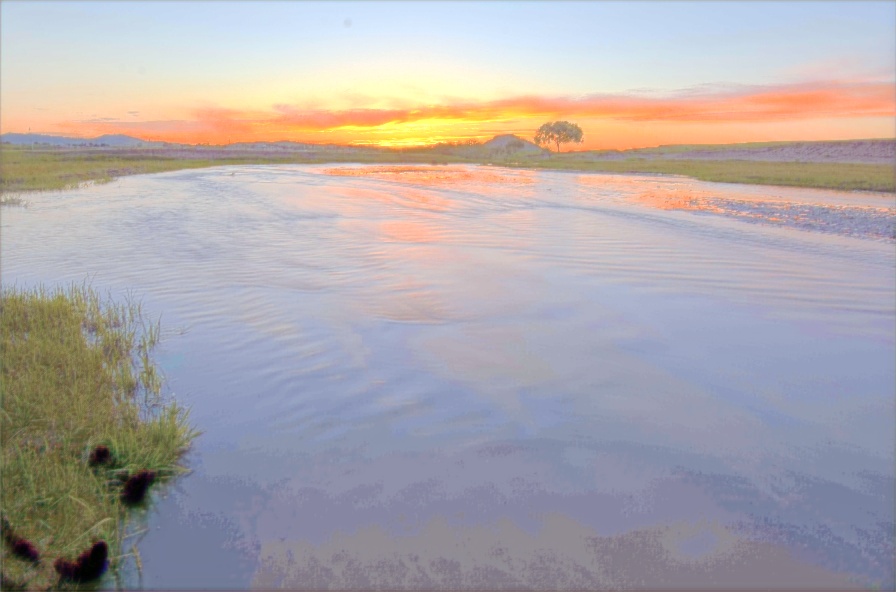}
        \caption*{Dimma 5 pairs (67\% lum)}
    \end{subfigure}
    
    \hspace{0.22\textwidth}
    \begin{subfigure}{0.22\textwidth}
        \includegraphics[width=\linewidth]{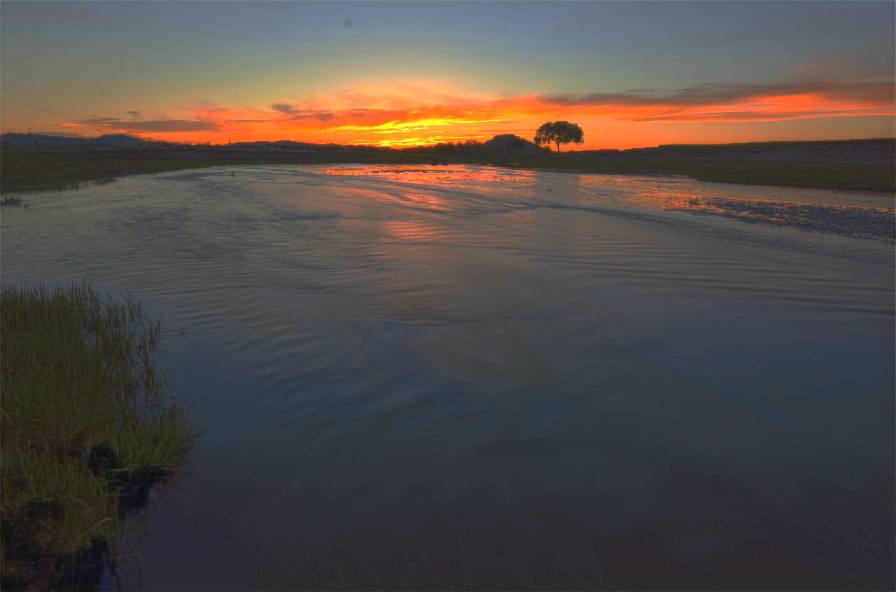}
        \caption*{Dimma 8 pairs (30\% lum)}
    \end{subfigure}
    \begin{subfigure}{0.22\textwidth}
        \includegraphics[width=\linewidth]{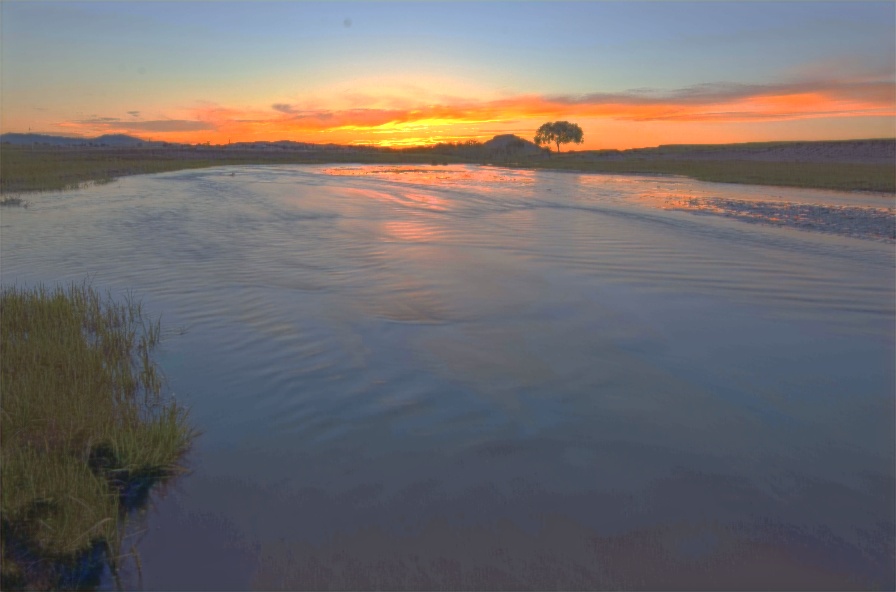}
        \caption*{Dimma 8 pairs (43\% lum)}
    \end{subfigure}
    \begin{subfigure}{0.22\textwidth}
        \includegraphics[width=\linewidth]{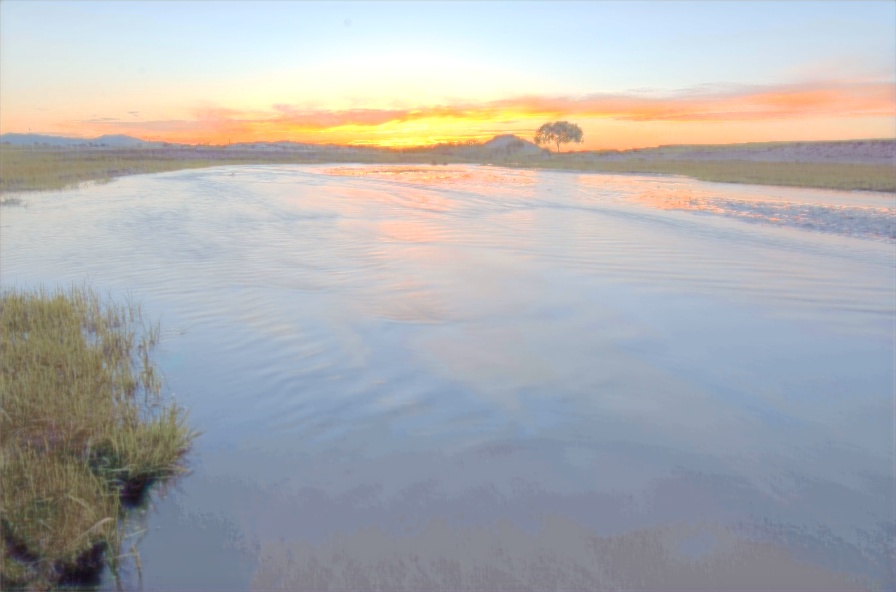}
        \caption*{Dimma 8 pairs (69\% lum)}
    \end{subfigure}

    \hspace{0.22\textwidth}
    \begin{subfigure}{0.22\textwidth}
        \includegraphics[width=\linewidth]{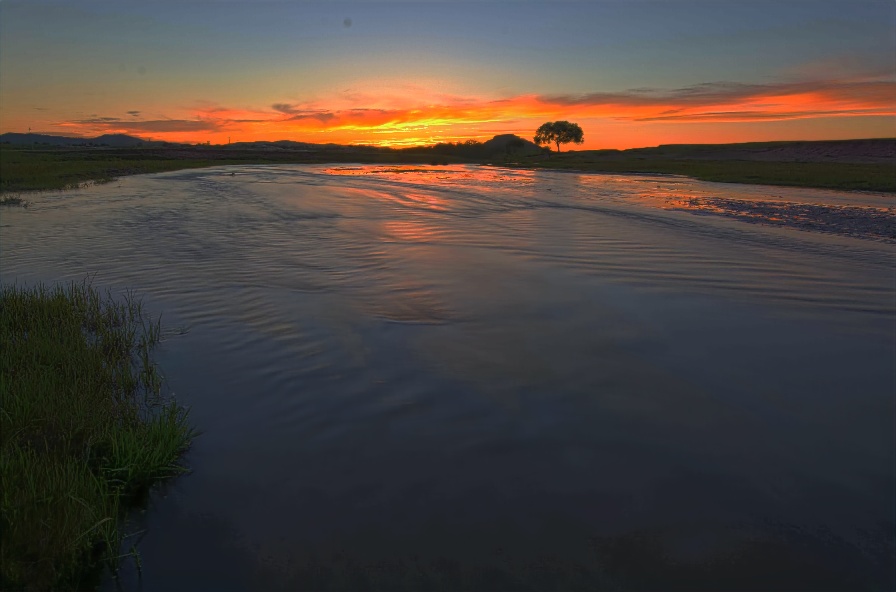}
        \caption*{Dimma full (23\% lum)}
    \end{subfigure}
    \begin{subfigure}{0.22\textwidth}
        \includegraphics[width=\linewidth]{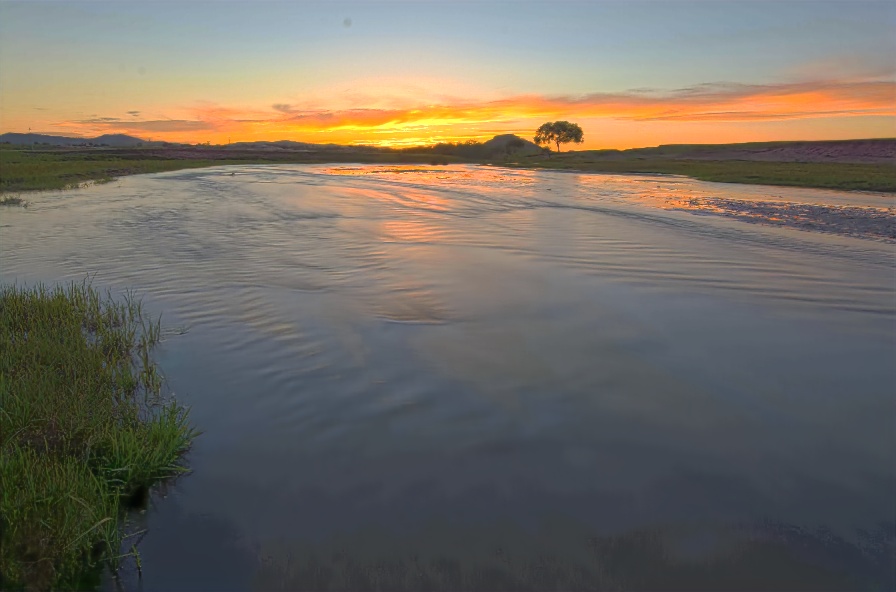}
        \caption*{Dimma full (37\% lum)}
    \end{subfigure}
    \begin{subfigure}{0.22\textwidth}
        \includegraphics[width=\linewidth]{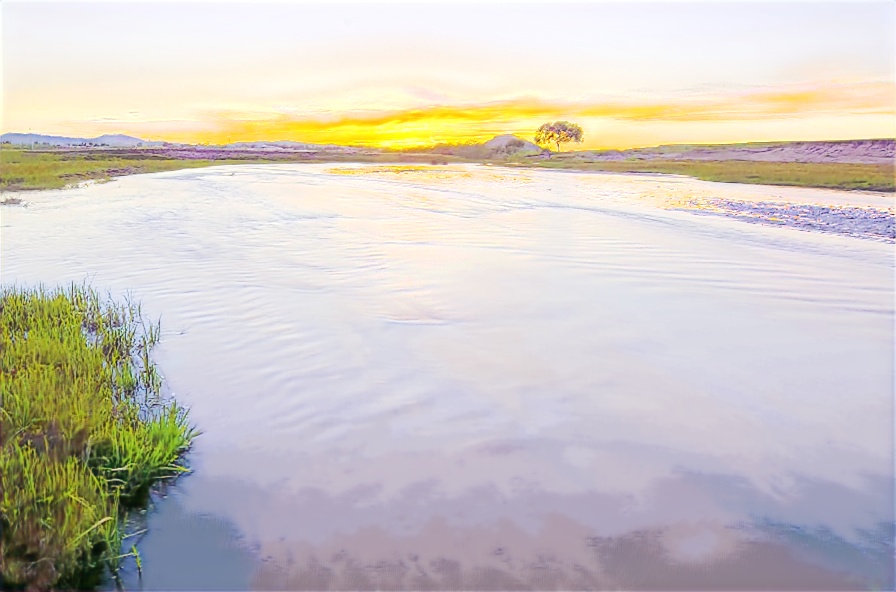}
        \caption*{Dimma full (74\% lum)}
    \end{subfigure}
    
    \begin{subfigure}{0.22\textwidth}
        \includegraphics[width=\linewidth]{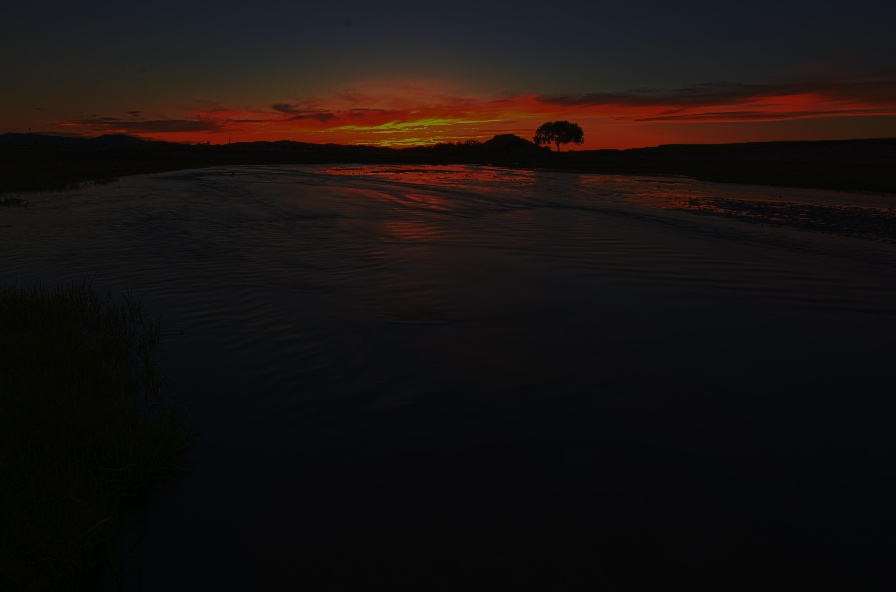}
        \caption*{Input (6\% lum)}
    \end{subfigure}
    \begin{subfigure}{0.22\textwidth}
        \includegraphics[width=\linewidth]{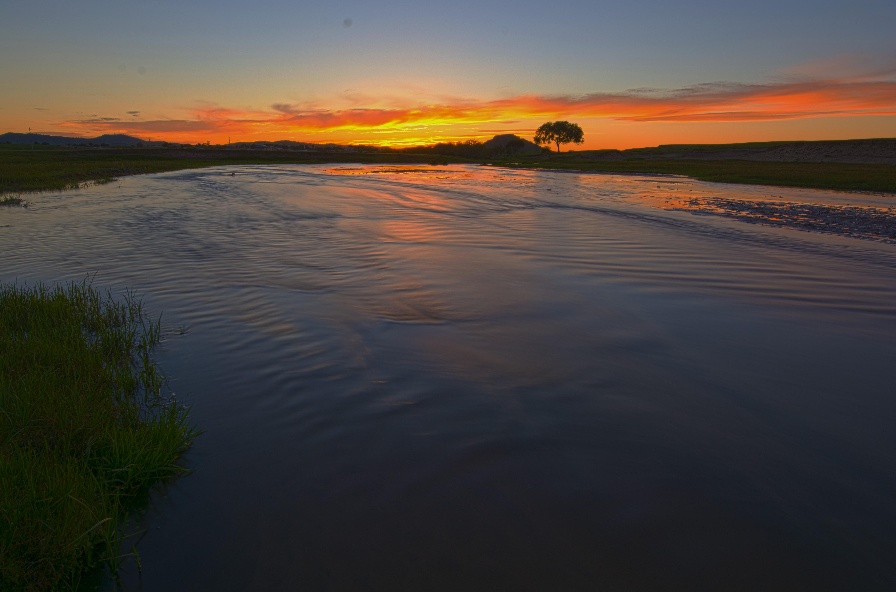}
        \caption*{Ground truth (23\% lum)}
    \end{subfigure}
    \begin{subfigure}{0.22\textwidth}
        \includegraphics[width=\linewidth]{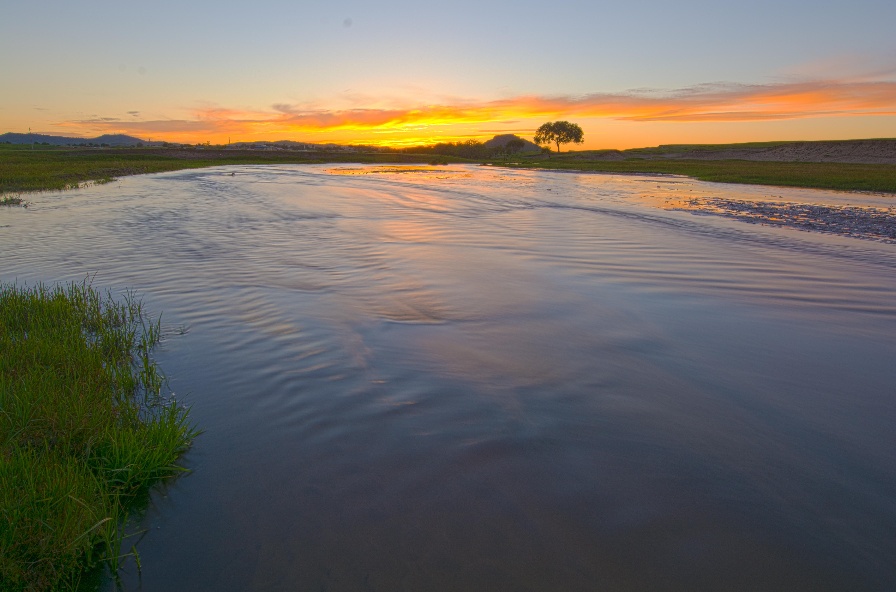}
        \caption*{Ground truth (37\% lum)}
    \end{subfigure}
    \begin{subfigure}{0.22\textwidth}
        \includegraphics[width=\linewidth]{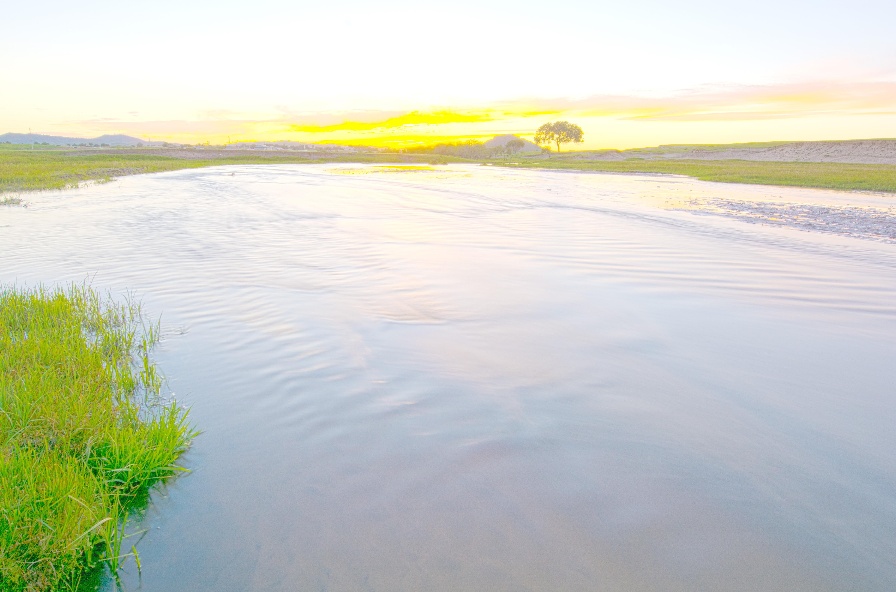}
        \caption*{Ground truth (76\% lum)}
    \end{subfigure}
    
    \caption{Visualization of low light enhancement with different brightening factors for Dimma and KinD++ on the image from SICE dataset. Both models were conditioned by the illumination values of the ground truth images displayed at the bottom. The average illumination of each image is shown in brackets, indicating the accuracy of the models in generating images with specific light levels.}
    \label{fig:dimma_light_results}
\end{figure*}

\begin{figure*}
    \centering

    \hspace{0.22\textwidth}
    \begin{subfigure}{0.22\textwidth}
        \includegraphics[width=\linewidth]{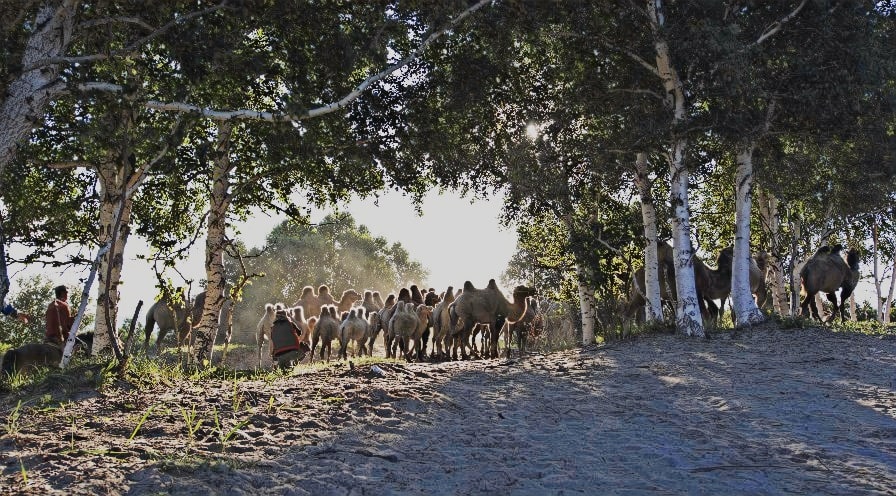}
        \caption*{KinD++ (32\% lum)}
    \end{subfigure}
    \begin{subfigure}{0.22\textwidth}
        \includegraphics[width=\linewidth]{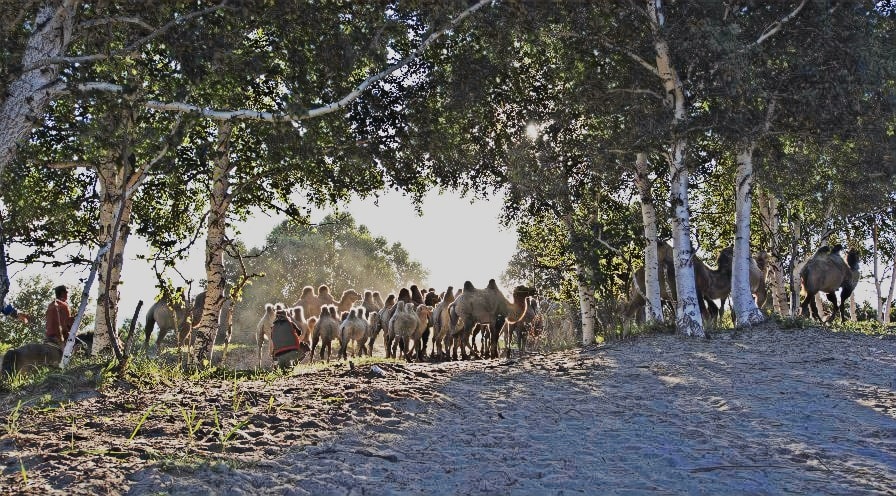}
        \caption*{KinD++ (36\% lum)}
    \end{subfigure}
    \begin{subfigure}{0.22\textwidth}
        \includegraphics[width=\linewidth]{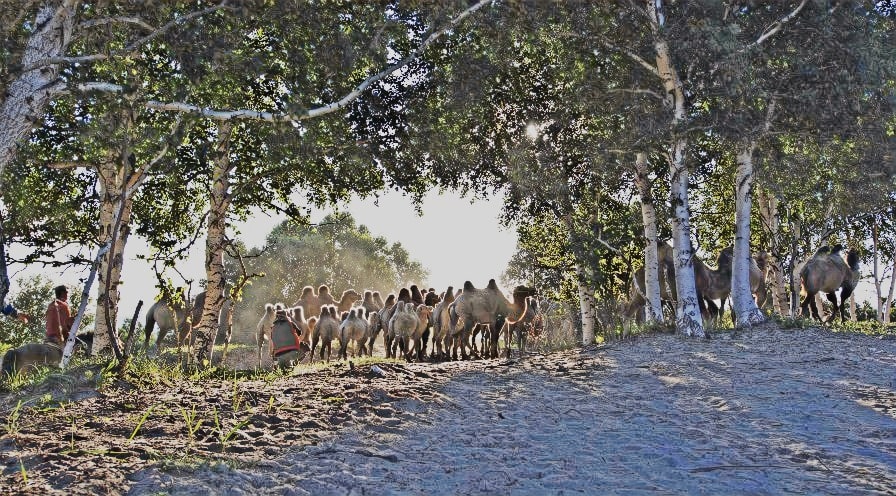}
        \caption*{KinD++ (40\% lum)}
    \end{subfigure}
    
    \hspace{0.22\textwidth}
    \begin{subfigure}{0.22\textwidth}
        \includegraphics[width=\linewidth]{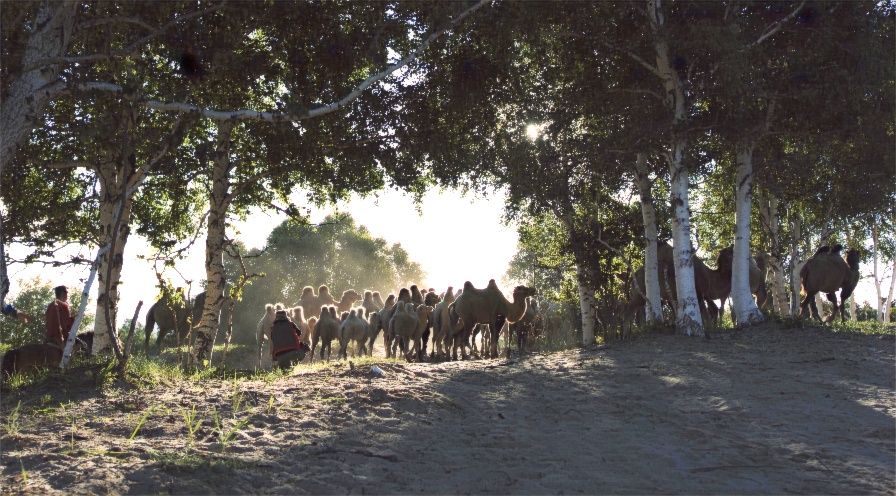}
        \caption*{Dimma 3 pairs (34\% lum)}
    \end{subfigure}
    \begin{subfigure}{0.22\textwidth}
        \includegraphics[width=\linewidth]{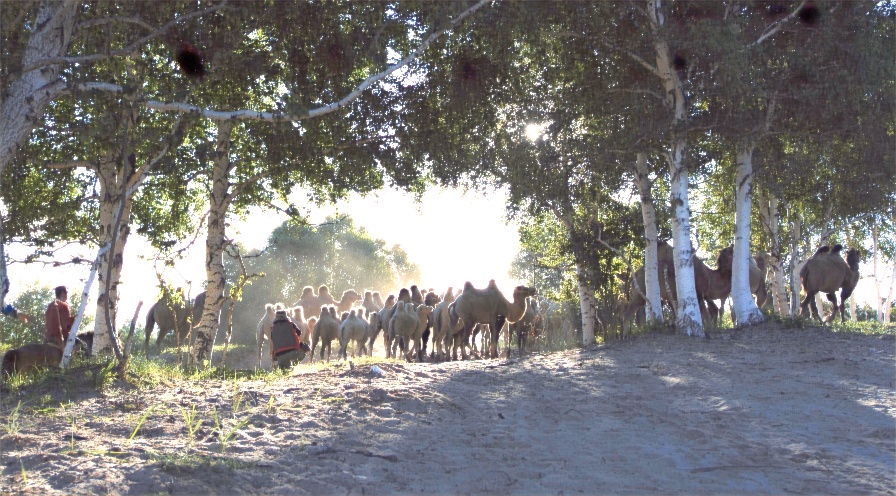}
        \caption*{Dimma 3 pairs (48\% lum)}
    \end{subfigure}
    \begin{subfigure}{0.22\textwidth}
        \includegraphics[width=\linewidth]{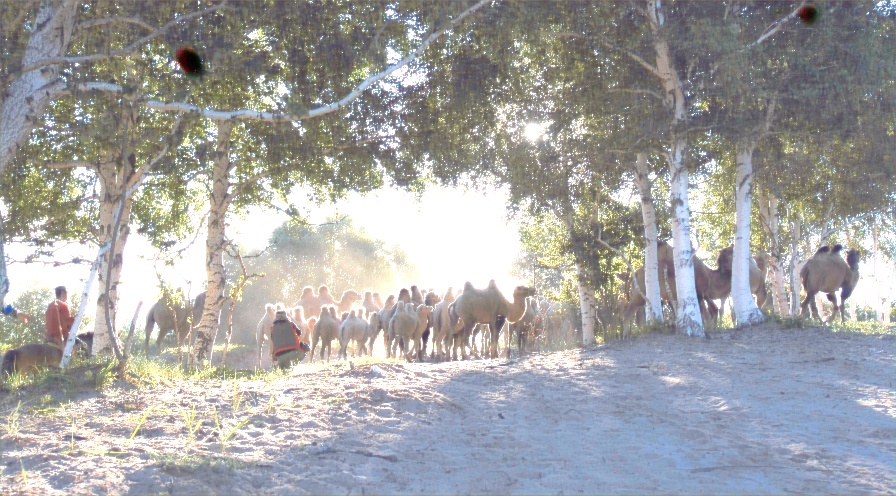}
        \caption*{Dimma 3 pairs (64\% lum)}
    \end{subfigure}
    
    \hspace{0.22\textwidth}
    \begin{subfigure}{0.22\textwidth}
        \includegraphics[width=\linewidth]{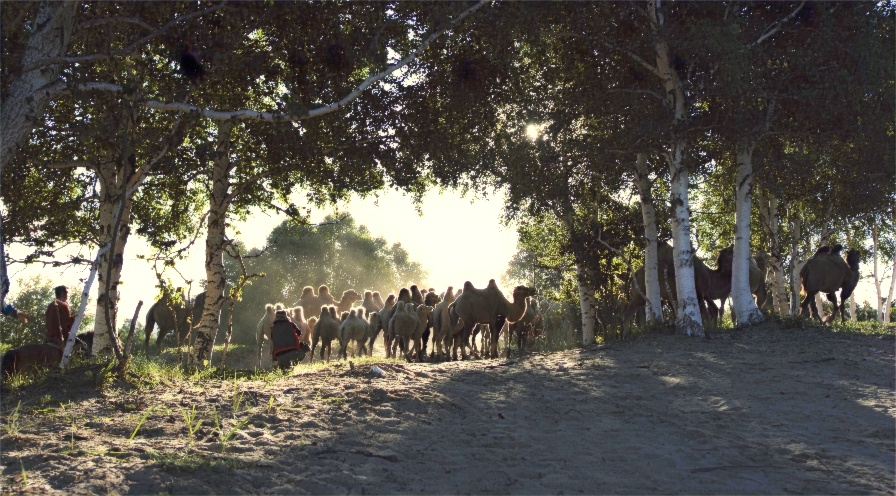}
        \caption*{Dimma 5 pairs (32\% lum)}
    \end{subfigure}
    \begin{subfigure}{0.22\textwidth}
        \includegraphics[width=\linewidth]{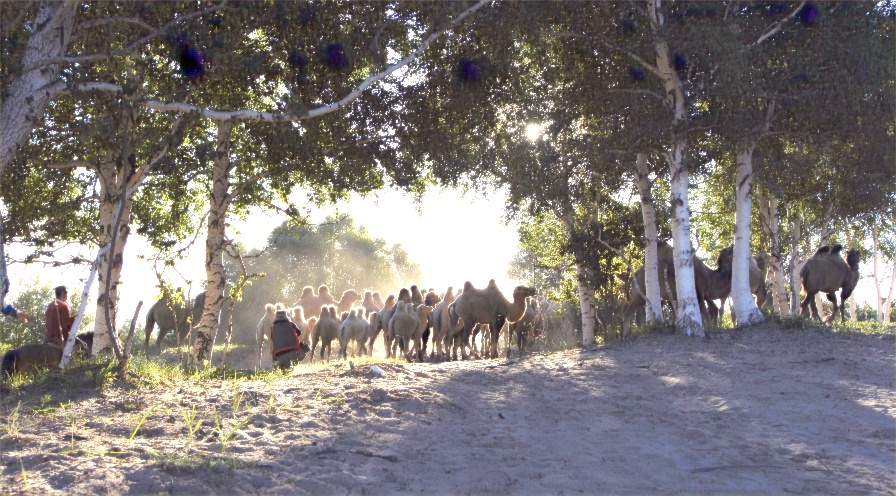}
        \caption*{Dimma 5 pairs (48\% lum)}
    \end{subfigure}
    \begin{subfigure}{0.22\textwidth}
        \includegraphics[width=\linewidth]{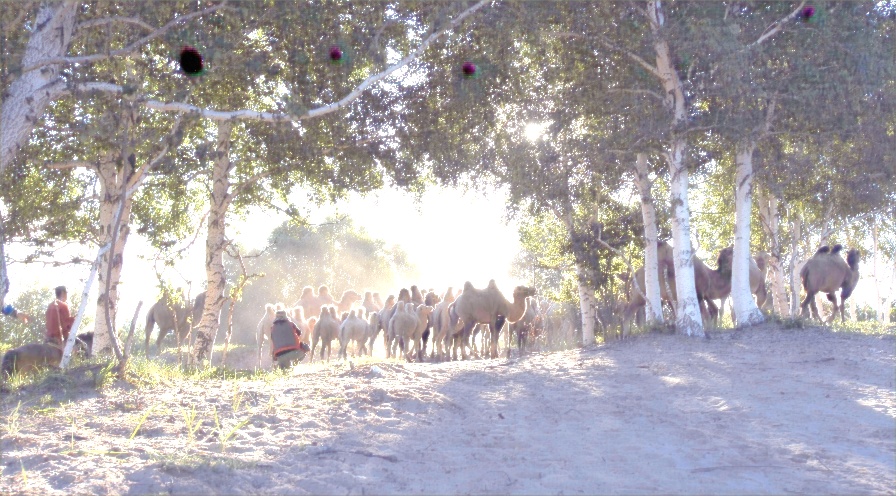}
        \caption*{Dimma 5 pairs (67\% lum)}
    \end{subfigure}
    
    \hspace{0.22\textwidth}
    \begin{subfigure}{0.22\textwidth}
        \includegraphics[width=\linewidth]{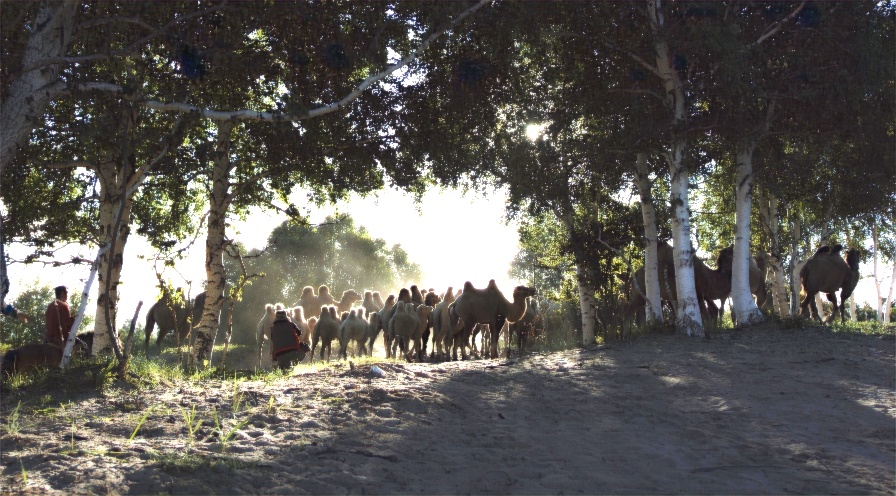}
        \caption*{Dimma 8 pairs (35\% lum)}
    \end{subfigure}
    \begin{subfigure}{0.22\textwidth}
        \includegraphics[width=\linewidth]{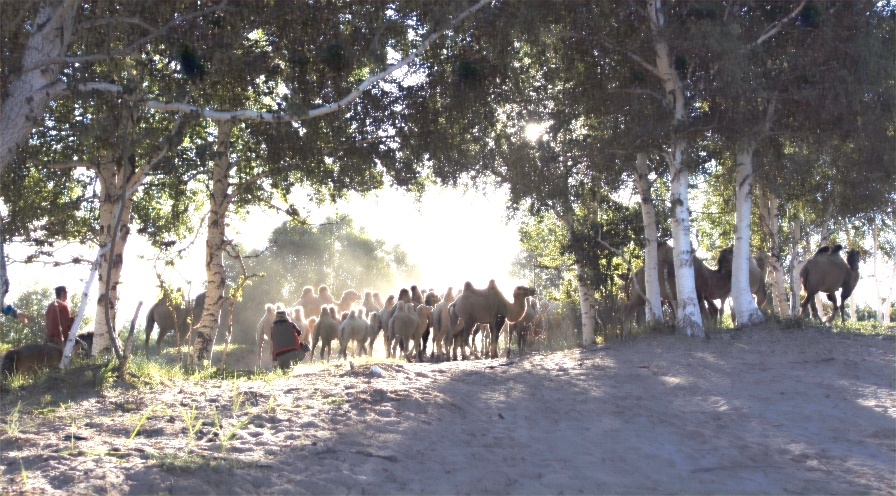}
        \caption*{Dimma 8 pairs (48\% lum)}
    \end{subfigure}
    \begin{subfigure}{0.22\textwidth}
        \includegraphics[width=\linewidth]{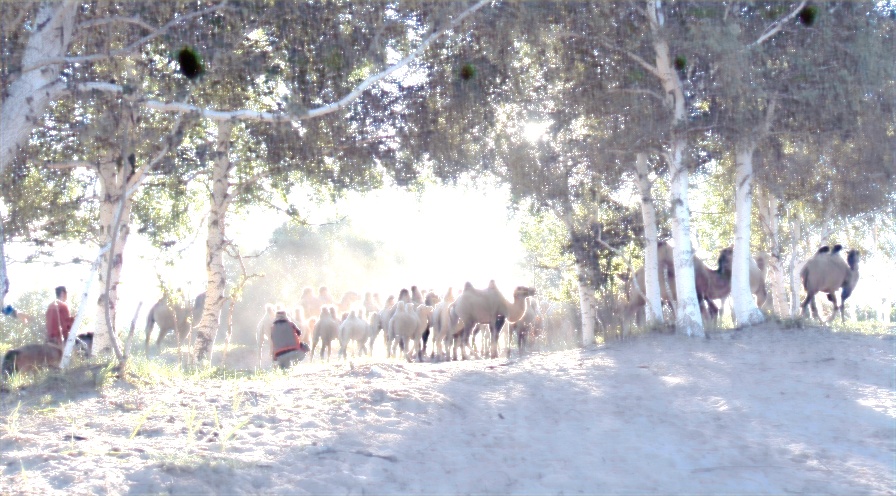}
        \caption*{Dimma 8 pairs (71\% lum)}
    \end{subfigure}

    \hspace{0.22\textwidth}
    \begin{subfigure}{0.22\textwidth}
        \includegraphics[width=\linewidth]{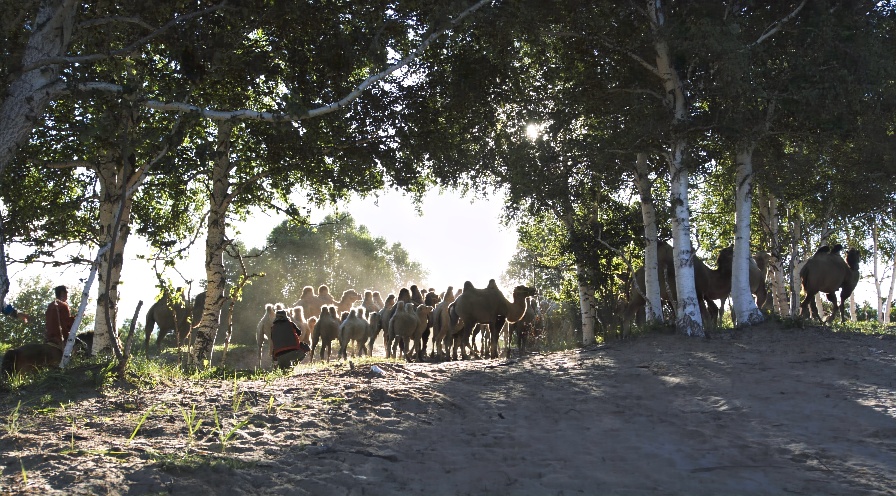}
        \caption*{Dimma full (34\% lum)}
    \end{subfigure}
    \begin{subfigure}{0.22\textwidth}
        \includegraphics[width=\linewidth]{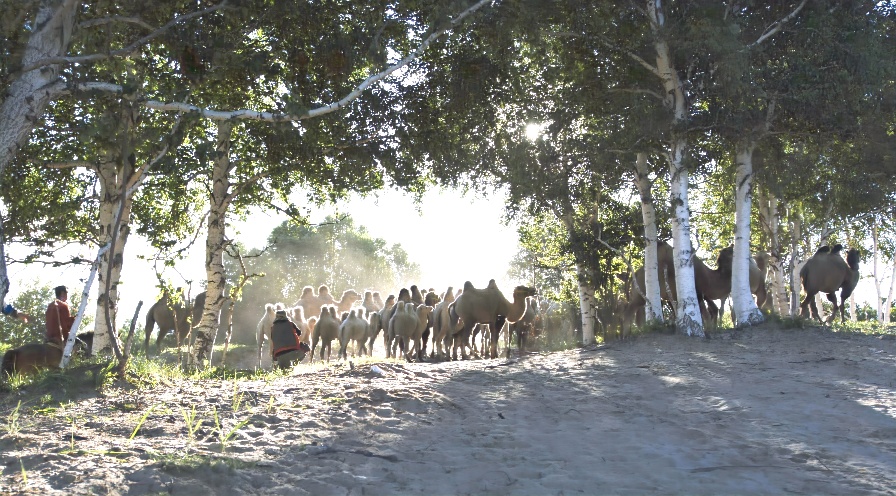}
        \caption*{Dimma full (48\% lum)}
    \end{subfigure}
    \begin{subfigure}{0.22\textwidth}
        \includegraphics[width=\linewidth]{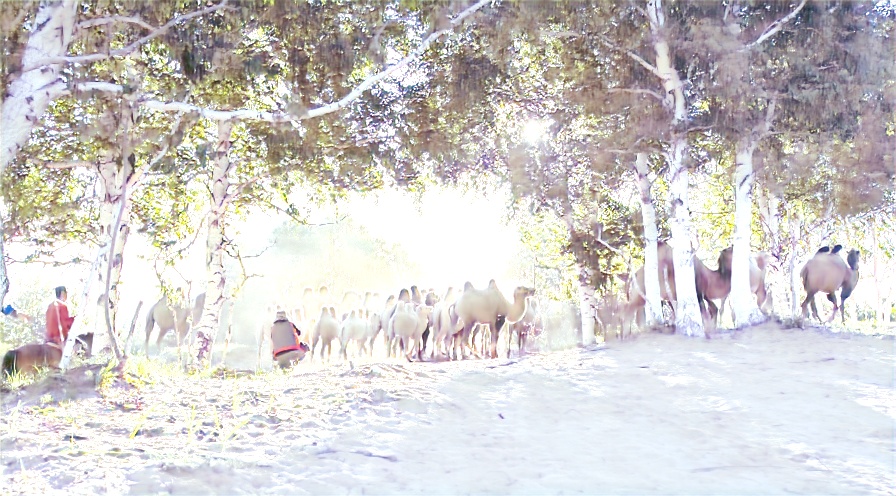}
        \caption*{Dimma full (81\% lum)}
    \end{subfigure}
    
    \begin{subfigure}{0.22\textwidth}
        \includegraphics[width=\linewidth]{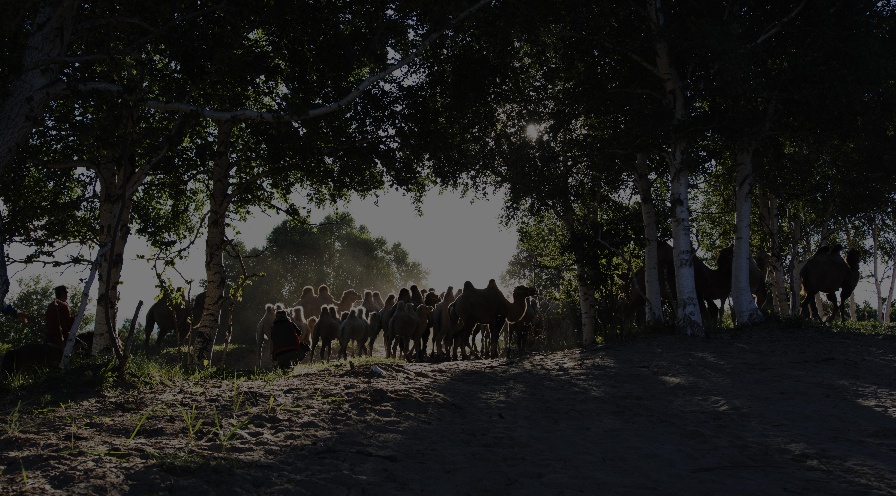}
        \caption*{Input (11\% lum)}
    \end{subfigure}
    \begin{subfigure}{0.22\textwidth}
        \includegraphics[width=\linewidth]{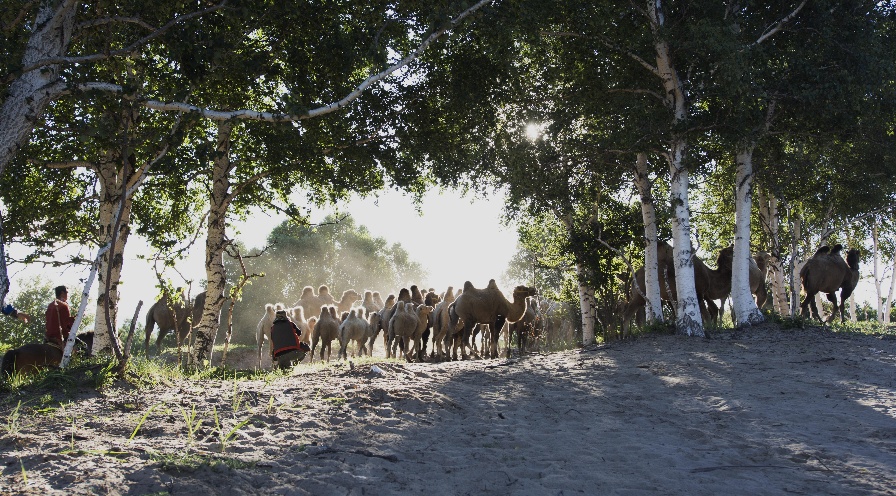}
        \caption*{Ground truth (36\% lum)}
    \end{subfigure}
    \begin{subfigure}{0.22\textwidth}
        \includegraphics[width=\linewidth]{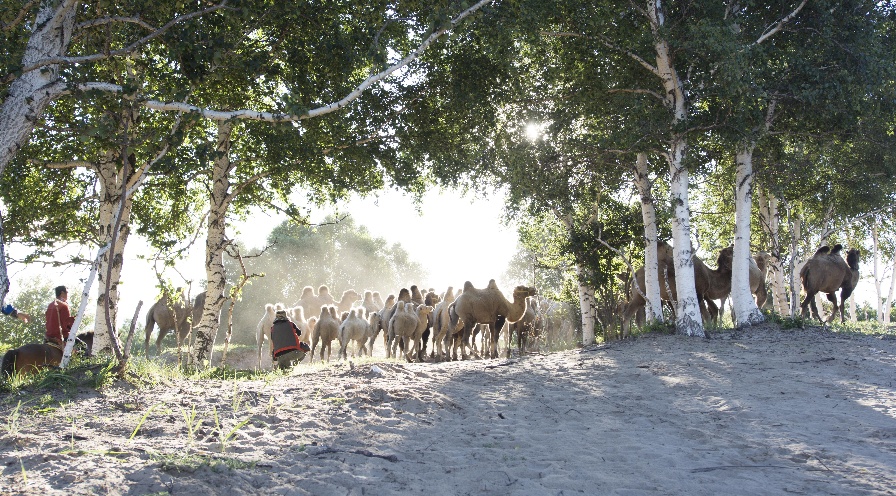}
        \caption*{Ground truth (49\% lum)}
    \end{subfigure}
    \begin{subfigure}{0.22\textwidth}
        \includegraphics[width=\linewidth]{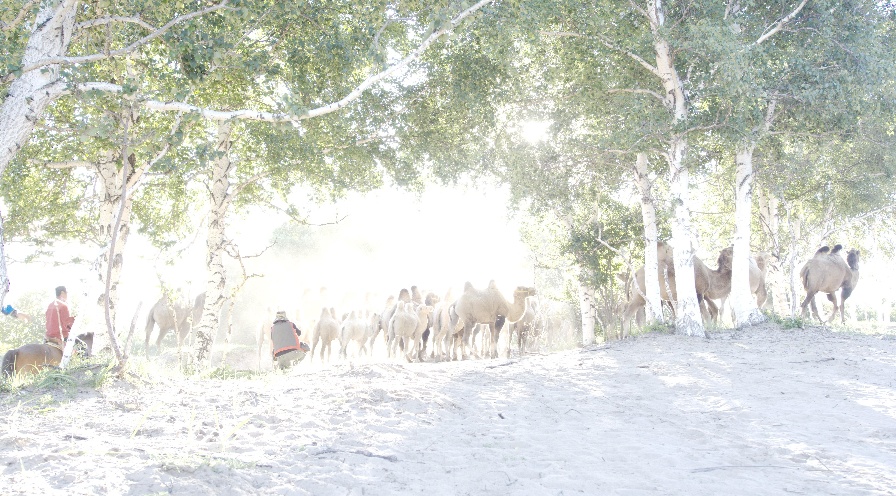}
        \caption*{Ground truth (83\% lum)}
    \end{subfigure}
    
    \caption{Visualization of low light enhancement with different brightening factors for Dimma and KinD++ on the image from SICE dataset. Both models were conditioned by the illumination values of the ground truth images displayed at the bottom. The average illumination of each image is shown in brackets, indicating the accuracy of the models in generating images with specific light levels.}
    \label{fig:dimma_light_results_2}
\end{figure*}

\begin{figure*}
    \centering

    \hspace{0.22\textwidth}
    \begin{subfigure}{0.22\textwidth}
        \includegraphics[width=\linewidth]{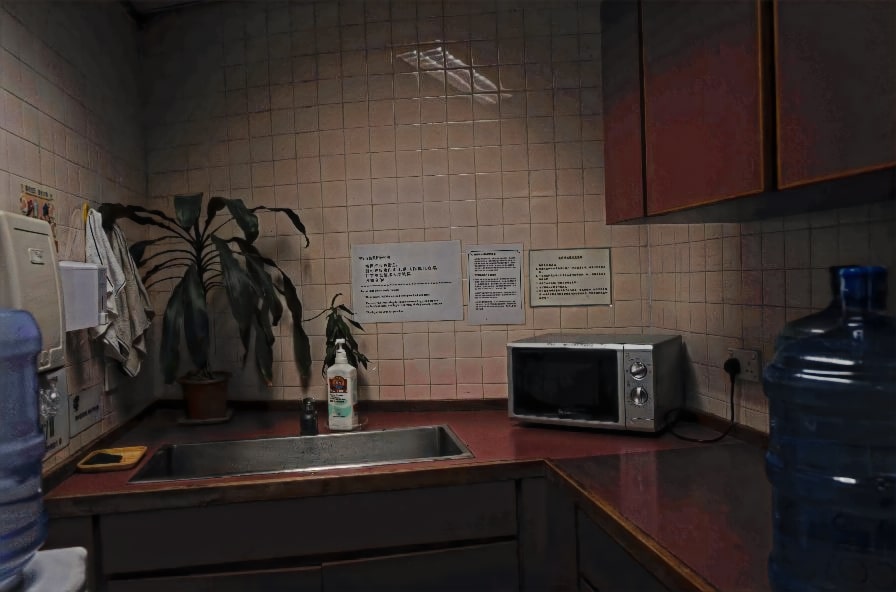}
        \caption*{KinD++ (20\% lum)}
    \end{subfigure}
    \begin{subfigure}{0.22\textwidth}
        \includegraphics[width=\linewidth]{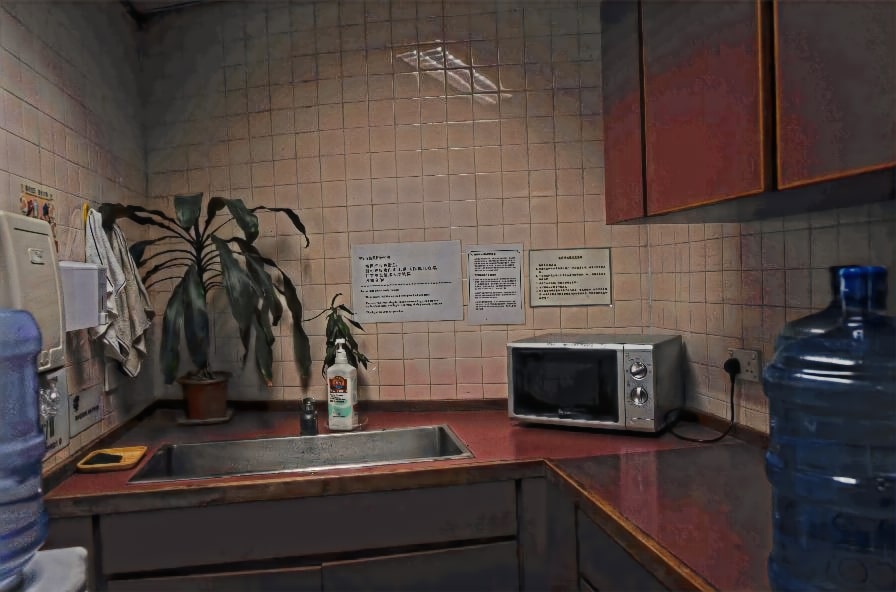}
        \caption*{KinD++ (28\% lum)}
    \end{subfigure}
    \begin{subfigure}{0.22\textwidth}
        \includegraphics[width=\linewidth]{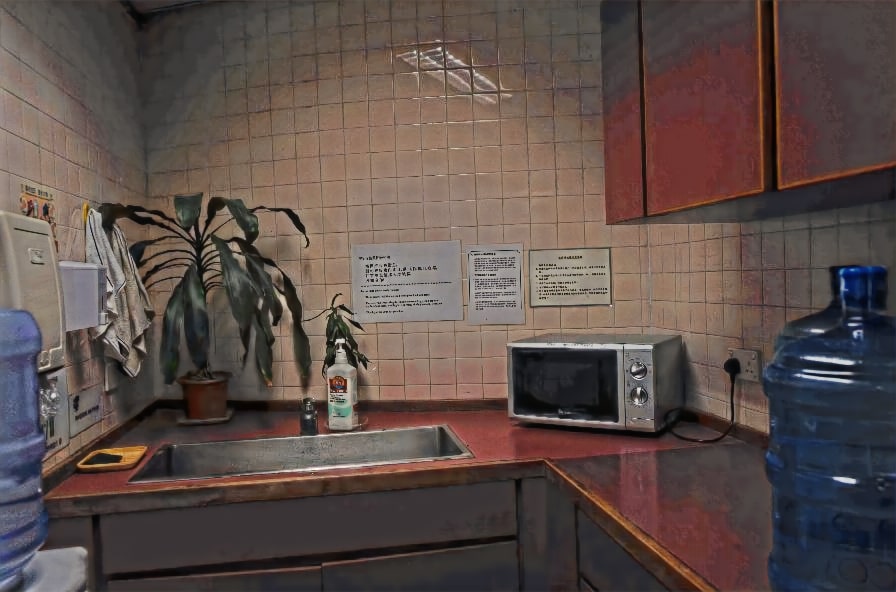}
        \caption*{KinD++ (32\% lum)}
    \end{subfigure}
    
    \hspace{0.22\textwidth}
    \begin{subfigure}{0.22\textwidth}
        \includegraphics[width=\linewidth]{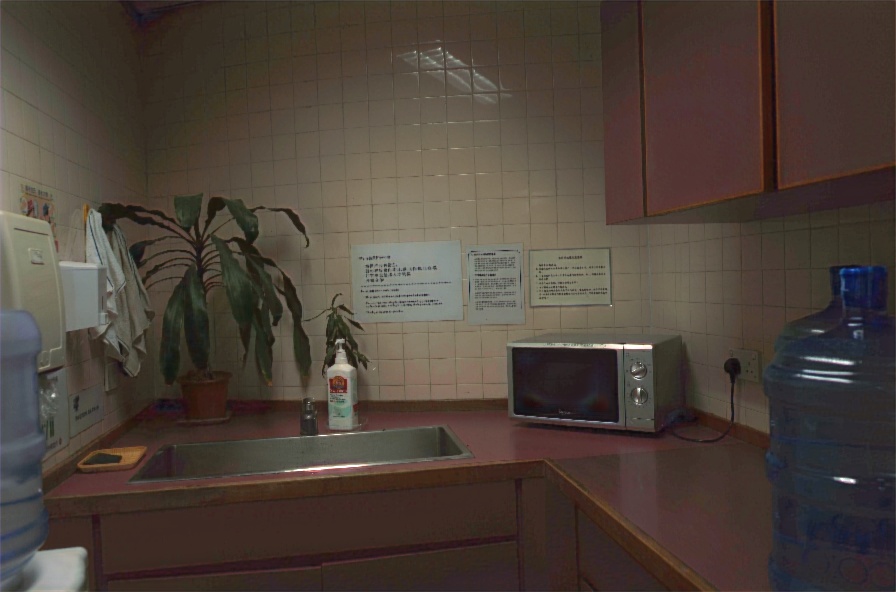}
        \caption*{Dimma 3 pairs (26\% lum)}
    \end{subfigure}
    \begin{subfigure}{0.22\textwidth}
        \includegraphics[width=\linewidth]{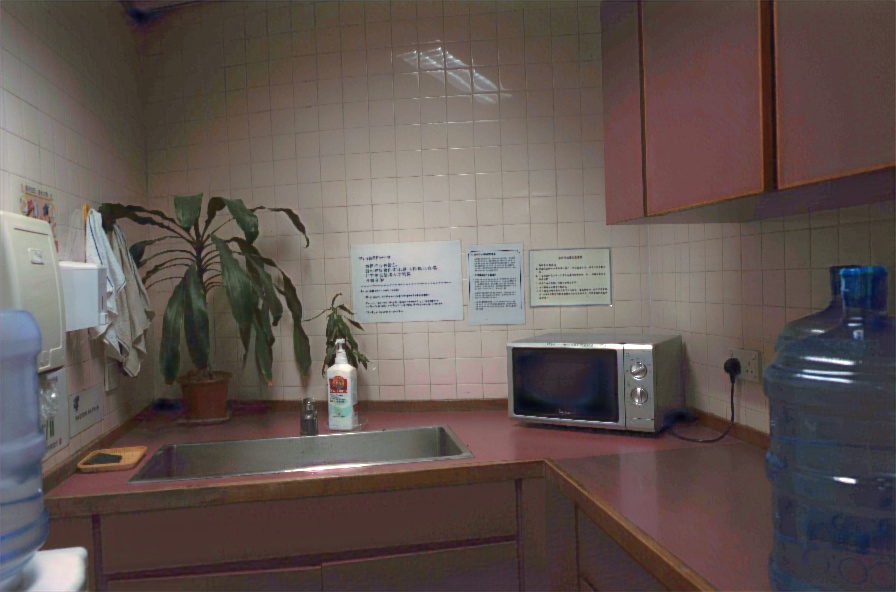}
        \caption*{Dimma 3 pairs (35\% lum)}
    \end{subfigure}
    \begin{subfigure}{0.22\textwidth}
        \includegraphics[width=\linewidth]{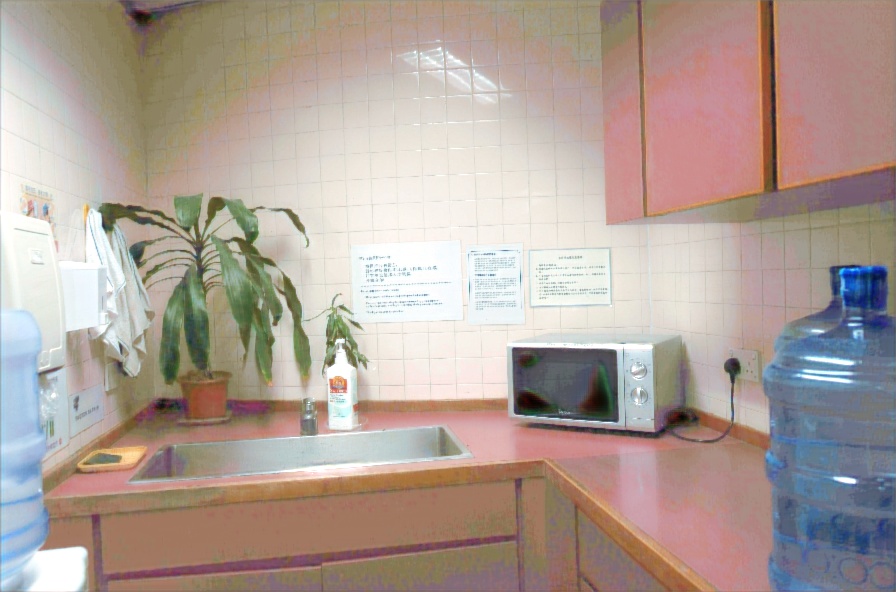}
        \caption*{Dimma 3 pairs (64\% lum)}
    \end{subfigure}
    
    \hspace{0.22\textwidth}
    \begin{subfigure}{0.22\textwidth}
        \includegraphics[width=\linewidth]{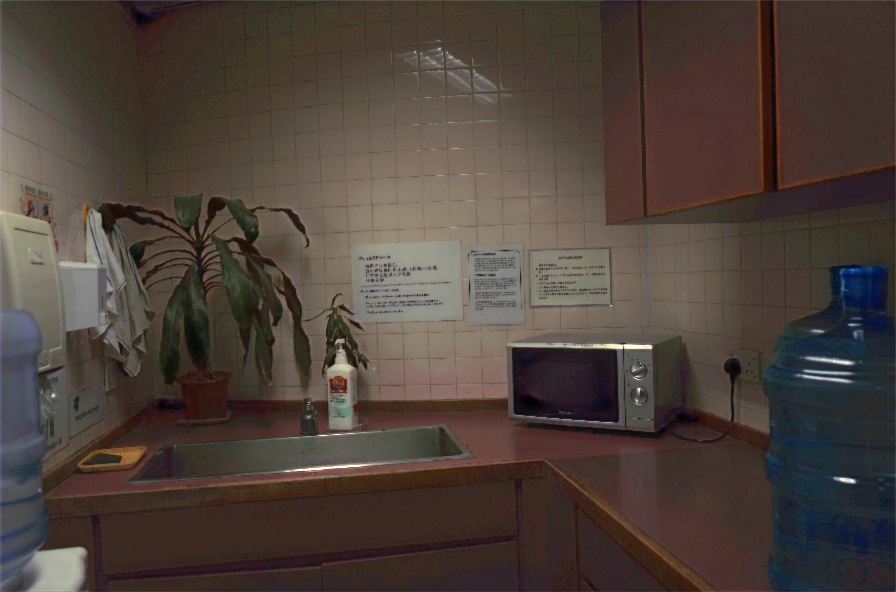}
        \caption*{Dimma 5 pairs (26\% lum)}
    \end{subfigure}
    \begin{subfigure}{0.22\textwidth}
        \includegraphics[width=\linewidth]{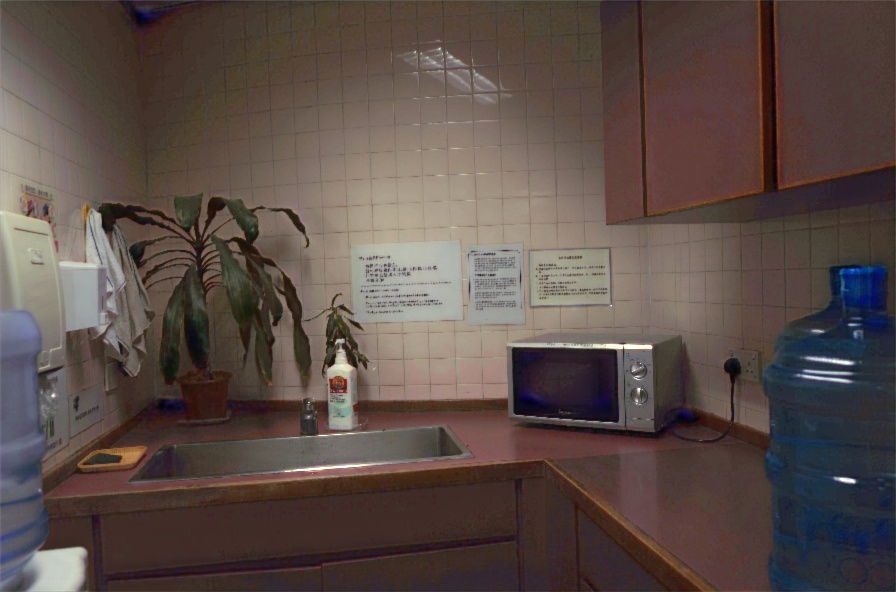}
        \caption*{Dimma 5 pairs (31\% lum)}
    \end{subfigure}
    \begin{subfigure}{0.22\textwidth}
        \includegraphics[width=\linewidth]{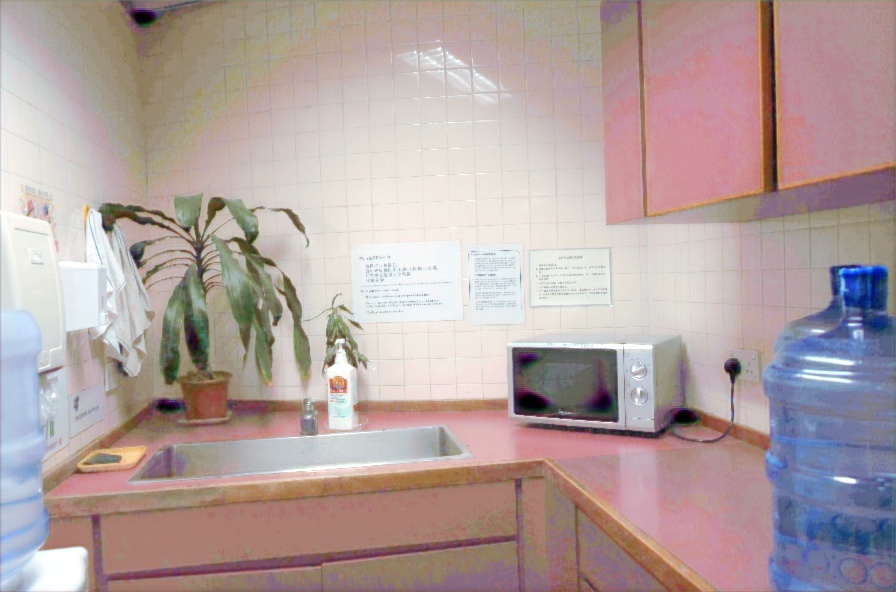}
        \caption*{Dimma 5 pairs (65\% lum)}
    \end{subfigure}
    
    \hspace{0.22\textwidth}
    \begin{subfigure}{0.22\textwidth}
        \includegraphics[width=\linewidth]{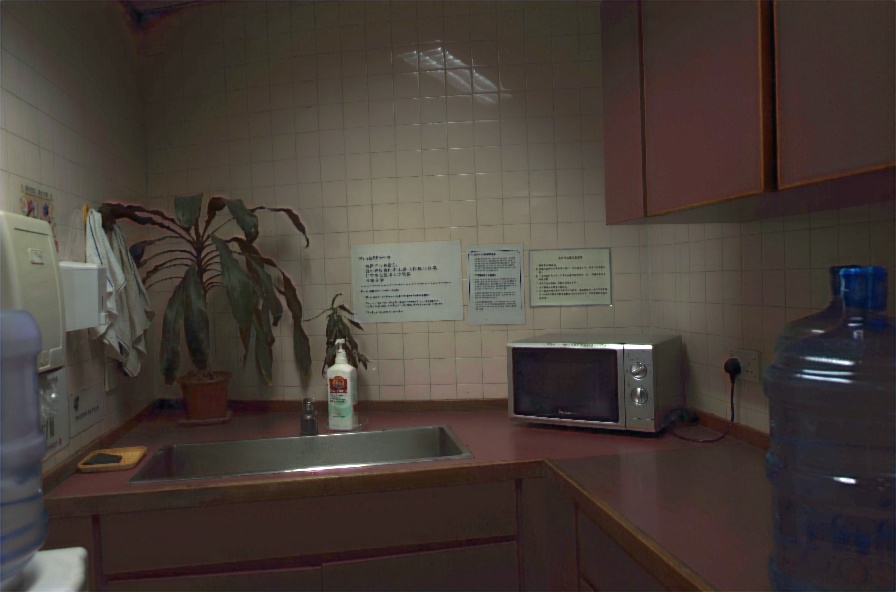}
        \caption*{Dimma 8 pairs (25\% lum)}
    \end{subfigure}
    \begin{subfigure}{0.22\textwidth}
        \includegraphics[width=\linewidth]{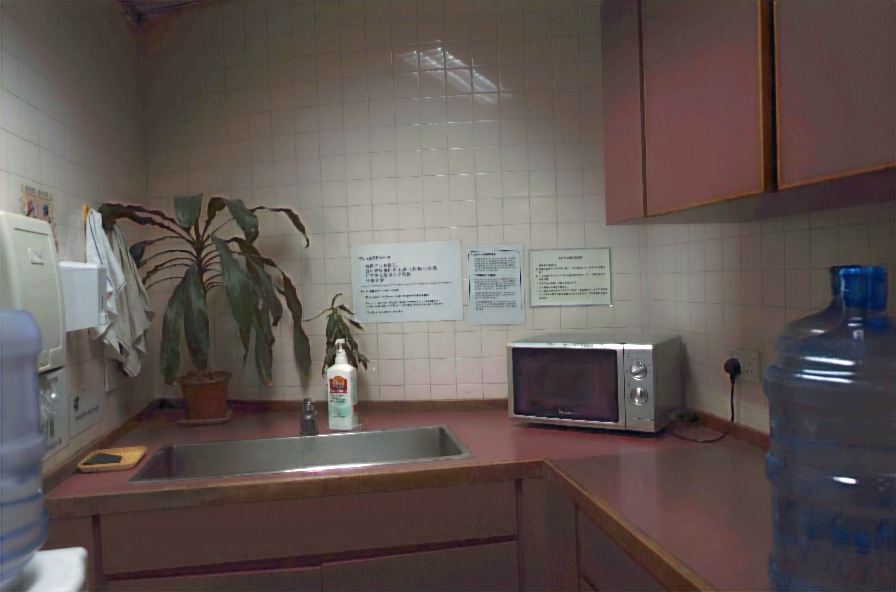}
        \caption*{Dimma 8 pairs (35\% lum)}
    \end{subfigure}
    \begin{subfigure}{0.22\textwidth}
        \includegraphics[width=\linewidth]{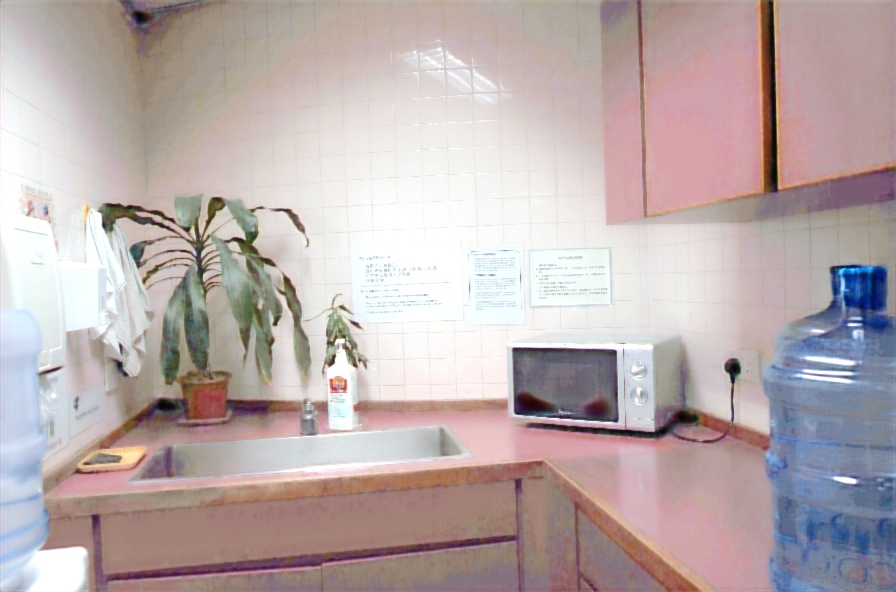}
        \caption*{Dimma 8 pairs (69\% lum)}
    \end{subfigure}

    \hspace{0.22\textwidth}
    \begin{subfigure}{0.22\textwidth}
        \includegraphics[width=\linewidth]{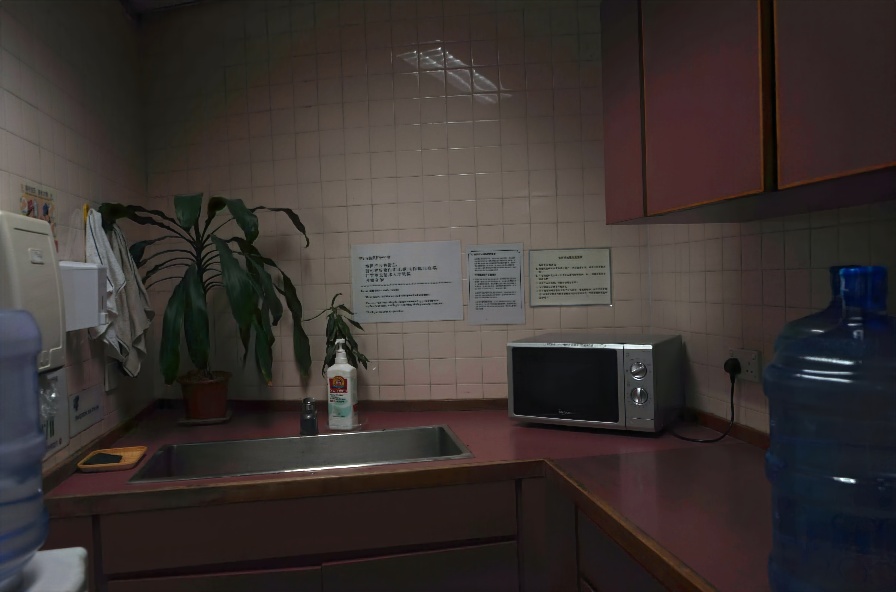}
        \caption*{Dimma full (21\% lum)}
    \end{subfigure}
    \begin{subfigure}{0.22\textwidth}
        \includegraphics[width=\linewidth]{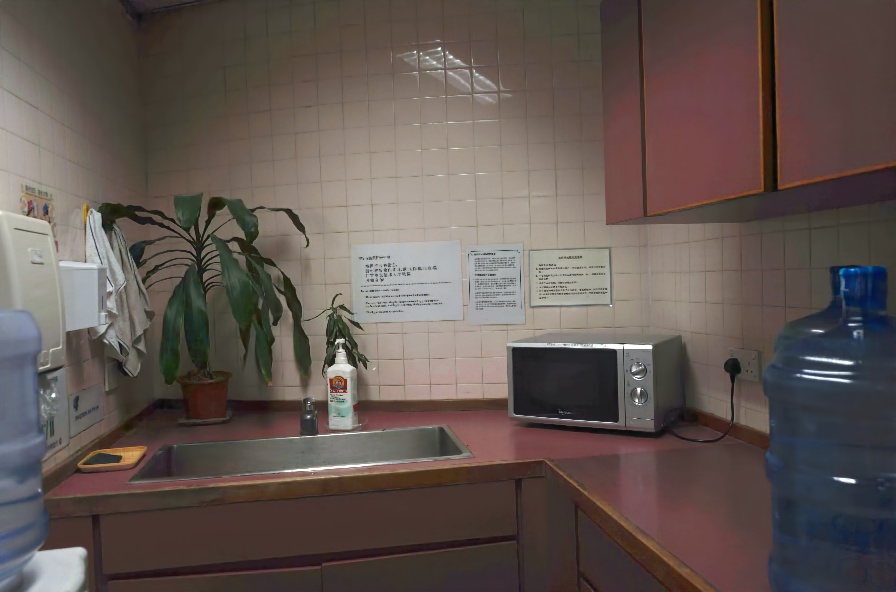}
        \caption*{Dimma full (35\% lum)}
    \end{subfigure}
    \begin{subfigure}{0.22\textwidth}
        \includegraphics[width=\linewidth]{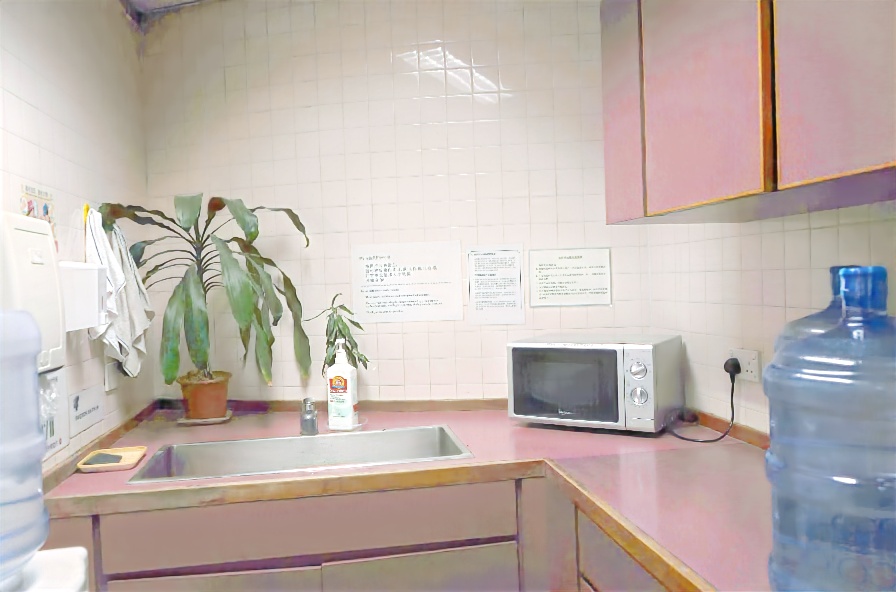}
        \caption*{Dimma full (71\% lum)}
    \end{subfigure}
    
    \begin{subfigure}{0.22\textwidth}
        \includegraphics[width=\linewidth]{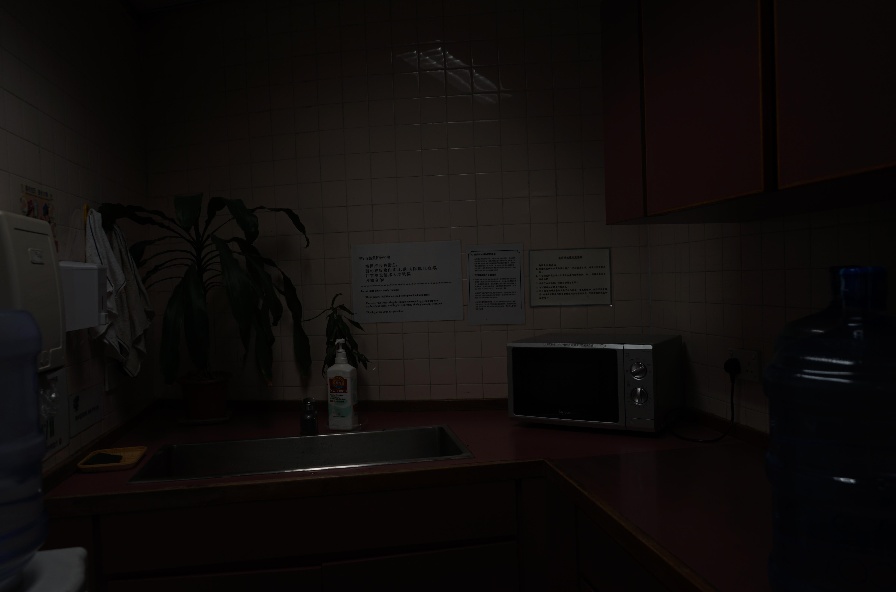}
        \caption*{Input (5\% lum)}
    \end{subfigure}
    \begin{subfigure}{0.22\textwidth}
        \includegraphics[width=\linewidth]{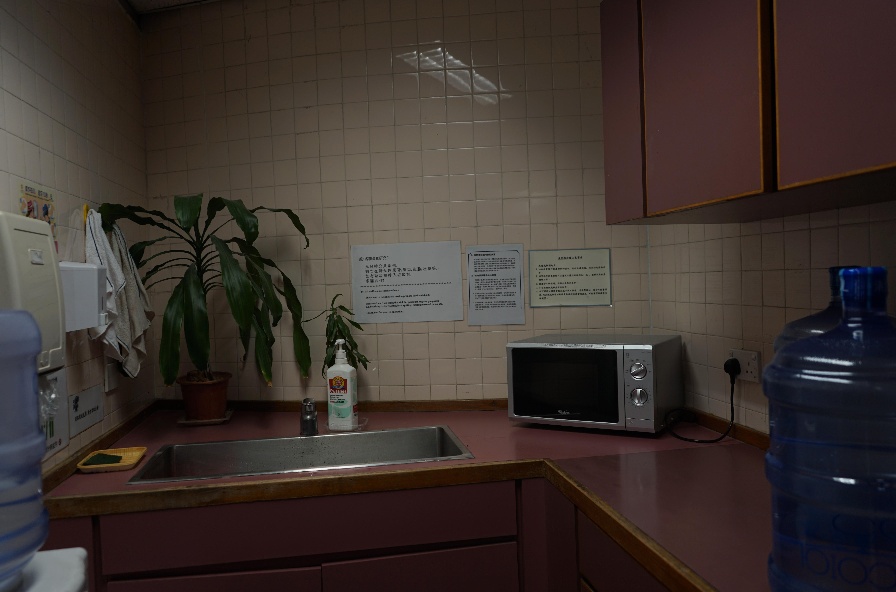}
        \caption*{Ground truth (20\% lum)}
    \end{subfigure}
    \begin{subfigure}{0.22\textwidth}
        \includegraphics[width=\linewidth]{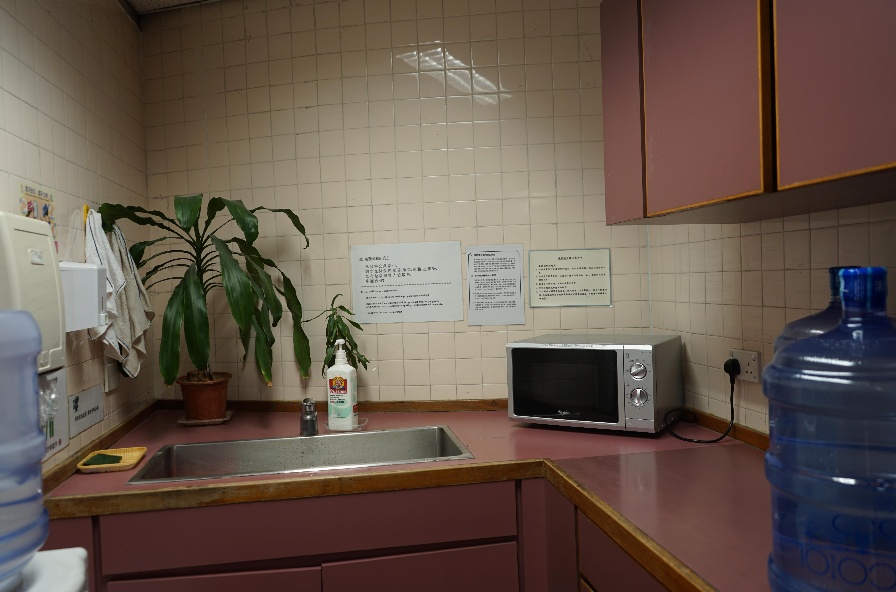}
        \caption*{Ground truth (35\% lum)}
    \end{subfigure}
    \begin{subfigure}{0.22\textwidth}
        \includegraphics[width=\linewidth]{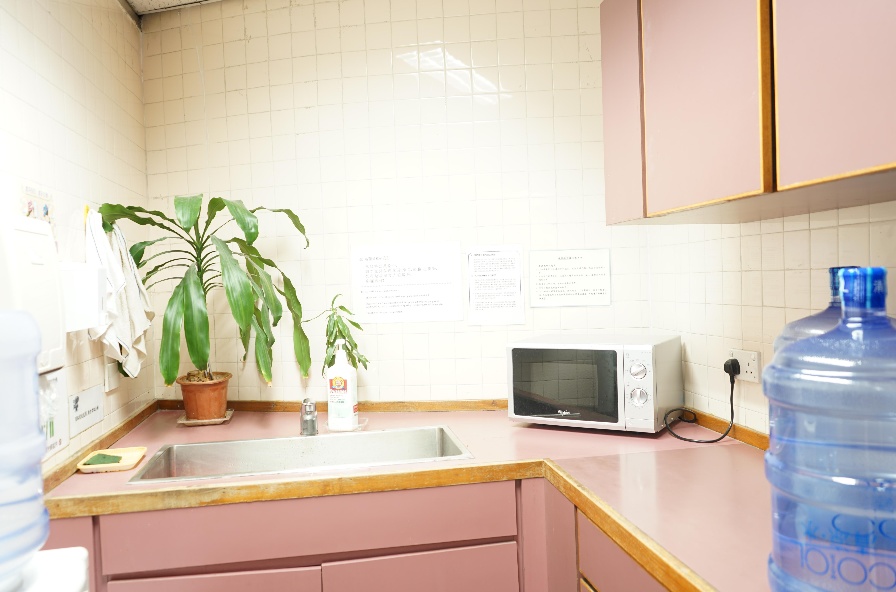}
        \caption*{Ground truth (79\% lum)}
    \end{subfigure}
    
    \caption{Visualization of low light enhancement with different brightening factors for Dimma and KinD++ on the image from SICE dataset. Both models were conditioned by the illumination values of the ground truth images displayed at the bottom. The average illumination of each image is shown in brackets, indicating the accuracy of the models in generating images with specific light levels.}
    \label{fig:dimma_light_results_3}
\end{figure*}

Figures \ref{fig:dimma_light_results}, \ref{fig:dimma_light_results_2}, and \ref{fig:dimma_light_results_3} demonstrate Dimma's remarkable ability to produce images with various desired levels of brightness. Each model was guided by the illumination of the ground truth images presented at the bottom of the figures, aiming to reproduce the distinct brightness levels from the corresponding dark inputs. Dimma excels in generating images that closely match the exact brightness of the ground truth images. Moreover, the model shows its ability to generate convincingly over-exposed images. 

\begin{figure*}
    \centering
    \begin{subfigure}{0.32\textwidth}
        \caption*{\Large Inputs}
        \includegraphics[width=\textwidth]{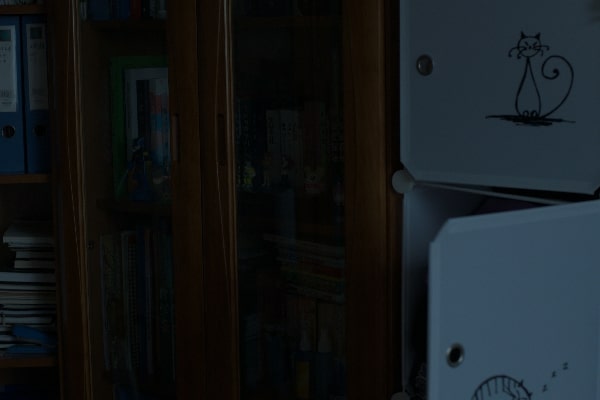}
    \end{subfigure}
    \begin{subfigure}{0.32\textwidth}
        \caption*{\Large Predictions}
        \includegraphics[width=\textwidth]{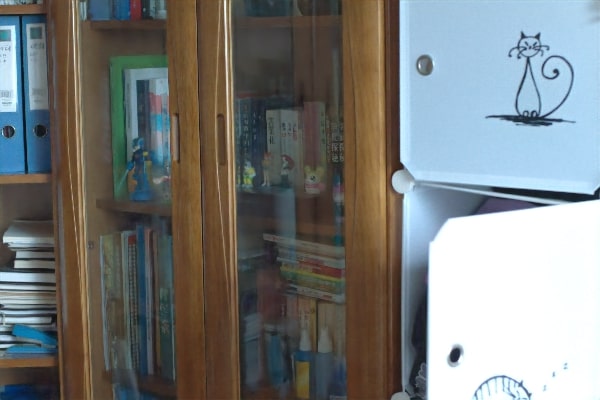}
    \end{subfigure}
    \begin{subfigure}{0.32\textwidth}
        \caption*{\Large Ground truth}
        \includegraphics[width=\textwidth]{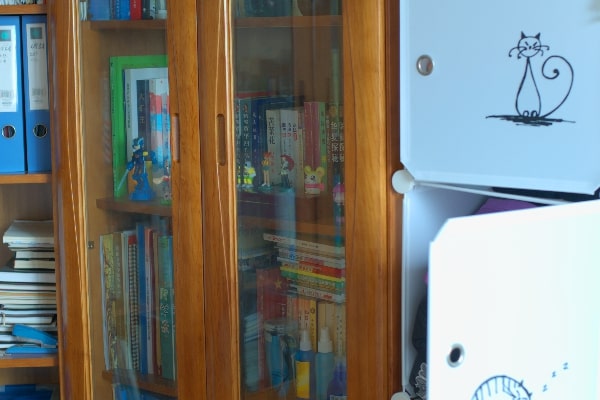}
    \end{subfigure}
    
    \includegraphics[width=0.32\textwidth]{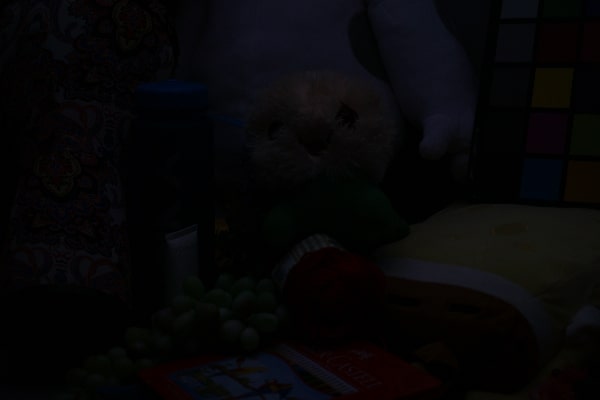}
    \includegraphics[width=0.32\textwidth]{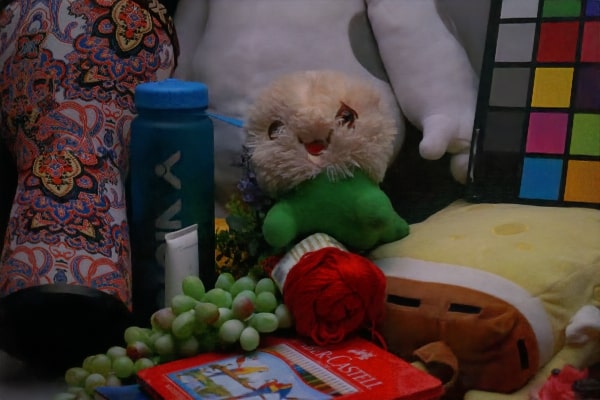}
    \includegraphics[width=0.32\textwidth]{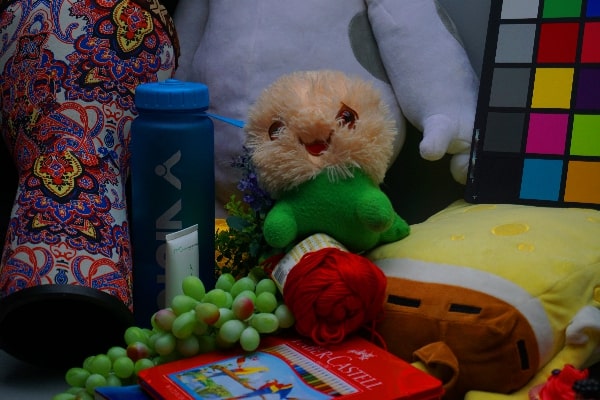}
    
    \includegraphics[width=0.32\textwidth]{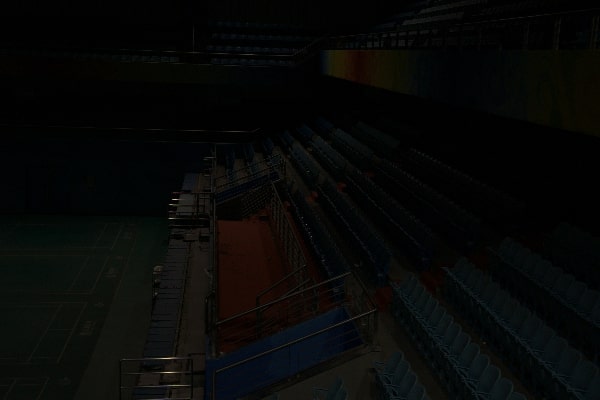}
    \includegraphics[width=0.32\textwidth]{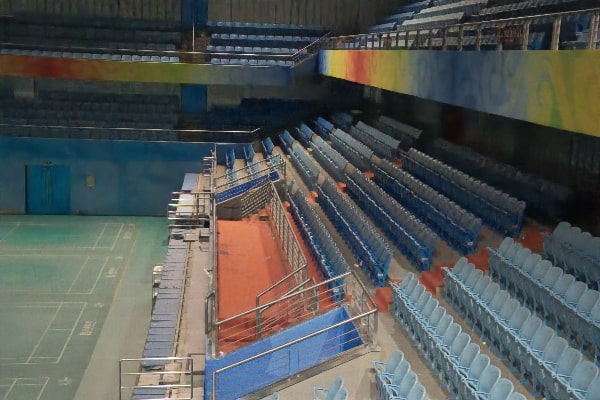}
    \includegraphics[width=0.32\textwidth]{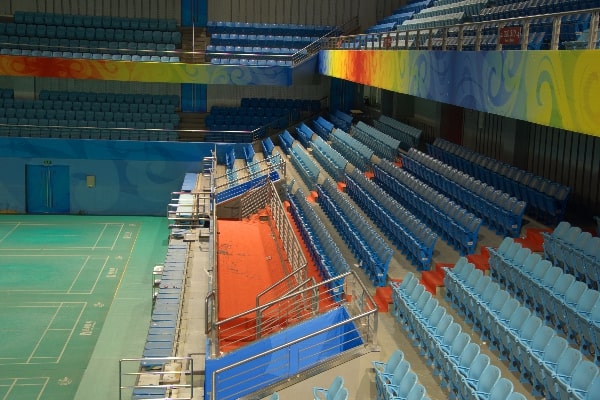}
    
    \includegraphics[width=0.32\textwidth]{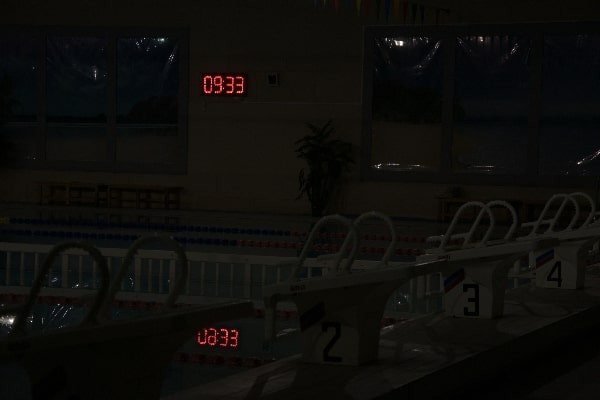}
    \includegraphics[width=0.32\textwidth]{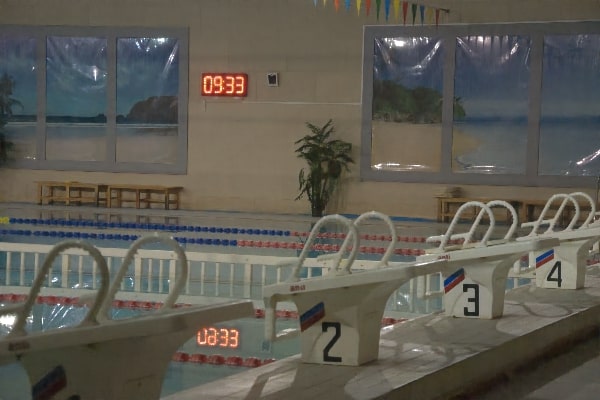}
    \includegraphics[width=0.32\textwidth]{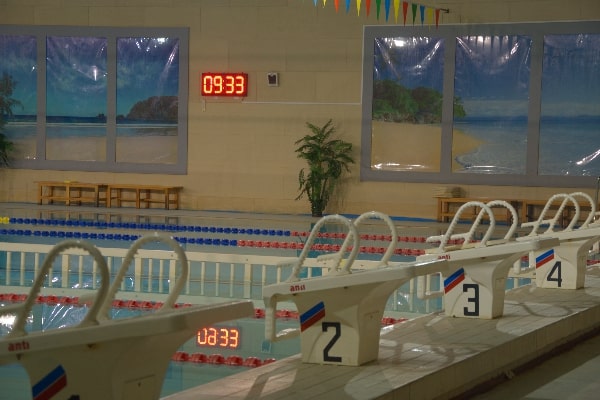}
    
    \includegraphics[width=0.32\textwidth]{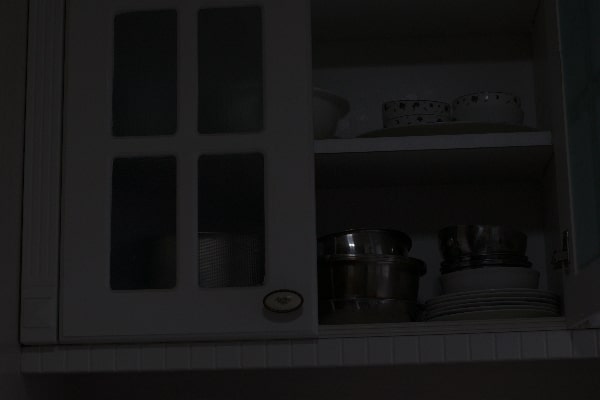}
    \includegraphics[width=0.32\textwidth]{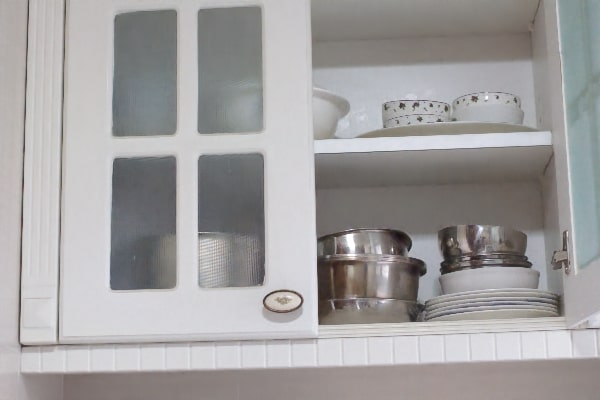}
    \includegraphics[width=0.32\textwidth]{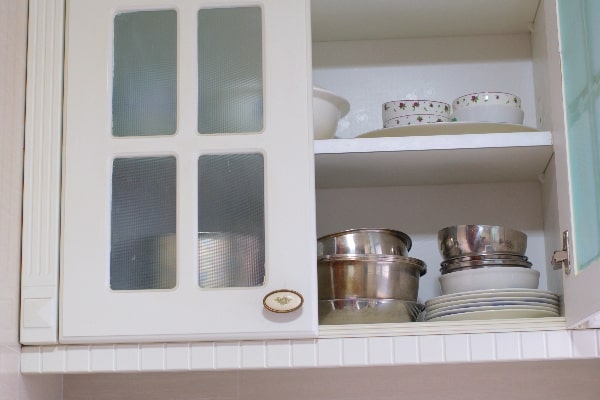}
    
    \includegraphics[width=0.32\textwidth]{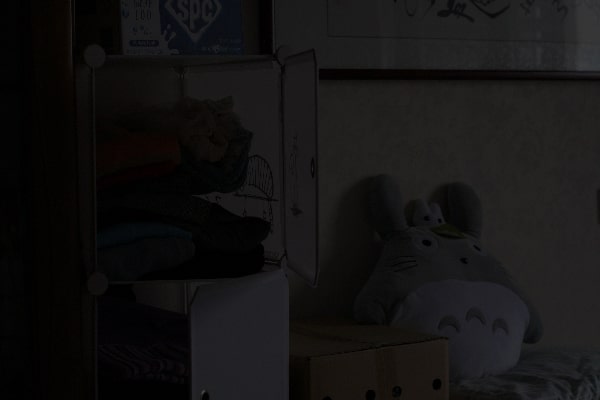}
    \includegraphics[width=0.32\textwidth]{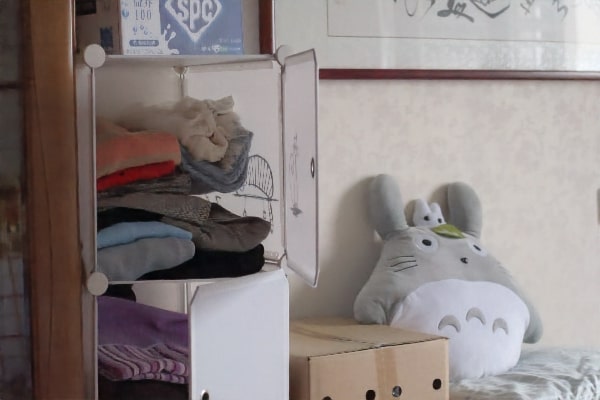}
    \includegraphics[width=0.32\textwidth]{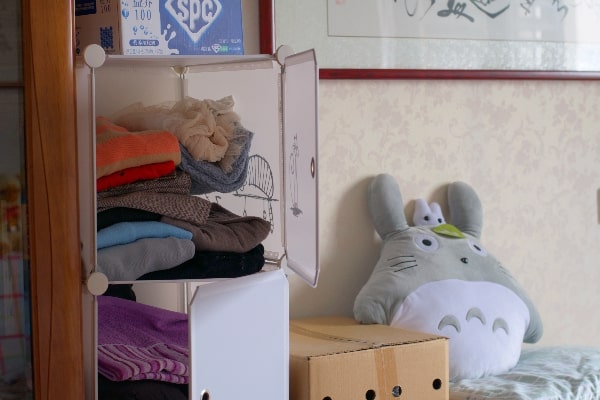}
    \caption{Visual results of Dimma model trained on 480 pairs from LOL dataset.}
    \label{fig:dimma_lol_visualization}
\end{figure*}

\begin{figure*}
    \centering
    \begin{subfigure}{0.32\textwidth}
        \caption*{\Large Inputs}
        \includegraphics[width=\textwidth]{imgs/paper/lol/dark/1.jpg}
    \end{subfigure}
    \begin{subfigure}{0.32\textwidth}
        \caption*{\Large Predictions}
        \includegraphics[width=\textwidth]{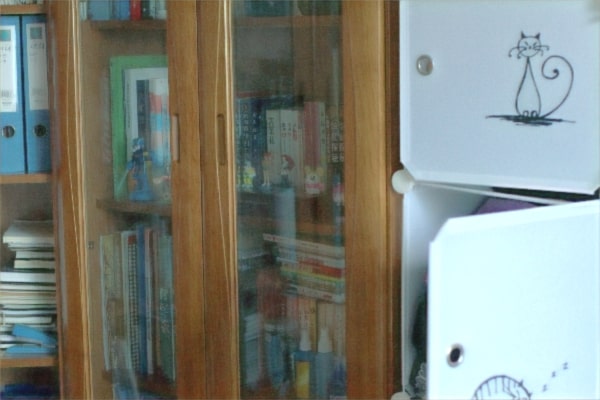}
    \end{subfigure}
    \begin{subfigure}{0.32\textwidth}
        \caption*{\Large Ground truth}
        \includegraphics[width=\textwidth]{imgs/paper/lol/GT/1.jpg}
    \end{subfigure}
    
    \includegraphics[width=0.32\textwidth]{imgs/paper/lol/dark/493.jpg}
    \includegraphics[width=0.32\textwidth]{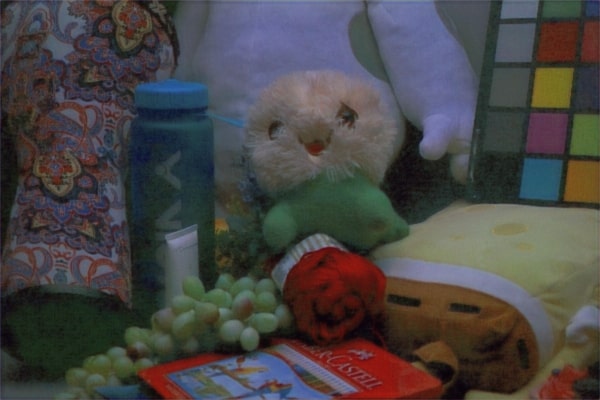}
    \includegraphics[width=0.32\textwidth]{imgs/paper/lol/GT/493.jpg}
    
    \includegraphics[width=0.32\textwidth]{imgs/paper/lol/dark/778.jpg}
    \includegraphics[width=0.32\textwidth]{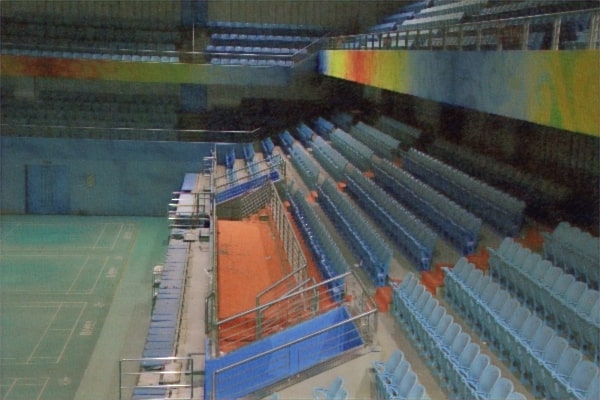}
    \includegraphics[width=0.32\textwidth]{imgs/paper/lol/GT/778.jpg}
    
    \includegraphics[width=0.32\textwidth]{imgs/paper/lol/dark/748.jpg}
    \includegraphics[width=0.32\textwidth]{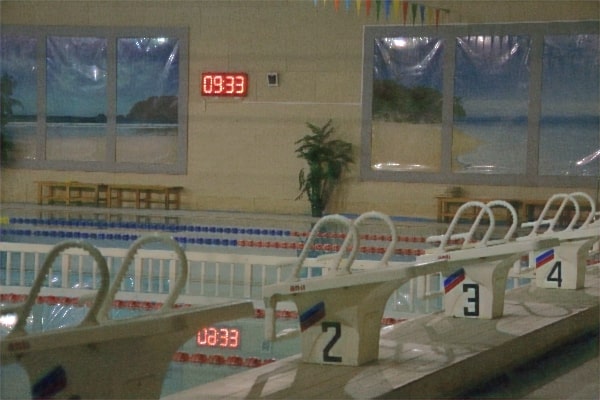}
    \includegraphics[width=0.32\textwidth]{imgs/paper/lol/GT/748.jpg}
    
    \includegraphics[width=0.32\textwidth]{imgs/paper/lol/dark/79.jpg}
    \includegraphics[width=0.32\textwidth]{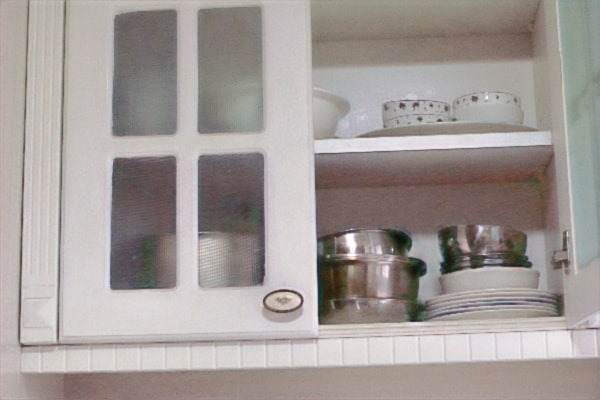}
    \includegraphics[width=0.32\textwidth]{imgs/paper/lol/GT/79.jpg}
    
    \includegraphics[width=0.32\textwidth]{imgs/paper/lol/dark/22.jpg}
    \includegraphics[width=0.32\textwidth]{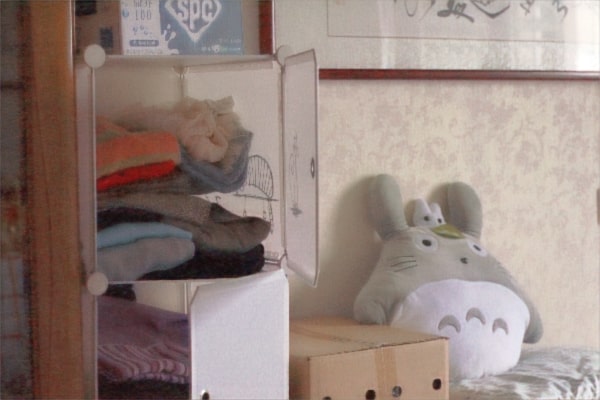}
    \includegraphics[width=0.32\textwidth]{imgs/paper/lol/GT/22.jpg}
    \caption{Visual results of Dimma model trained on 3 pairs from LOL dataset.}
    \label{fig:dimma_3shot_lol_visualization}
\end{figure*}

\begin{figure*}
    \centering
    \begin{subfigure}{0.32\textwidth}
        \caption*{\Large Inputs}
        \includegraphics[width=\textwidth]{imgs/paper/lol/dark/1.jpg}
    \end{subfigure}
    \begin{subfigure}{0.32\textwidth}
        \caption*{\Large Predictions}
        \includegraphics[width=\textwidth]{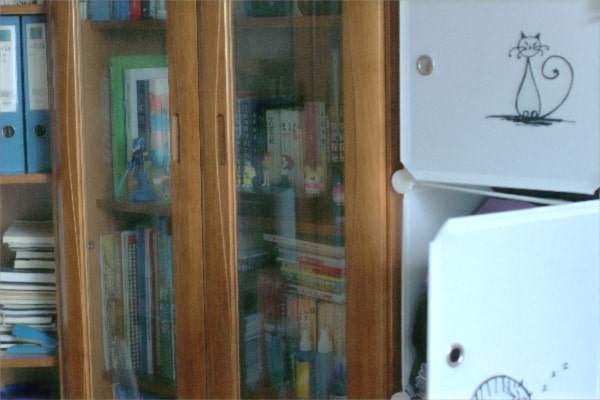}
    \end{subfigure}
    \begin{subfigure}{0.32\textwidth}
        \caption*{\Large Ground truth}
        \includegraphics[width=\textwidth]{imgs/paper/lol/GT/1.jpg}
    \end{subfigure}
    
    \includegraphics[width=0.32\textwidth]{imgs/paper/lol/dark/493.jpg}
    \includegraphics[width=0.32\textwidth]{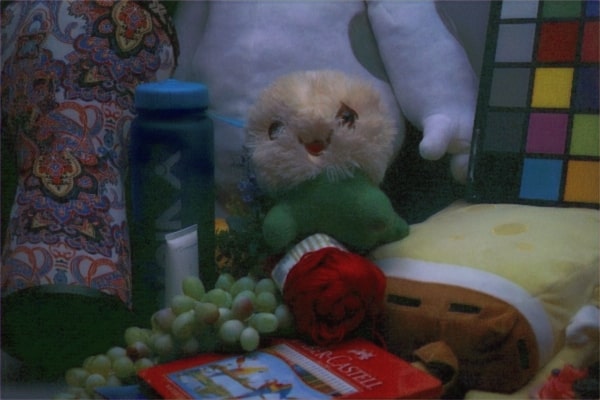}
    \includegraphics[width=0.32\textwidth]{imgs/paper/lol/GT/493.jpg}
    
    \includegraphics[width=0.32\textwidth]{imgs/paper/lol/dark/778.jpg}
    \includegraphics[width=0.32\textwidth]{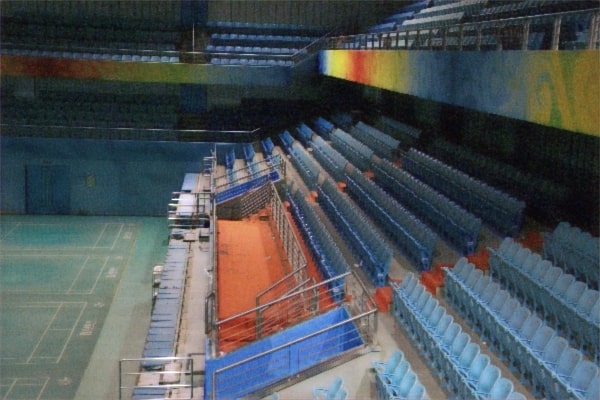}
    \includegraphics[width=0.32\textwidth]{imgs/paper/lol/GT/778.jpg}
    
    \includegraphics[width=0.32\textwidth]{imgs/paper/lol/dark/748.jpg}
    \includegraphics[width=0.32\textwidth]{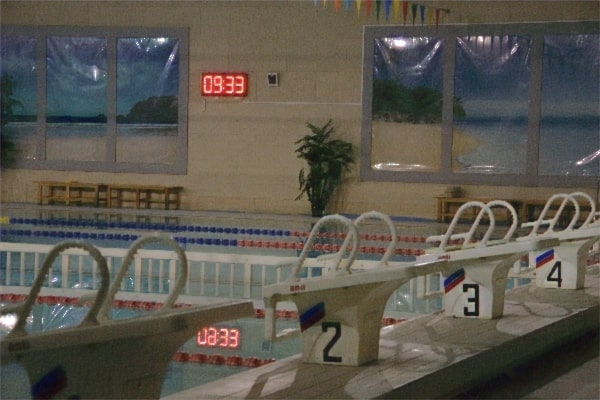}
    \includegraphics[width=0.32\textwidth]{imgs/paper/lol/GT/748.jpg}
    
    \includegraphics[width=0.32\textwidth]{imgs/paper/lol/dark/79.jpg}
    \includegraphics[width=0.32\textwidth]{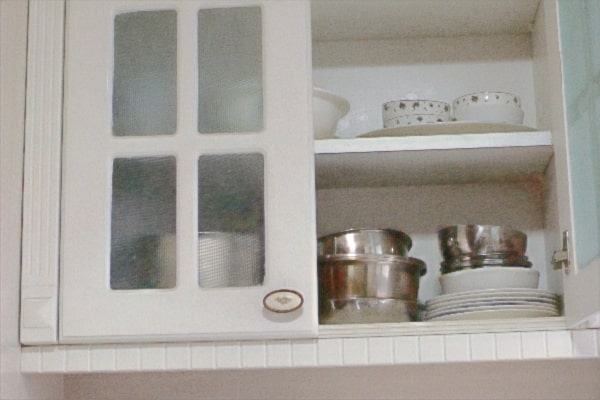}
    \includegraphics[width=0.32\textwidth]{imgs/paper/lol/GT/79.jpg}
    
    \includegraphics[width=0.32\textwidth]{imgs/paper/lol/dark/22.jpg}
    \includegraphics[width=0.32\textwidth]{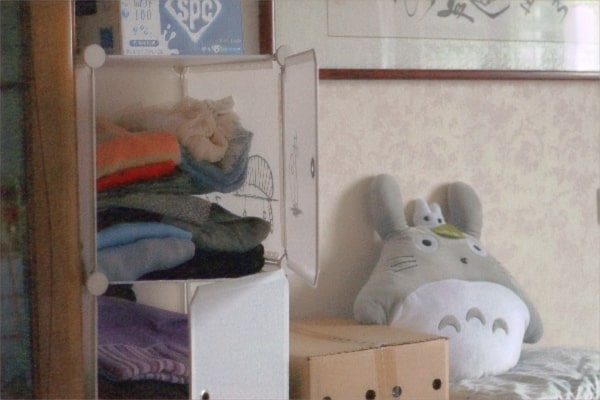}
    \includegraphics[width=0.32\textwidth]{imgs/paper/lol/GT/22.jpg}
    \caption{Visual results of Dimma model trained on 5 pairs from LOL dataset.}
    \label{fig:dimma_5shot_lol_visualization}
\end{figure*}

\begin{figure*}
    \centering
    \begin{subfigure}{0.32\textwidth}
        \caption*{\Large Inputs}
        \includegraphics[width=\textwidth]{imgs/paper/lol/dark/1.jpg}
    \end{subfigure}
    \begin{subfigure}{0.32\textwidth}
        \caption*{\Large Predictions}
        \includegraphics[width=\textwidth]{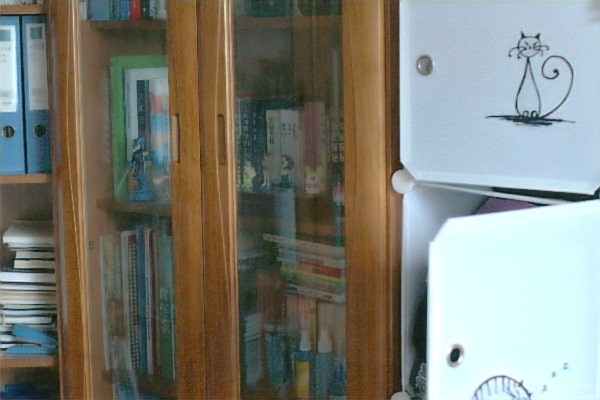}
    \end{subfigure}
    \begin{subfigure}{0.32\textwidth}
        \caption*{\Large Ground truth}
        \includegraphics[width=\textwidth]{imgs/paper/lol/GT/1.jpg}
    \end{subfigure}
    
    \includegraphics[width=0.32\textwidth]{imgs/paper/lol/dark/493.jpg}
    \includegraphics[width=0.32\textwidth]{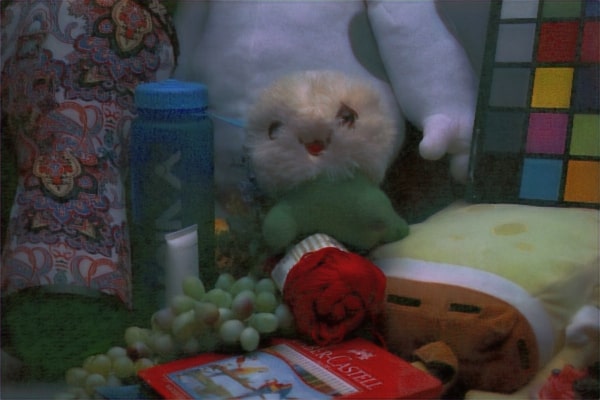}
    \includegraphics[width=0.32\textwidth]{imgs/paper/lol/GT/493.jpg}
    
    \includegraphics[width=0.32\textwidth]{imgs/paper/lol/dark/778.jpg}
    \includegraphics[width=0.32\textwidth]{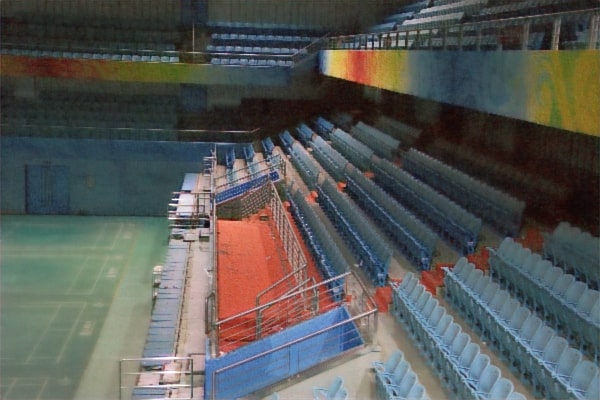}
    \includegraphics[width=0.32\textwidth]{imgs/paper/lol/GT/778.jpg}
    
    \includegraphics[width=0.32\textwidth]{imgs/paper/lol/dark/748.jpg}
    \includegraphics[width=0.32\textwidth]{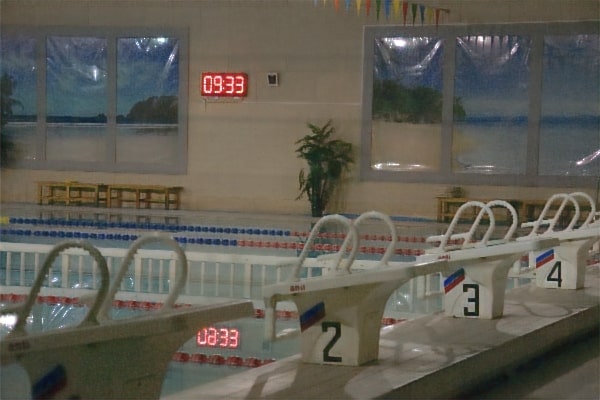}
    \includegraphics[width=0.32\textwidth]{imgs/paper/lol/GT/748.jpg}
    
    \includegraphics[width=0.32\textwidth]{imgs/paper/lol/dark/79.jpg}
    \includegraphics[width=0.32\textwidth]{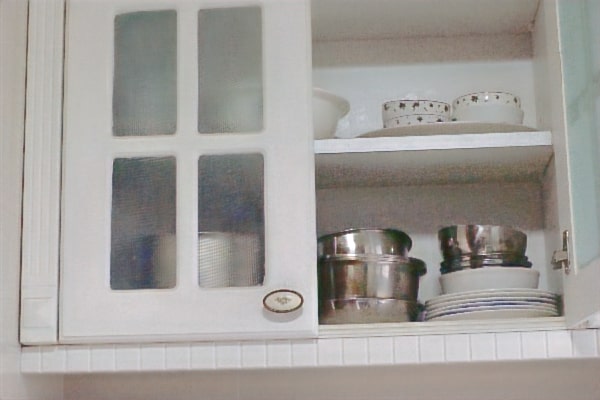}
    \includegraphics[width=0.32\textwidth]{imgs/paper/lol/GT/79.jpg}
    
    \includegraphics[width=0.32\textwidth]{imgs/paper/lol/dark/22.jpg}
    \includegraphics[width=0.32\textwidth]{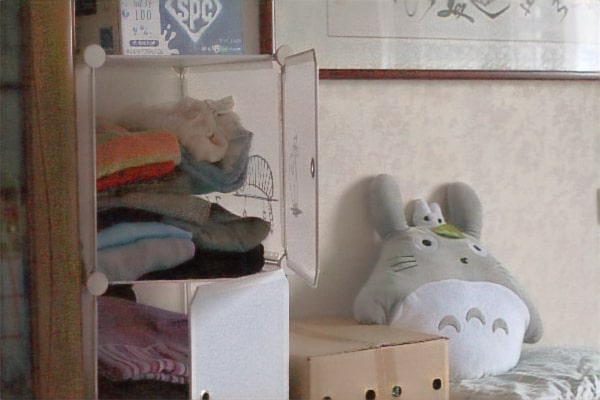}
    \includegraphics[width=0.32\textwidth]{imgs/paper/lol/GT/22.jpg}
    \caption{Visual results of Dimma model trained on 8 pairs from LOL dataset.}
    \label{fig:dimma_8shot_lol_visualization}
\end{figure*}

Figures \ref{fig:dimma_lol_visualization}, \ref{fig:dimma_3shot_lol_visualization}, \ref{fig:dimma_5shot_lol_visualization}, \ref{fig:dimma_8shot_lol_visualization} present the results on LOL dataset produced by our four models trained on respectively 480, 3, 5 and 8 image pairs.

\end{document}